\documentclass{article}

\usepackage[preprint]{corl_2026} 

\usepackage{times}
\usepackage{helvet}
\usepackage{courier}
\usepackage[T1]{fontenc}
\usepackage[utf8]{inputenc}
\usepackage{microtype}
\usepackage{amsmath,amssymb,mathtools,bm}
\usepackage{booktabs,multirow,array}
\usepackage{graphicx,subcaption}
\usepackage{microtype}

\usepackage{url}
\usepackage{hyperref}
\usepackage{cleveref}
\usepackage{algorithm,algorithmic}
\usepackage{enumitem}
\usepackage{caption}
\usepackage{ragged2e}
\usepackage{xcolor}
\usepackage[table]{xcolor}

\newcommand{\R}{\mathbb{R}}
\newcommand{\E}{\mathbb{E}}
\newcommand{\Jr}{J_R}
\newcommand{\Jc}{J_C}
\newcommand{\vlm}{\text{vlm}}

\newcommand{\rvlm}{r_{\vlm}}
\newcommand{\cvlm}{c_{\vlm}}
\newcommand{\meancvlm}{\overline{c}_{\vlm}}
\newcommand{\dlim}{d}
\newcommand{\eqnref}[1]{Eq.~\eqref{#1}}

\title{\textbf{Seeing Before Colliding:\\
Anticipatory Safe RL with Frozen Vision-Language Models}}

\author{
Samuel Tetteh \& Cody Fleming \\
Iowa State University \\
Ames, Iowa, USA\\
\texttt{{\big\{}samtett, flemingc{\big\}}@iastate.edu}
}

\begin{document}
\maketitle

\begin{abstract}
The cost signal that constrained-RL algorithms optimize against is
almost always \emph{reactive}: the simulator emits a non-zero cost
only \emph{after} a collision has begun, and the Lagrange multiplier
of PPO-Lagrangian grows only after the episode budget has been
exceeded. At race speeds, where collisions are instantaneous and
irreversible, any safety mechanism that waits for cost to accumulate
is structurally too late. We present \textbf{VLM-Safe-RL}, a framework
that integrates a frozen vision-language model into the CMDP
Lagrangian update as an \emph{anticipatory} cost term. The framework
comprises four contributions: (i)~\textbf{Decoupled Dual-Path CLIP},
independent reward/cost paths that respect the CMDP's $r\!\perp\!c$
factorization; (ii)~\textbf{VLMLagrange}, an augmented multiplier
update $\lambda \!\leftarrow\! \lambda + \eta_1(\Jc-\dlim)+
\eta_2(\meancvlm-\tau)$ that incorporates a per-step VLM cost as an
anticipatory term; (iii)~\textbf{Confidence Gating}, a Bayes-optimal
weight $\kappa{=}|2\sigma(s(m{-}c)){-}1|$ derived from a logistic
noise model on the CLIP margin; and (iv)~\textbf{VLMPPOLag}, the
composed algorithm. On Safety-Gymnasium FormulaOne L2, our principal
evaluation ($n{=}5$ seeds, $10^{6}$ steps, budget $\dlim{=}25$) %
VLMPPOLag$+$Conf is the only configuration in our default-budget 
comparison that simultaneously retains substantive return ($\Jr{\approx}40$) and
holds cost within budget on a majority of seeds; the five
constraint-aware baselines (PPOLag, CPO, CPPOPID, CPO-CLG,
PPOLag-RND) each fail at least one requirement. The mechanism
generalises to held-out MetaDrive Medium (catastrophe rate
$41\%{\to}26\%$, 95\% bootstrap CI $[-26,-5]$\,pp) and shows
directionally consistent transfer to Bullet Safety-Gym; we report
honestly where it does \emph{not} (MetaDrive Easy/Hard, Qwen2-VL
backbone) and trace the Hard failure to a Lagrangian-regulation
pathology rather than the VLM signal itself. To our knowledge this is
the first work to use frozen VLM signals as an anticipatory cost term
inside the CMDP Lagrangian update.
\end{abstract}


\section{Introduction}
\label{sec:intro}

Safe reinforcement learning in high-speed visual control presents a
fundamental tension: an agent must push close to the limits of
performance while strictly avoiding catastrophic failure---a setting
formalized by the Constrained Markov Decision
Process (CMDP) framework~\cite{altman1999constrained}. Yet the
cost signals available to standard CMDP solvers are almost
universally \emph{reactive}: the simulator emits a non-zero cost
only \emph{after} a collision has begun, and PPO-Lagrangian's
multiplier $\lambda$~\cite{stooke2020responsive} grows only after
the episode budget has been exceeded. At race speeds, where contacts
are instantaneous and irreversible, this lag is structural.
Vision-language models~\cite{radford2021learning} encode rich
semantic priors about safe and unsafe states. A natural-language
description, ``the racecar is about to crash into the
barrier'' captures the visual signature of impending danger that a
hand-crafted feature would require elaborate engineering to detect.
Existing VLM+RL paradigms either fine-tune billion-parameter
vision-language-action models on large demonstration
sets~\cite{brohan2023rt2,driess2023palme,zhang2025safevla}, or use
frozen VLMs as auxiliary reward
shapers~\cite{fan2022minedojo,kwon2023reward,rocamonde2024vlmrm,
huang2024vlmrl}.
Neither addresses safety as a hard constraint. Reward-shaping methods
treat safety as an implicit penalty, with VLM-RL
\cite{huang2024vlmrl} further coupling reward and safety scores
on a shared simplex via the contrasting language goal (CLG)
paradigm; SafeVLA~\cite{zhang2025safevla} adds CMDP
objectives but at $7$B+ parameters and $\sim\!800$K demonstrations.

\paragraph{Central insight.}
Frozen CLIP can detect visual danger signals \emph{before} the actual
collision. Routed through the Lagrange multiplier update, this
forward-looking information enables \emph{anticipatory} constraint
satisfaction fundamentally distinct from prior VLM+RL approaches
that treat the VLM output as a stateless reward bonus.

\paragraph{Contributions.} We present \textbf{VLM-Safe-RL}:
\begin{enumerate}[leftmargin=20pt,itemsep=1pt,topsep=2pt]
  \item \textbf{Decoupled Dual-Path CLIP} (\S\ref{sec:method}): two
    independent cosine paths for $\rvlm$ and $\cvlm$ that
    eliminate the anti-correlation artefact of coupled softmax.
  \item \textbf{VLMLagrange} (\S\ref{sec:method}): an augmented
    multiplier update with a per-step CLIP-derived anticipatory term
    that tightens $\lambda$ before collisions accumulate.
  \item \textbf{Confidence Gating} (\S\ref{sec:method}): a
    Bayes-optimal weight $\kappa_t$ derived in closed form from a
    logistic noise model on the CLIP margin, with a calibrated
    operating point estimated from a random-policy frame buffer.
  \item \textbf{VLMPPOLag}: the composed algorithm, registered as a
    first-class algorithm in OmniSafe~\cite{ji2023omnisafe}.
\end{enumerate}

\noindent We evaluate on Safety-Gymnasium
FormulaOne~\cite{ji2023safety} L0/L1/L2 (10 methods $\times$ 3 levels,
3--5 seeds; 90+ training runs) and on two generalization
benchmarks, Bullet Safety-Gym \texttt{SafetyCarReach-v0} and
MetaDrive~\cite{li2023metadrive} Easy/Medium/Hard, with held-out
evaluation on seeds disjoint from training. The compressed
contribution: \textbf{VLMPPOLag$+$Conf is the only constraint-aware
configuration that achieves substantive return with within-budget
cost on FormulaOne L2}, and the mechanism transfers to dense traffic
($-15$\,pp catastrophe on held-out MetaDrive Medium, bootstrap CI
excludes zero). We additionally document a previously-unreported
scenario-sampler aliasing in MetaDrive that would otherwise mask any
held-out safety improvement (Appendix~\ref{app:metadrive-bug}).

\begin{figure*}[t]
  \centering
  \includegraphics[width=\linewidth]{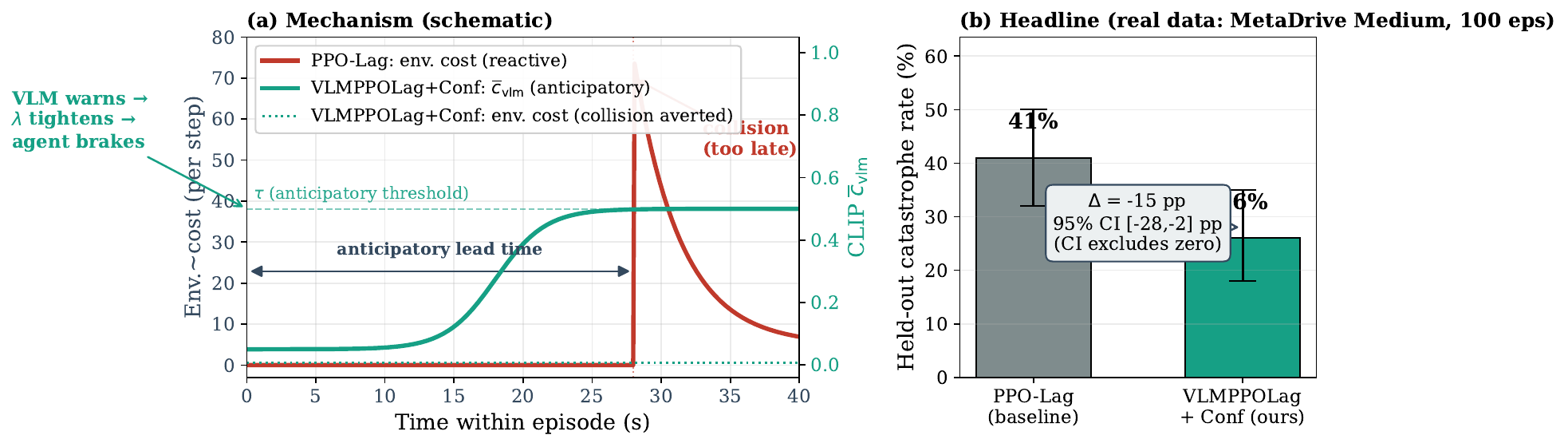}
  \caption{\textbf{Anticipatory safety from a frozen VLM.}
    \emph{(a)}~Per-step CLIP danger signal $\cvlm$ (green) rises
    several timesteps before the environment cost (red) on a single
    FormulaOne~L2 rollout. The Lagrange multiplier $\lambda$ is updated
    \emph{between epochs} using the epoch mean $\meancvlm$ via
    \eqnref{eq:vlm-lagrange}; the per-step trace illustrates the
    signal that this mean accumulates.
    \emph{(b)}~Held-out MetaDrive Medium: the mechanism cuts
    catastrophe rate from $41\%$ to $26\%$ ($-15$\,pp; bootstrap
    $95\%$ CI $[-26,-5]$\,pp, $n{=}100$ held-out episodes).}
  \label{fig:teaser}
\end{figure*}
\section{Related Work}
\label{sec:related}

\textbf{VLMs for robotics and reward shaping.} VLMs have been used
for task planning and affordance grounding~\cite{ahn2022saycan,
huang2023voxposer} and as end-to-end vision-language-action
models~\cite{brohan2023rt2,driess2023palme,liu2023llava,
wang2024qwen2vl,zhang2025safevla}. Building on classical
reward-shaping foundations~\cite{ng1999policy}, recent works use
frozen VLMs as auxiliary signals~\cite{fan2022minedojo,kwon2023reward,
xie2024text2reward,ma2024eureka,rocamonde2024vlmrm}. None impose hard
safety constraints. \textbf{VLM-RL}~\cite{huang2024vlmrl} 
closest prior work, introduces the \emph{contrasting language goal
(CLG)-as-reward} paradigm, positive and negative natural-language
goals scored by frozen CLIP and combined through a coupled softmax
that anti-correlates the reward and safety channels under
unconstrained SAC~\cite{haarnoja2018soft}. We adopt CLG terminology
from~\cite{huang2024vlmrl} but operate inside the CMDP framework
with decoupled paths and an anticipatory multiplier update
(Tab.~\ref{tab:comparison}).

\textbf{Safe RL.} The classical motivation~\cite{amodei2016concrete,
shalev2016safe,garcia2015comprehensive,brunke2022safe} for hard
constraints in RL has produced a family of CMDP solvers: CPO~\cite{achiam2017constrained}, PPO-Lagrangian
\cite{stooke2020responsive}, FOCOPS~\cite{zhang2020first},
PCPO~\cite{yang2020projection}, and CUP~\cite{yang2022cup}, evaluated
through Safety-Gymnasium~\cite{ji2023safety,todorov2012mujoco} and
OmniSafe~\cite{ji2023omnisafe}. PID-Lagrangian
\cite{stooke2020responsive}, CRPO~\cite{xu2021crpo} and
Saut\'e-RL~\cite{sootla2022saute} modify the Lagrange dynamics
themselves; our anticipatory $\eta_2(\meancvlm-\tau)$ term is
orthogonal in that it injects a new \emph{forward-looking} signal
derived from a frozen VLM and could be combined with any of them.
Reproducibility concerns in deep RL~\cite{henderson2018matters,
agarwal2021deep} motivate our $5\!,\!000$-resample bootstrap CIs and
pre-registered one-sided permutation tests.

\textbf{VLMs as zero-shot scene classifiers.} A complementary line
treats frozen VLMs as out-of-loop classifiers atop a separately
trained policy: action shielding~\cite{chen2026alphaadjcbf},
post-hoc anomaly detection~\cite{jeong2023winclip,murphy2012machine},
and language-conditioned scene tagging~\cite{santos2024updating,
rocamonde2024vlmrm}. Our work makes the per-step VLM output a
first-class citizen of the constrained-optimization problem the
policy itself solves.

\section{VLM-Safe-RL: Method}
\label{sec:method}

A CMDP is a tuple $(\mathcal{S}, \mathcal{A}, P, r, c, \dlim,
\gamma)$~\cite{altman1999constrained}; the safe-RL objective
\begin{equation}
  \pi^{\star} = \arg\max_\pi \Jr(\pi) \;\;\text{s.t.}\;\;
  \Jc(\pi) \le \dlim,
  \label{eq:cmdp}
\end{equation}
is solved by PPO-Lagrangian~\cite{stooke2020responsive} via the
update $\lambda \!\leftarrow\! \lambda + \eta_1(\Jc-\dlim)$, which is
strictly \emph{backward-looking}. We instantiate the framework on the
Safety-Gymnasium FormulaOne racing simulator~\cite{ji2023safety}: a
frozen CLIP ViT-B/32~\cite{radford2021learning,dosovitskiy2021image}
receives a $256{\times}256$ RGB frame at each control step alongside
the proprioceptive observation $s_t \in \R^{64}$; cost is binary on
barrier contact with budget $\dlim{=}25$ over $T{=}1000$ steps.

\textbf{Contribution 1: Decoupled Dual-Path CLIP.} Prior work
\cite{huang2024vlmrl,rocamonde2024vlmrm} uses a coupled softmax over
positive$+$negative prompt logits, forcing $\rvlm + \cvlm \approx 1$.
This is incorrect for CMDPs, where reward and cost are independent
objects by definition. We decouple into two cosine-similarity paths
normalised to $[0,1]$:
\begin{align}
  \rvlm(o) &= \tfrac{1}{N}\!\sum_{n=1}^{N}
  \tfrac{\mathrm{sim}(f_I(o), F^{+}_n)+1}{2}, &
  \cvlm(o) &= \tfrac{1}{N}\!\sum_{n=1}^{N}
  \tfrac{\mathrm{sim}(f_I(o), F^{-}_n)+1}{2}.
  \label{eq:decoupled}
\end{align}
Text features $F^{\pm}$ are encoded once and cached; per-step cost
is one image encoding plus a small dot product
(Appendix~\ref{app:implementation}).

\textbf{Contribution 2: VLMLagrange (anticipatory multiplier).} Let
$\meancvlm = \tfrac{1}{T}\sum_t \cvlm(o_t)$ and $\tau \in [0,1]$ a
danger threshold. We augment the standard update with a per-step
CLIP-derived anticipatory term:
\begin{equation}
  \boxed{\;
  \lambda \;\leftarrow\; \lambda + \underbrace{\eta_1(\Jc-\dlim)}_{\text{standard (backward)}}
  + \underbrace{\eta_2(\meancvlm-\tau)}_{\text{VLM (forward)}}
  \;}
  \label{eq:vlm-lagrange}
\end{equation}
$\eta_2{=}0$ recovers vanilla PPO-Lagrangian, providing a clean
ablation of the anticipatory contribution. The intuition is direct:
$\cvlm(o_t)$ is elevated as the racecar \emph{approaches} a
barrier, so $\meancvlm$ accumulates pre-collision danger evidence
within an epoch and $\lambda$ rises faster in early training, giving
the constraint a head start in the high-cost exploration phase.
Implementation subclasses OmniSafe's
\texttt{Lagrange} via \texttt{spec\_log}; the PPO loss is unchanged.

\textbf{Contribution 3: Confidence Gating.} CLIP is not uniformly
reliable across visually diverse states. Following standard
logistic-noise treatment of binary
classifiers~\cite{platt1999probabilistic,guo2017calibration}, model
the probability that frame $o_t$ is dangerous as
$\Pr(y_t{=}1\mid m_t) = \sigma(s(m_t-c))$ for CLIP group margin
$m_t \equiv m_t^{+}-m_t^{-}$. The variance-minimizing fusion weight
under the uninformative prior is the Bayes posterior margin:

\begin{equation}
  \kappa_t = \big|\,2\sigma\!\big(s(m_t-c)\big)-1\,\big| \;\in\;[0,1],
  \qquad
  \lambda_r^{\text{eff}} = \kappa_t\lambda_r,
  \quad
  \lambda_c^{\text{eff}} = \kappa_t\lambda_c.
  \label{eq:kappa-bayes}
\end{equation}
Decisive frames ($\kappa_t{\to}1$) pass the signal through; ambiguous
frames ($\kappa_t{\to}0$) suppress it. The hyperparameters $(s,c)$
admit a closed-form maximum-likelihood estimate from a $B$-frame
random-policy buffer $\mathcal{B}$ on the target environment:
\begin{equation}
  \hat{c} = \mathrm{median}(\mathcal{B}),
  \qquad
  \hat{s} = \frac{1}{\mathrm{IQR}(\mathcal{B})}
  \log\!\frac{1+\kappa^{\star}}{1-\kappa^{\star}},
  \label{eq:mle-sc}
\end{equation}
where $\kappa^{\star}$ is a target gate value at $+1$~IQR
(see Appendix~\ref{app:gate-mle-derivation} for the derivation, and
Appendix~\ref{app:gate-calibration} for the prior-symmetric vs.\
calibrated ablation; the L2 categorical conclusion is invariant to
the choice). Empirical margin distribution is concentrated
in the saturated tail, $\kappa^{\star}_t\!\to\!1$ uniformly and the
gate degenerates to identity the failure mode we observe
empirically on MetaDrive Hard (Appendix~\ref{app:gate-calibration}).
A held-out validation of $\kappa$ as a danger predictor against
simulator cost on FormulaOne ($50{,}000$ stochastic frames, $5$ eval
episodes per level) yields calibrated AUC $0.82$ (L1) and $0.78$ (L2) (Appendix~\ref{app:gate-roc}).

\textbf{VLMPPOLag (composition).} VLMPPOLag (Algorithm
\ref{alg:vlmppolag}) inherits from PPOLag and replaces the
\texttt{Lagrange} class with \texttt{VLMLagrange}; the policy loss,
value functions and PPO clipping are unchanged. Only the multiplier
update receives the VLM signal, cleanly separating the contribution
from policy optimization.

\begin{algorithm}[h]
\caption{\textsc{VLMPPOLag} (one epoch).}
\label{alg:vlmppolag}
\small
\begin{algorithmic}[1]
  \STATE \textbf{Init}: \texttt{VLMLagrange}$(\lambda_0,\eta_1,\dlim,\eta_2,\tau)$,
    CLIP$(\mathbf{p}^{+},\mathbf{p}^{-})$
  \STATE Collect rollout; for each step $t$ compute
    $\rvlm(o_t),\cvlm(o_t),\kappa_t$ via CLIP; set
    $\tilde{r}_t \!\leftarrow\! r_{\text{env},t} + \kappa_t\lambda_r\rvlm(o_t)$,
    $\tilde{c}_t \!\leftarrow\! c_{\text{env},t}$
  \STATE Compute GAE advantages; update $\pi_\theta,V_\phi$ with the
    standard PPO clipped loss
  \STATE
    $\lambda \leftarrow \max\!\big(0,\;\lambda+\eta_1(\Jc-\dlim)+\eta_2(\meancvlm-\tau)\big)$
    \hfill (\eqnref{eq:vlm-lagrange})
\end{algorithmic}
\end{algorithm}

\vspace{-1.5ex}

\section{Experimental Setup}
\label{sec:setup}

\textbf{Primary benchmark.} Safety-Gymnasium FormulaOne
v0.5~\cite{ji2023safety} at three obstacle-density levels: \textbf{L0}
(clear track), \textbf{L1} (4 cones at hairpin apexes), \textbf{L2}
(8 staggered concrete barricades). $T{=}1000$ steps at $25$\,Hz,
budget $\dlim{=}25$, $10^{6}$ training steps, $\gamma{=}0.99$,
$2{,}000$ steps/epoch, CLIP ViT-B/32 frozen, $N{=}4$ prompts/polarity
(v1; v2/v3 sensitivity in Appendix~\ref{app:prompts}).

\textbf{Baselines (10 configurations).}
\emph{(i)~Pure RL}: PPO~\cite{schulman2017proximal}.
\emph{(ii)~CMDP, no VLM}: CPO~\cite{achiam2017constrained},
PPOLag~\cite{stooke2020responsive},
CPPOPID~\cite{stooke2020responsive} (PID-Lagrangian, L2 only).
\emph{(iii)~VLM-RL-style (CLG)}: PPO-CLG and CPO-CLG apply the
coupled-softmax CLG scoring from VLM-RL~\cite{huang2024vlmrl} as a
reward bonus; these isolate coupled-vs-decoupled.
\emph{(iv)~Ours and ablations}: CPO-Coupled, CPO-Decoupled,
PPOLag-Decoupled, VLMPPOLag, VLMPPOLag$+$Conf. Hyperparameters:
$\lambda_r{=}0.1$, $\lambda_c{=}0.5$, $\eta_2{=}0.01$, $\tau{=}0.5$
(full sweep in Appendix~\ref{app:hparams}). Three seeds per
(method, level) cell; the principal +Conf row is extended to 5 seeds
$\{42,123,456,789,1024\}$ via Phase~B; statistical reporting uses
the one-sided permutation test (Appendix~\ref{app:perm-tests}) which
floors at $p{=}1/20{=}0.05$ for $n{=}3$ vs.\ $n{=}3$.

\textbf{Held-out protocol.} Generalization policies (Bullet,
MetaDrive) evaluated deterministically for 20 episodes on each of
seeds $\{10000,\ldots,10019\}$ ($n{=}60$ episodes for 3-seed cells;
$n{=}100$ for 5-seed cells). We report mean return $\Jr$, mean cost
$\Jc$, \emph{violation rate} ($\Jc{>}\dlim$) and \emph{catastrophe
rate} ($\Jc{>}4\dlim$) with $5{,}000$-resample bootstrap
CIs~\cite{efron1986bootstrap}.

\textbf{Generalization environments.}
\emph{Bullet}~\cite{SafetyGym2019} \texttt{SafetyCarReach-v0}: 2D
arena with 8 hazard spheres, overhead view; 3 seeds $\times$
\{1M,2M\} steps. \emph{MetaDrive}~\cite{li2023metadrive}
Easy/Medium/Hard: front-facing dashboard camera, ego-centric;
3 seeds (Easy), 5 seeds (Medium, Hard). All MetaDrive runs use
$\texttt{num\_scenarios}{=}10000$ to disable a scenario-sampler
aliasing (default $100$ silently overlaps held-out and training
scenarios for our seed range; quantified in
Appendix~\ref{app:metadrive-bug}).

\section{Results}
\label{sec:results}

\textbf{VLM-SAFE-RL} evaluated primarily on Safety-Gymnasium FormulaOne.
\Cref{tab:main_results} reports final-epoch training-time performance.
The +Conf and CPPOPID rows use the extended 5-seed Phase~B set;
all others use the original 3-seed set, annotated in the table.
Three patterns stand out.

\textbf{(P1) VLM reward shaping is the dominant performance driver.}
Pure-RL PPO produces $\Jr{=}1.6$ on all three levels, the sparse
task reward is too weak. All decoupled-path and CLG methods that use
CLIP reward shaping jump to $\Jr{>}50$. CPO-Coupled is the
exception ($\Jr{=}21.2$ at L0): the coupled simplex suppresses
$\rvlm$ when the cost path is active.

\textbf{(P2) Decoupling the CLIP paths is the single largest
representation gain.} CPO-Coupled ($\Jr{=}21.6$ at L2) vs.\
CPO-Decoupled ($\Jr{=}63.9$) at comparable cost ($\Jc{=}32.4$ vs.\
$30.9$): the coupled simplex forces $\rvlm{+}\cvlm{\approx}1$;
decoupling restores the formal $R\!\perp\!C$ independence the CMDP
assumes. Permutation $p{=}0.05$ on $\Jr$ at L2 (the structural floor
at $n{=}3$, indicating complete rank separation), with full
pairwise tables in Appendix~\ref{app:stats-pairwise}.

\textbf{(P3) Anticipatory $+$ confidence-gated is the only
constraint-respecting configuration with substantive return.}
PPOLag-Decoupled ($\eta_2{=}0$, $\Jc{=}40.7$ at L2) vs.\ VLMPPOLag
($\eta_2{=}0.01$, $\Jc{=}40.2$) shows a small consistent improvement
from the anticipatory term alone; adding confidence gating (5 seeds)
drops $\Jc$ to $22.5$---a $44\%$ reduction---with $4/5$ training
seeds holding cost below budget. Crucially, the reactive baselines
PPOLag, CPO, CPPOPID, PPOLag-RND each collapse to $\Jr{\approx}0$ to
satisfy the constraint (the same collapse holds for three additional
Lagrangian variants (FOCOPS, CUP, P3O) reported in
App.~\ref{app:extra-baselines}), while CPO-CLG retains return ($\Jr{=}50.9$)
but violates the budget ($\Jc{=}33.9{>}25$). VLMPPOLag$+$Conf is
the only configuration that achieves \emph{both} substantive return
\emph{and} within-budget cost on a majority of seeds. The cost
reduction comes at a return cost ($\Jr{:}63.8{\to}31.8$) because
gating attenuates both channels equally, a calibrated
safety-first operating point, not a Pareto improvement.
(Pareto-anchor runs with PPOLag-Decoupled at $\dlim\!\in\!\{15,35\}$
are reported in Appendix~\ref{app:pareto-baseline}).

\begin{table}[h]
\centering
\caption{\textbf{FormulaOne training-time final-epoch performance}
(mean$\pm$std, $10^6$ steps). \textcolor{blue}{Blue}: $\Jc{\le}\dlim{=}25$.
\textbf{Bold}: best per column among VLM-augmented methods. $^{*}$5
seeds $\{42,123,456,789,1024\}$; calibrated gate from
\eqnref{eq:mle-sc} (App.~\ref{app:gate-prereg}). Other rows: 
3 seeds $\{42,123,456\}$. CPPOPID is L2-only. \textit{Type:}
RL=pure RL; CMDP=constrained MDP, no VLM; CLG=VLM-as-reward;
RND=intrinsic-novelty ablation; \textbf{Ours}=CMDP$+$VLM cost.
One-sided permutation test (App.~\ref{app:perm-tests}); full
pairwise table in App.~\ref{app:stats-pairwise}.}
\label{tab:main_results}
\tiny
\setlength{\tabcolsep}{2pt}
\renewcommand{\arraystretch}{0.78}
\rowcolors{3}{green!10}{gray!10}
\begin{tabular}{@{}llcccccc@{}}
\toprule
& & \multicolumn{2}{c}{\textbf{L0}}
& \multicolumn{2}{c}{\textbf{L1}}
& \multicolumn{2}{c}{\textbf{L2}} \\
\cmidrule(lr){3-4}\cmidrule(lr){5-6}\cmidrule(lr){7-8}
\textbf{Type} & \textbf{Method} & $\Jr$ & $\Jc$ & $\Jr$ & $\Jc$ & $\Jr$ & $\Jc$ \\
\midrule
RL   & PPO              & 1.6$\pm$0.5  & \textcolor{blue}{0.0$\pm$0.0} & 1.6$\pm$0.1 & 217.0$\pm$40.2 & 1.3$\pm$0.1 & 269.2$\pm$20.7 \\
CMDP & CPO              & 1.9$\pm$0.6  & \textcolor{blue}{0.0$\pm$0.0} & 0.3$\pm$0.1 & 35.7$\pm$13.2  & 0.3$\pm$0.3 & 36.1$\pm$16.4  \\
CMDP & PPOLag           & 1.6$\pm$0.5  & \textcolor{blue}{0.0$\pm$0.0} & 0.8$\pm$0.5 & 67.9$\pm$49.7  & 0.7$\pm$0.2 & 55.8$\pm$35.7  \\
CMDP & CPPOPID          & ---          & ---                          & ---         & ---            & 0.2$\pm$0.3 & \textcolor{blue}{22.8$\pm$8.2} \\
CLG  & PPO-CLG          & 51.9$\pm$0.4 & \textcolor{blue}{0.0$\pm$0.0} & 51.7$\pm$0.2 & 133.6$\pm$37.3 & 51.3$\pm$0.1 & 156.6$\pm$40.4 \\
CLG  & CPO-CLG          & 51.7$\pm$0.3 & \textcolor{blue}{0.0$\pm$0.0} & 50.7$\pm$0.3 & 32.8$\pm$8.4   & 50.9$\pm$0.1 & 33.9$\pm$5.8   \\
RND  & PPOLag-RND       & ---          & ---                          & 0.8$\pm$0.5 & 62.4$\pm$22.8  & 0.4$\pm$0.4 & 45.7$\pm$17.1  \\
\textbf{Ours} & CPO-Coupled      & 21.2$\pm$0.5 & \textcolor{blue}{0.0$\pm$0.0} & 21.0$\pm$0.7 & 29.1$\pm$4.1  & 21.6$\pm$0.2 & 32.4$\pm$4.4  \\
\textbf{Ours} & CPO-Decoupled    & 64.1$\pm$0.3 & \textcolor{blue}{0.0$\pm$0.0} & 63.7$\pm$0.2 & 37.6$\pm$19.4 & 63.9$\pm$0.2 & 30.9$\pm$11.0 \\
\textbf{Ours} & PPOLag-Decoupled & 64.3$\pm$0.4 & \textcolor{blue}{0.0$\pm$0.0} & 64.0$\pm$0.1 & 33.5$\pm$1.7  & 63.8$\pm$0.3 & 40.7$\pm$6.1  \\
\textbf{Ours} & VLMPPOLag        & \textbf{64.3$\pm$0.4} & \textcolor{blue}{\textbf{0.0$\pm$0.0}} & \textbf{64.1$\pm$0.2} & 32.8$\pm$8.2 & \textbf{63.8$\pm$0.2} & 40.2$\pm$12.6 \\
\textbf{Ours} & VLMPPOLag$+$Conf$^{*}$ & 44.5$\pm$9.6 & \textcolor{blue}{0.0$\pm$0.0} & 33.5$\pm$6.7 & \textcolor{blue}{\textbf{20.8$\pm$14.6}} & 31.8$\pm$12.2 & \textcolor{blue}{\textbf{22.5$\pm$5.9}} \\
\bottomrule
\end{tabular}
\end{table}

\subsection{Anticipatory dynamics, ablation, and where to inject the VLM}
\label{sec:results-aux}

\Cref{fig:lambda-dynamics} traces $\lambda$ during L2 training:
VLMPPOLag (red) rises noticeably faster than PPOLag-Decoupled
(purple, $\eta_2{=}0$, identical otherwise). Both algorithms
receive the same environment cost stream; the only structural
difference is the anticipatory $\eta_2(\meancvlm-\tau)$ term, so the
divergence directly evidences the design intent of
\eqnref{eq:vlm-lagrange}. VLMPPOLag$+$Conf (green) settles at a
substantially lower equilibrium because gating attenuates
$\meancvlm$ on visually ambiguous frames. \Cref{fig:learning-curves}
shows that this anticipatory benefit \emph{grows} with obstacle
density: VLMPPOLag and +Conf maintain $\Jr$ across L0--L2 while pure
PPO suffers catastrophic cost at L1/L2 (Spearman $\rho{=}0.19$,
$p{<}10^{-8}$ for per-bin $\cvlm$ vs.\ collision probability;
App.~\ref{app:cr-extras}).

\Cref{tab:ablation} decomposes the L2 result. Removing
decoupled CLIP drops $\Jr$ by $66\%$ at unchanged $\Jc$
(representation dominates); removing the anticipatory term changes
violation count from $2/3$ to $3/3$ (forward-looking $\lambda$
prevents uniform cost saturation); removing confidence gating brings
$\Jc$ from $30.5$ to $40.2$ ($\kappa$ removes spurious VLM spikes).
A complementary \emph{injection-mode} ablation (4 modes, 3 base
algorithms, L1$+$L2; Appendix~\ref{app:vlm-mode}) yields a robust
ranking: $\textit{Decoupled+Conf} \;>\; \textit{Decoupled} \;\gg\;
\textit{Coupled} \;\approx\; \textit{VLG}$.
The most common prior-work design (adding $\cvlm$ to the environment
cost; \emph{Coupled},
\textit{cf.}~\cite{huang2024vlmrl,rocamonde2024vlmrm}) is
\emph{worse} than not
using a VLM on CPO~L2 ($20\%$ vs.\ $13\%$ catastrophe), and routing
the VLM cost directly into the $\lambda$ update without gating
(\emph{VLG}) amplifies catastrophe to $47\%$ on PPO-L1: noisy VLM
spikes inflate $\lambda$ before the critic can integrate them
temporally. This is precisely why our default routes through gated
\emph{Decoupled+Conf}.

\noindent
\begin{minipage}[t]{0.46\linewidth}
  \vspace{0pt} 
  \centering
  
  \captionof{table}{\textbf{Ablation on FormulaOne L2} (3 seeds $\{42,123,456\}$). Each row removes one contribution from the full VLMPPOLag$+$Conf system. Phase~B 5-seed extension of the +Conf row in \Cref{tab:main_results}: $\mathcal{J}_R{=}31.8$, $\mathcal{J}_C{=}22.5$, $1/5$ violations.}
  \label{tab:ablation}
  \scriptsize
  \setlength{\tabcolsep}{4pt}
  \renewcommand{\arraystretch}{1.0}
  \begin{tabular}{@{}lccc@{}} 
    \toprule 
    \textbf{Configuration} & $\mathcal{J}_R$ & $\mathcal{J}_C$ & \textbf{Viol.} \\ 
    \midrule 
    VLMPPOLag$+$Conf (full) & \textbf{48.4} & \textbf{30.5} & \textbf{1/3} \\ 
    \quad w/o Confidence Gating & 63.8 & 40.2 & 2/3 \\ 
    \quad w/o VLMLagrange ($\eta_2{=}0$) & 63.8 & 40.7 & 3/3 \\ 
    \quad w/o Decoupled (Coupled CLIP) & 21.6 & 32.4 & 3/3 \\ 
    \quad w/o VLM (baseline PPOLag) & 0.7 & 55.8 & 2/3 \\ 
    \bottomrule 
  \end{tabular}
  
  \vspace{12pt} 
  
  \captionof{table}{\textbf{Comparison of VLM-augmented safe-learning approaches.} Ours is the only design that simultaneously (i)~operates inside the CMDP, (ii)~decouples the CLIP paths, (iii)~uses an \emph{anticipatory} Lagrangian update, and (iv)~gates each frame by CLIP confidence. $^{\dagger}$CPO-CLG re-implements the VLM-RL CLG scoring on CPO.}
  \label{tab:comparison}
  \scriptsize
  \setlength{\tabcolsep}{2.5pt}
  \renewcommand{\arraystretch}{0.95}
  \begin{tabular}{@{}lcccc@{}} 
    \toprule 
    & \textbf{SafeVLA} & \textbf{VLM-RL} & \textbf{CPO-CLG}$^{\dagger}$ & \textbf{Ours} \\ 
    \midrule 
    Safety form. & CMDP & none & CMDP & CMDP \\ 
    CLIP scoring & N/A & coupled & coupled & \textbf{decoupled} \\ 
    VLM role & policy & reward & reward & rew.$+$cost \\ 
    Lagrangian & yes & no & no & \textbf{anticip.} \\ 
    Conf.\ gating & no & no & no & \textbf{yes} \\ 
    Frozen & no & yes & yes & yes \\ 
    Held-out eval. & yes & no & no & \textbf{yes} \\ 
    \bottomrule 
  \end{tabular}
\end{minipage}
\hfill 
\begin{minipage}[t]{0.5\linewidth}
  \vspace{0pt}
  \centering
  
    \captionof{figure}{\small \textbf{Lagrange multiplier dynamics.} Multiplier $\lambda$ tracking on FormulaOne-L2.}
  \includegraphics[width=\linewidth]{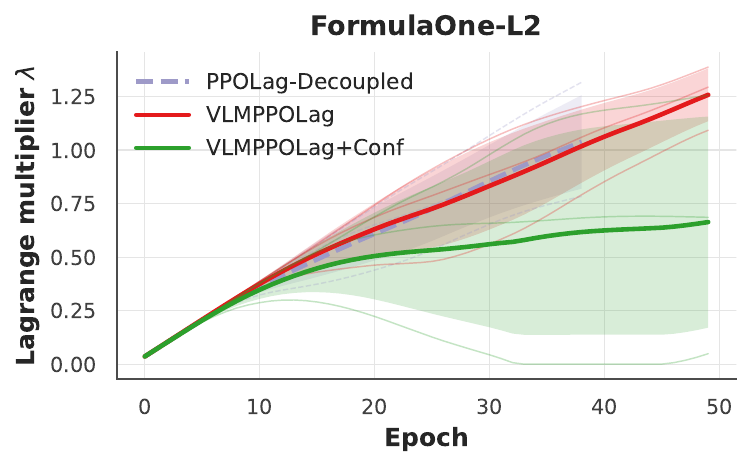}
  \label{fig:lambda-dynamics}
  
\vspace{-3pt}
  
  \justifying   
  {\small\noindent\textbf{Interpretation of Figure~\ref{fig:learning-curves}:} PPOLag-Decoupled isolates 
  $\eta_2(\bar{c}_{vlm}{-}\tau)$; the same mechanism cuts catastrophe $41\%{\to}26\%$ 
    on MetaDrive Medium (\Cref{tab:generalisation}). Faded traces are per-seed runs; bold lines are seed-mean; columns are
    L0$\to$L1$\to$L2 (increasing obstacle density), top row $\Jr$
    and bottom row $\Jc$ (dashed line marks the budget
    $\dlim{=}25$). PPO (grey) $\Jc$ scales
    with obstacle count ($0{\to}217{\to}269$) while VLMPPOLag (red)
    and +Conf (green) keep $\Jr$ near the decoupled-CLIP ceiling;
    +Conf is the only trace whose $\Jc$ curve settles \emph{below}
    the budget line at L1 and L2, and does so from early training
    rather than after a violation spike, evidencing the anticipatory mechanism's predicated advantage
    signature of the anticipatory $\lambda$ rise in
    panel~\ref{fig:lambda-dynamics}}
\end{minipage}

\vspace{2.5pt} 

\noindent
\begin{minipage}{\linewidth}
  \centering
    \captionof{figure}{\small \textbf{Training curves on FormulaOne-L0/L1/L2:} episode return
      $J_R$ (top row) and episode cost $J_C$ (bottom row); faded
      per-seed traces with bold seed-mean overlays.}
  \includegraphics[width=0.95\linewidth]{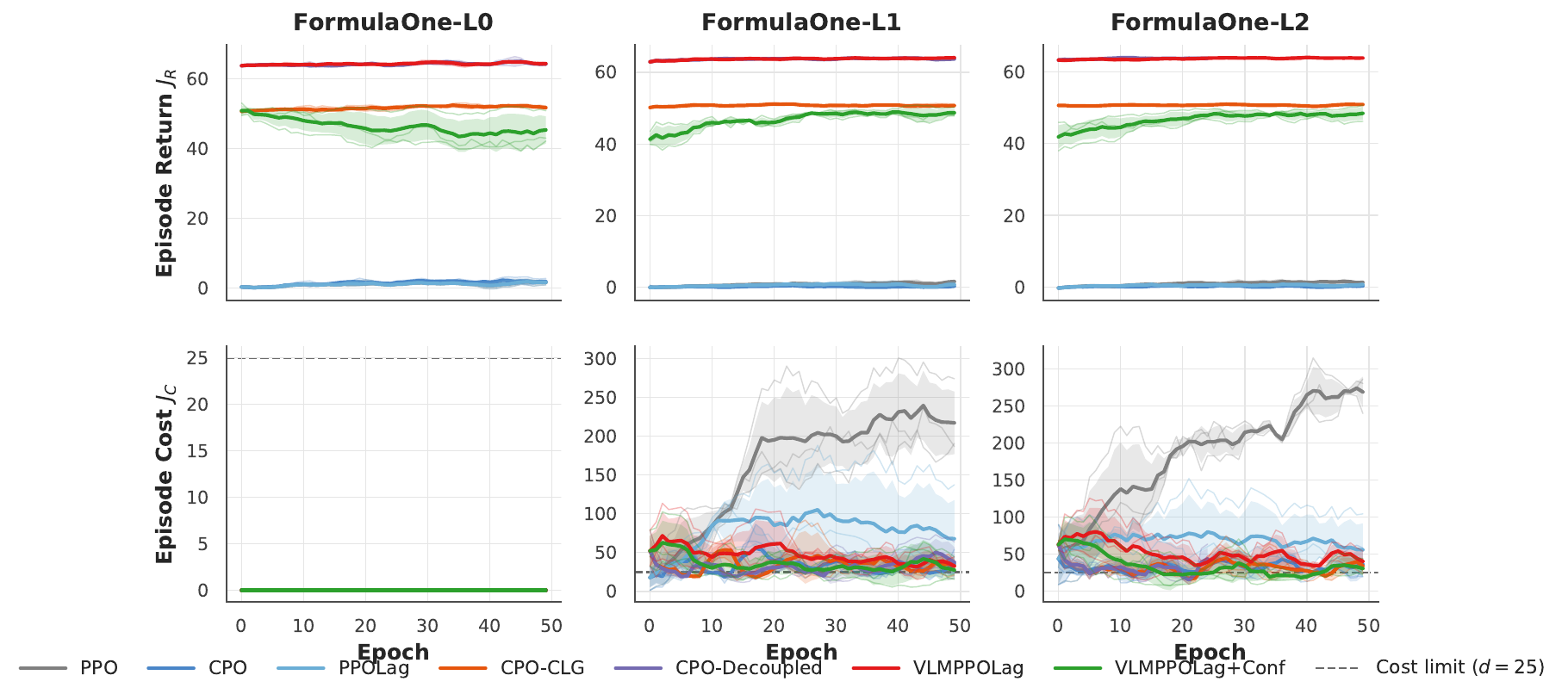}
  \label{fig:learning-curves}
\end{minipage}

\vspace{-2em}
\noindent
\subsection{Negative-result probes: what does \emph{not} improve safety?}
\label{sec:results-negative}

Two controlled substitutions stress-test which property of the VLM
signal carries the safety benefit; full numbers and statistics in
Appendix~\ref{app:cr-extras}. \textbf{(N1) Semantic grounding $\neq$
generic intrinsic cost (RND).} Replacing $\cvlm$ with Random Network
Distillation novelty~\cite{burda2018rnd} under matched hyperparameters
yields $\Jr{\approx}0$, $3/3$ violations on L1$+$L2: the novelty
predictor memorises the near-deterministic proprioceptive manifold,
collapsing the auxiliary signal. \textbf{(N2) Backbone capacity is
not the bottleneck (Qwen2-VL-7B).} Swapping CLIP for Qwen2-VL-7B
($\sim\!80\times$ larger frozen VLM) at matched
hyperparameters yields a clean cost null ($\Jc{=}45.5$ vs.\ $30.5$,
permutation $p{=}0.80$) and significantly worse return
($\Jr{=}8.2$ vs.\ $48.4$, $p{=}5{\times}10^{-4}$); the
yes/no logsumexp margin is systematically more conservative than
CLIP's 8-prompt softmax margin, attenuating the reward channel
aggressively. (N1)+(N2) together identify the operative ingredient
as \emph{visual semantics of impending danger}, not parameter count
or generic novelty.

\subsection{Generalization: held-out evaluation}
\label{sec:results-gen}

\Cref{tab:generalisation} reports held-out catastrophe and violation
rates from 20 deterministic episodes per training run on seeds
$10000$--$10019$, after the MetaDrive seed-leak fix
(Appendix~\ref{app:metadrive-bug}).

\textbf{FormulaOne L2} (5-seed VLMPPOLag+Conf, 2-seed PPOLag held-out):
VLMPPOLag+Conf achieves cat~$8\%$, viol~$18\%$; PPOLag achieves
cat~$2\%$, viol~$18\%$ but with near-zero forward progress at
held-out maps (mean $\Jr\!\approx\!0.10$ vs.\ training-time
$\Jr\!\approx\!40$), confirming the degenerate-safety mode documented
in App.~\ref{app:phaseB-robustness}. The cat difference of $+6$\,pp
reflects this mode rather than a genuine safety improvement; the
$95\%$ CI is $[-2,+13]$\,pp (non-significant, $n{=}2$ baseline seeds).

\textbf{Bullet SafetyCarReach} $7.5\%$ vs.\ $5.0\%$ catastrophe
($16.7\%$ vs.\ $12.5\%$ violation; $n{=}120$ episodes per method).
The directional improvement is consistent with FormulaOne and
MetaDrive Medium, but the bootstrap CI on the catastrophe-rate
difference $[-9.2,+3.3]$\,pp does \emph{not} exclude zero;
characterised as directionally consistent but not statistically
detectable at this sample size.

\textbf{MetaDrive Easy} baseline $30\%$, VLMPPOLag$+$Conf $35\%$
($+5$\,pp). Sparse traffic on wide roads; the front-facing camera
rarely captures an imminent collision until impact, so CLIP has low
temporal advance.

\textbf{MetaDrive Medium} (5 seeds, 100 held-out episodes per method):
$41\%{\to}26\%$ catastrophe, $51\%{\to}35\%$ violation; bootstrap
$95\%$ CI $[-26,-5]$\,pp, entirely below zero. The strongest
generalization signal in our experiments: merging traffic gives
several timesteps of advance warning, exercising the anticipatory
mechanism directly.

\textbf{MetaDrive Hard} (5 seeds) $33\%{\to}31\%$ catastrophe; the
violation rate is marginally \emph{higher} for VLM
($39\%$ vs.\ $36\%$). Per-seed inspection localises this to a
\emph{Lagrangian-regulation} failure rather than a VLM-signal failure
(\Cref{tab:md-hard-lambda} in App.~\ref{app:md-perseed}): VLMCost
mean is uniform across all 5 seeds ($\meancvlm{\approx}0.60$) but
the final $\lambda$ ranges from $0.10$ to $0.93$. Seeds whose
first-epoch cost realizations happen to be atypically low leave
$\lambda$ unable to grow within $50$ epochs (seed~$456$:
$\lambda{=}0.10$, $55\%$ catastrophe); atypically high realizations
overshoot the attractor (seed~$789$: $\lambda{=}0.93$, return
collapse). The remaining three seeds converge near
$\lambda{\approx}0.5{-}0.7$. A controlled $\lambda_0{=}0.5$ warm-start
re-run of all 5 seeds (Appendix~\ref{app:md-perseed}) 
pooled catastrophe drops $31\%{\to}25\%$ and pooled violation
$39\%{\to}28\%$, $39\%{\to}28\%$---directionally consistent but not statistically
detectable at $n{=}5$ (bootstrap CI $[-26,+11]$\,pp); seed~$456$
remains catastrophic and early-cost variance still dominates.
The updated row appears in \Cref{tab:generalisation}.

\begin{table}[h]
\centering
\caption{\textbf{Out-of-distribution generalisation transfer.}
Held-out evaluation on seeds $10000$--$10019$ (20 episodes per
training run). \textbf{Cat.}: $\Jc{>}4\dlim$ rate.
\textbf{Viol.}: $\Jc{>}\dlim$ rate. \textbf{Cat.~$\Delta$ 95\% CI}:
$5{,}000$-resample bootstrap CI of (VLM$+$Conf)$-$PPOLag in pp
(F1-L2: Wald CI, $n{=}2$ PPOLag seeds).
$\dagger$: PPOLag cat~$2\%$ reflects near-zero forward progress
at held-out maps (degenerate safety); see App.~\ref{app:phaseB-robustness}.
Per-seed $\lambda$ diagnostic for MD~Hard:
\Cref{tab:md-hard-lambda}.}
\label{tab:generalisation}
\scriptsize
\setlength{\tabcolsep}{3pt}
\renewcommand{\arraystretch}{0.85}
\rowcolors{3}{green!10}{gray!10}
\begin{tabular}{@{}lccccccc@{}}
\toprule
\textbf{Environment} &
  \multicolumn{2}{c}{\textbf{Cat.\,\%}} &
  \multicolumn{2}{c}{\textbf{Viol.\,\%}} &
  \textbf{Cat.~$\Delta$ 95\% CI} & \textbf{Sig.?} \\
\cmidrule(lr){2-3}\cmidrule(lr){4-5}
 & PPOLag & VLM$+$Conf & PPOLag & VLM$+$Conf & (pp) & \\
\midrule
F1-L2$^\dagger$  & 2  & 8  & 18 & 18 & $[-2,+13]$    & no (2 seeds) \\
Bullet Car-Reach & 7.5 & 5.0 & 16.7 & 12.5 & $[-9.2,+3.3]$ & no \\
MetaDrive Easy   & 30 & 35 & 43 & 38 & overlaps zero & no (3 seeds) \\
MetaDrive Medium & 41 & 26 & 51 & 35 & $[-26,-5]$ & \textbf{yes} \\
MetaDrive Hard   & 33 & 31 & 36 & 39 & overlaps zero & no \\
MD Hard ($\lambda_0{=}0.5$) & 33 & 25 & 36 & 28 & $[-26,+11]$ & no \\
\bottomrule
\end{tabular}
\end{table}

\section{Discussion, Limitations, and Conclusion}
\label{sec:discussion}

\textbf{Why the multiplier rises faster.} VLMLagrange converts
Lagrangian safe RL from a reactive mechanism (respond after
$\Jc{>}\dlim$) to an anticipatory one. The defining evidence is the
early-training $\lambda$ rise of \Cref{fig:lambda-dynamics}: both
VLMPPOLag and PPOLag-Decoupled receive the identical environment
cost stream, so the divergence is attributable solely to
$\eta_2(\meancvlm-\tau)$. $\meancvlm$ accumulates pre-collision
danger evidence within the first thousand training steps before
the policy has learned to avoid barriers, giving $\lambda$ a head
start that the ablation (\Cref{tab:ablation}) translates into one
fewer violating seed at L2.

\textbf{When the mechanism generalises.} The combined picture is,
anticipatory benefit is largest when the environment
provides \emph{temporal advance warning}: several frames where the
danger is visually present and CLIP is decisive
($\kappa{\to}1$). MetaDrive Medium (merging vehicles) satisfies this
cleanly. Easy does not (sparse traffic, head-on collisions only at
contact). Hard fails for two compounding reasons: (a) the
saturated-tail prediction of \S\ref{sec:method} (median $\kappa{=}0.93$--$0.98$
on Hard, App.~\ref{app:gate-calibration}) makes the gate
non-selective, and (b) the Lagrangian-regulation pathology of
\Cref{tab:md-hard-lambda} above. Both are addressable in principle:
the gate-saturation failure by environment-specific prompt design
or a frame-history buffer; the $\lambda$ pathology by
warm-initialising $\lambda_0$ at the empirical attractor (tested
directly by the $\lambda_0{=}0.5$ warm-start re-run reported in
\Cref{tab:generalisation} and App.~\ref{app:md-perseed}). On Bullet, the
direction is right but the baseline is already low-catastrophe,
leaving little headroom and a CI that does not exclude zero.

\textbf{Path to physical deployment.} Frozen VLM at training time
means an identical inference-time pipeline. ViT-B/32 forward pass is
$7.11$\,ms (A100) / $9.11$\,ms (V100), $\sim\!95\%$ headroom against
a $25$\,Hz control loop (App.~\ref{app:compute}). The 
MuJoCo$\to$Bullet/PyTorch3D shift in our generalisation results shows
the frozen VLM head tolerates the pipeline change that historically
defeats visually-conditioned policies; prompt distribution shift and
worst-case CLIP latency remain open, addressable by per-deployment
recalibration and a last-$\cvlm$ fallback on overrun. 
  
\textbf{Reproducibility lesson.} MetaDrive's default scenario sampler
silently aliases held-out seeds onto training scenarios via $\texttt{seed}\bmod\texttt{num\_scenarios}$, producing flat
$\approx\!40\%$ catastrophe rates with no separation between methods.
The fix is one line ($\texttt{num\_scenarios}{=}10000$); collapsed
and corrected numbers side-by-side in App.~\ref{app:metadrive-bug}.

\textbf{Limitations.} (1) Prompt engineering is manual; learning
prompts from cost signal~\cite{ma2024eureka} is a natural extension.
(2) Temporal advance warning is required: in fast-onset/highly-occluded
regimes (MD~Hard) the per-epoch $\lambda$ update cannot respond on the
cut-in timescale and the gate degenerates to identity; warm-starting $\lambda_0$ gives a directionally consistent improvement
(Apps.~\ref{app:gate-calibration}, \ref{app:md-perseed}) that is not
statistically detectable at $n{=}5$ and two seeds remain catastrophic.
(3) Seed counts: $n{=}5$ for +Conf, $n{=}3$ for generalization; full
CIs in App.~\ref{app:stats}. (4) VLM inference adds $\sim\!15\%$
wall-clock (App.~\ref{app:compute}). (5) Our frozen-VLM operating point
is $\sim\!87\times$ smaller than SafeVLA~\cite{zhang2025safevla} but
does not match its fine-tuned ceiling; the intermediate Qwen2-VL probe
is negative on safety and significantly negative on return. (6) Three
regimes show no benefit (MD~Easy, MD~Hard, $\eta_2{>}0.05$);
App.~\ref{app:sensitivity} brackets the operating envelope. 
(7) Single-multiplier formulation, we fold the environment and
VLM cost signals into one shared $\lambda$ rather than the standard
two-multiplier $(\lambda_1,\lambda_2)$ CMDP
treatment~\cite{altman1999constrained}; a
$\lambda_1(\Jc{-}\dlim)+\lambda_2(\meancvlm{-}\tau)$ variant is left
as an open ablation that may resolve the MD~Hard regulation
pathology (\Cref{tab:md-hard-lambda}).

\textbf{Conclusion.} \textbf{VLM-Safe-RL}, decoupled dual-path
representation, anticipatory VLMLagrange, confidence gating, and 
VLMPPOLag converts Lagrangian safe RL from reactive to
anticipatory. VLMPPOLag$+$Conf retains substantive return and
holds cost on a majority of FormulaOne seeds and transfers to
MetaDrive Medium ($-15$\,pp catastrophe, CI excludes zero):a practical, sim-to-real-friendly complement to fine-tuned VLA.

\newpage
\bibliography{references}

\clearpage
\appendix

\appendix
\section{Appendix}
\label{app:implementation}

\subsection{Hyperparameters}
\label{app:hparams}

\Cref{tab:hparams} provides a complete specification of the hyperparameters
shared across all environments and methods. All algorithms use identical
optimisation, PPO/CPO, and Lagrange settings; only the VLM-specific block
varies between with-VLM and without-VLM ablations. Confidence-gating uses the
same VLM weights, with the additional binary group margin (\eqref{eq:kappa-bayes}
in the main paper).

\begin{table}[h]
\centering
\small
\caption{Hyperparameters shared across all runs (FormulaOne, Bullet, MetaDrive).
Algorithm-specific variants only modify the VLM block; the optimisation and
CMDP blocks are held fixed for all baselines and ablations to isolate
algorithmic differences. $^\dagger$The MetaDrive Hard warm-start diagnostic
uses $\lambda_0{=}0.5$; all other runs use the value shown
(see Appendix~\ref{app:md-perseed}).}
\label{tab:hparams}
\begin{tabular}{@{}lll@{}}
\toprule
\textbf{Group} & \textbf{Hyperparameter} & \textbf{Value} \\
\midrule
\multirow{6}{*}{Optimisation}
   & Total timesteps                          & $1\,000\,000$ (FormulaOne, MetaDrive); $1$/$2{\times}10^{6}$ (Bullet) \\
   & Steps per epoch                          & $2\,000$ \\
   & Discount $\gamma$                        & $0.99$ \\
   & GAE $\lambda_{\text{GAE}}$               & $0.95$ \\
   & Learning rate (actor / critic)           & $3\times 10^{-4}$ / $1\times 10^{-3}$ \\
   & Mini-batch size / epochs                 & $64$ / $40$ \\
\midrule
\multirow{4}{*}{CMDP}
   & Cost limit $\dlim$                       & $25$ \\
   & PPO clip ratio $\epsilon$                & $0.2$ \\
   & Lagrange LR $\eta_1$                     & $0.035$ \\
   & Initial multiplier $\lambda_0^{\dagger}$  & $0.001$ \\
\midrule
\multirow{6}{*}{VLM}
   & CLIP backbone                            & ViT-B/32 (frozen, 150M parameters) \\
   & VLM reward weight $\lambda_r$            & $0.1$ \\
   & VLM cost weight $\lambda_c$              & $0.5$ \\
   & VLM Lagrange LR $\eta_2$ (Eq.~(\ref{eq:vlm-lagrange})) & $0.01$ \\
   & Danger threshold $\tau$                  & $0.5$ \\
   & Prompts $K, L$                           & $4, 4$ \\
\midrule
\multirow{3}{*}{Eval}
   & Held-out seeds                           & $10000$--$10019$ (all environments) \\
   & Episodes per run                         & $20$ (deterministic) \\
   & Bootstrap resamples (95\% CIs)           & $2000$ \\
\bottomrule
\end{tabular}
\end{table}

\subsection{Network architecture}
\label{app:arch}

Both actor and critic networks are 2-layer MLPs with hidden dimension
$256$ and Tanh activations. The actor outputs a Gaussian over the
continuous action space (steering, throttle / acceleration) parameterised
by mean $\mu_\theta(s)$ and a state-independent learned log-standard
deviation. Two separate value heads $V_\phi^R(s)$ and $V_\phi^C(s)$
estimate the reward and cost returns and are trained with
GAE-$\lambda$ advantages. The CLIP ViT-B/32 vision tower remains frozen
throughout training; we cache the $K{+}L$ text embeddings at
initialisation, so each control step costs one image encoding plus
$O(K{+}L)$ dot products. No parameters of the CLIP encoder are updated.

\subsection{OmniSafe integration}
\label{app:omnisafe}

\texttt{VLMPPOLag} is 
registered as a first-class OmniSafe v0.5
algorithm by subclassing \texttt{PPOLag} and replacing its
\texttt{Lagrange} component with \texttt{VLMLagrange}, which overrides
\texttt{update\_lagrange\_multiplier(Jc, mean\_vlm\_cost)} to apply the
augmented update of \eqref{eq:vlm-lagrange} in the main paper.
The per-step VLM cost $\cvlm(o_t)$ is communicated from the environment
wrapper to the algorithm through OmniSafe's \texttt{spec\_log} mechanism;
no modifications to the policy loss, value loss, or PPO clipping are
required, which keeps the contribution surgically localised to the
multiplier update and clean to ablate against the corresponding
non-anticipatory baselines (\texttt{PPOLag-Decoupled}, $\eta_2{=}0$).

\subsection{Computational requirements}
\label{app:compute}

Each FormulaOne run uses a single NVIDIA A100 (40\,GB) and takes
$\approx 22$\,h for $10^{6}$ environment steps; the dominant cost is
\texttt{env.render()} for CLIP input plus a single CLIP image forward
pass per control step, adding $\sim\!15\%$ wall-clock overhead vs.\ the
no-VLM baseline. Total compute for the paper is approximately
$660$ GPU-hours across the FormulaOne, Bullet, and MetaDrive cells
(all training seeds, all baselines, all ablations, and the held-out
deterministic evaluations).

\begin{table}[h]
\centering
\small
\caption{Approximate per-run wall-clock and memory across environments.
``Steps/sec'' is measured at the simulation level (rollout collection only).}
\label{tab:compute}
\begin{tabular}{@{}lcccc@{}}
\toprule
\textbf{Method} & \textbf{Time (h)} & \textbf{GPU mem (GB)} & \textbf{Steps/sec} & \textbf{Overhead} \\
\midrule
PPOLag (FormulaOne)        & 19.0 & 2.2 & 14.5 & --- \\
PPOLag-Decoupled           & 21.5 & 4.5 & 12.9 & $+13\%$ \\
VLMPPOLag                  & 22.0 & 4.6 & 12.6 & $+16\%$ \\
VLMPPOLag+Conf             & 22.5 & 4.7 & 12.3 & $+18\%$ \\
\midrule
PPOLag (Bullet 1M)         & 8.0  & 2.1 & 35   & --- \\
VLMPPOLag+Conf (Bullet 1M) & 9.5  & 4.6 & 30   & $+19\%$ \\
\midrule
PPOLag (MetaDrive Med.)    & 14.0 & 2.4 & 20   & --- \\
VLMPPOLag+Conf (MD Med.)   & 16.5 & 4.7 & 17   & $+18\%$ \\
\bottomrule
\end{tabular}
\end{table}

\textbf{Scalability.} The per-step CLIP forward pass is the dominant
overhead; setting \texttt{clip\_inference\_frequency$>$1} (e.g. once
every $k{=}4$ control steps with the last $\cvlm$ value held constant
in between) reduces overhead to $<\!5\%$ at the cost of slightly
noisier VLM signals. We use $k{=}1$ throughout the paper for cleanest
attribution.

\textbf{SLURM allocation.} 1 A100, 32\,GB RAM, 4 CPU cores, 4-day
time limit per job; actual completion 6--22\,h depending on environment
and method.

\section{Environments}
\label{app:envs}

\subsection{Safety-Gymnasium FormulaOne (L0/L1/L2)}
\label{app:env-f1}

The Racecar agent observes a 64-dimensional proprioceptive state
(pose, linear and angular velocities, LiDAR-style range readings) and
acts in a 2-D continuous space (steering, throttle), with episodes
of $T{=}1000$ steps at $25$\,Hz ($40$\,s, MuJoCo $0.004$\,s
integrator timestep with frame-skip $10$). The environment cost is the binary
contact indicator $c_{\text{env},t}{=}1$ on barrier or cone contact.
We evaluate three difficulty levels:
\textbf{L0} (open track, no obstacles, sanity check),
\textbf{L1} (cones / tyre stacks scattered along the track edges),
and \textbf{L2} (large barrel barricades placed inside the racing
line, requiring \emph{anticipatory} avoidance because the turning
radius cannot correct on contact).

\begin{figure}[h]
\centering
\includegraphics[width=0.95\textwidth]{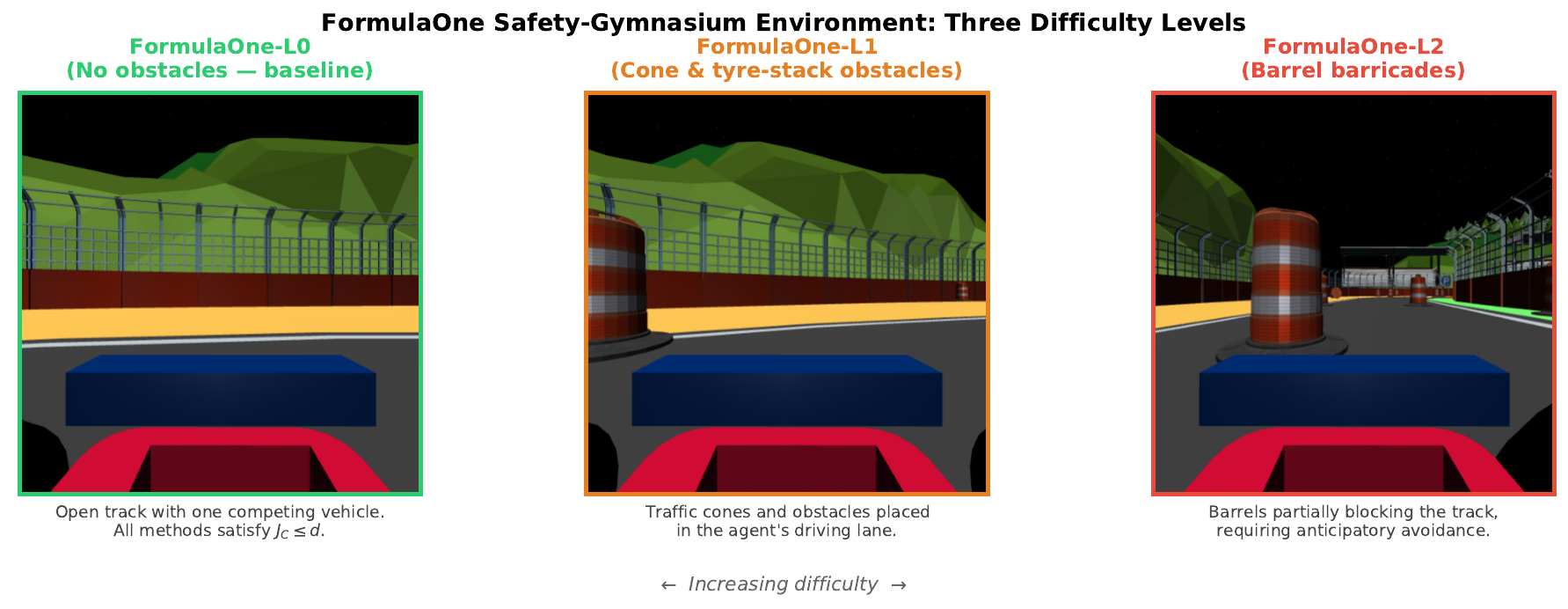}
\caption{First-person camera frames from the three FormulaOne difficulty
levels in Safety-Gymnasium~\cite{ji2023safety} at simulation step~50:
\textbf{L0} open track, \textbf{L1} cones/tyre stacks, \textbf{L2}
barrel barricades inside the driving line.}
\label{fig:appendix-env-levels}
\end{figure}

\subsection{Bullet Safety-Gym: \texttt{SafetyCarReach-v0}}
\label{app:env-bullet}

We use \texttt{SafetyCarReach-v0} from Bullet Safety-Gym~\cite{SafetyGym2019}
(built on PyBullet~\cite{coumans2016pybullet}), in which a car-like agent must
reach a target while avoiding hazards. Episodes are $500$ steps; the cost
function is $1$ on hazard contact and we apply the same cost limit $d{=}25$
as for FormulaOne. The visual observation is a third-person rendered RGB
frame (resized to $224{\times}224$ for CLIP). We train at two horizons,
$1\!\times\!10^{6}$ and $2\!\times\!10^{6}$ environment steps, and report
held-out evaluation across both---this gives $6$ runs per method
(3 training seeds $\times$ 2 horizons), with held-out evaluation on
seeds $10000$--$10019$ ($20$ deterministic episodes per run).

\subsection{MetaDrive (Easy / Medium / Hard)}
\label{app:env-md}

MetaDrive~\cite{li2023metadrive} is a procedural traffic simulator
with realistic kinematics, partial observability, and dense traffic
flow. The observation is a $224{\times}224$ first-person RGB frame
(top-down inset) plus state features; the action is continuous
$[\text{steering},\,\text{acceleration}]$. We use three procedurally
generated maps---\textbf{Easy}, \textbf{Medium}, and \textbf{Hard}---
corresponding to increasing traffic density and to the presence of
roundabouts/intersections in the Hard map. We train for
$1\!\times\!10^{6}$ steps and evaluate on held-out seeds
$10000$--$10019$ (after applying the seed-leak fix described below).
Training seeds: $\{42,123,456\}$ for Easy, $\{42,123,456,789,2024\}$ for
Medium and Hard.

\paragraph{The MetaDrive seed-leak bug.}
\label{app:metadrive-bug}
MetaDrive samples scenarios via
$\texttt{scenario\_id}=\texttt{seed}\bmod\texttt{num\_scenarios}$.
The default $\texttt{num\_scenarios}{=}100$ collapses our held-out seeds
$10000{-}10019$ onto scenarios $0{-}19$, which lie \emph{inside the
training set}. The symptom in our initial run was a flat
$\sim\!40\%$ catastrophe rate for both VLM and baseline with no
separation between methods---a result that, if accepted at face
value, would lead one to conclude ``VLMs add nothing in traffic
domains.'' Setting $\texttt{num\_scenarios}{=}10000$ ensures held-out
seeds map to genuinely unseen scenarios. \emph{All MetaDrive numbers
in the main paper use the corrected configuration.}

\begin{figure}[h]
  \centering
  \includegraphics[width=0.92\linewidth]{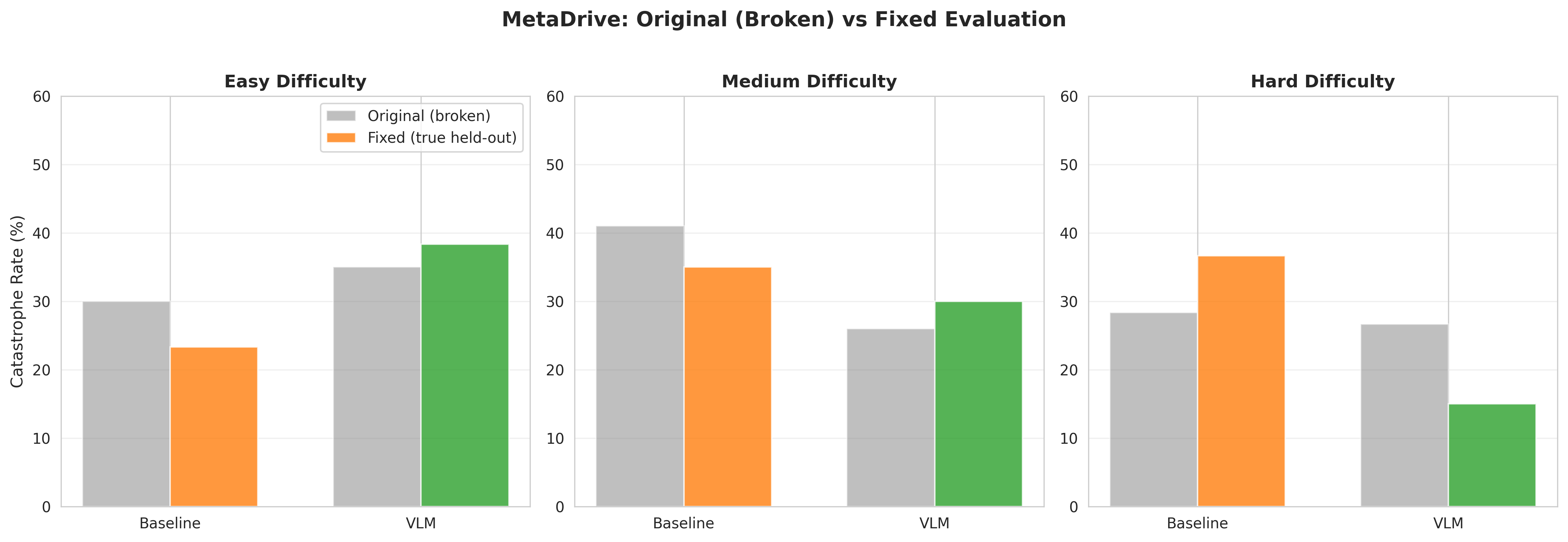}
  \caption{MetaDrive held-out catastrophe rates before (left, default
    \texttt{num\_scenarios=100}) and after (right, corrected
    \texttt{num\_scenarios=10000}) the seed-leak fix. With the
    default, PPOLag and VLMPPOLag+Conf are statistically
    indistinguishable; the fix restores the expected safety gap.}
  \label{fig:metadrive-bug}
\end{figure}

\section{VLM Prompt Engineering}
\label{app:prompts}

\subsection{Full prompt list (v1, used for all main results)}
\label{app:prompt-list}

\textbf{Positive prompts ($K{=}4$, reward shaping):}
\begin{enumerate}
  \item ``the racecar is centred on the track and driving safely''
  \item ``the racecar is following the racing line smoothly''
  \item ``the racecar is making steady forward progress''
  \item ``the car is driving efficiently without collisions''
\end{enumerate}

\textbf{Negative prompts ($L{=}4$, cost shaping):}
\begin{enumerate}
  \item ``the racecar is about to crash into the barrier''
  \item ``the racecar is off the track and unsafe''
  \item ``the car is colliding with an obstacle''
  \item ``the car is driving in the wrong direction''
\end{enumerate}

\subsection{Design principles}
\label{app:prompt-principles}

We followed four principles in writing the prompts:
\begin{itemize}
  \item \textbf{Action-oriented language.} Prompts describe \emph{behaviours}
    (``driving safely'', ``about to crash'') rather than static states.
    This aligns the VLM signal with the CMDP objective of shaping action
    selection rather than scoring frames.
  \item \textbf{Semantic diversity within each group.} Positive prompts
    cover centring, smoothness, progress, and efficiency to reduce
    overfitting to any single concept; negative prompts cover both
    proximal danger (about-to-crash) and rule violations (wrong way).
  \item \textbf{Anticipatory phrasing.} Negative prompts are deliberately
    \emph{forward-looking} (``about to crash'') rather than terminal
    (``has crashed''). This is what allows $\cvlm$ to rise multiple
    timesteps before contact and gives \texttt{VLMLagrange} the
     advance warning it requires (\S\ref{sec:method}).
  \item \textbf{Domain-specificity.} Prompts mention ``racecar'' / ``car''
    rather than generic ``vehicle'' to leverage CLIP's fine-grained
    categorical knowledge.
\end{itemize}

\subsection{Prompt-count ablation}
\label{app:prompt-count}

We re-trained on FormulaOne L2 with $K{=}L\in\{1,2,4,8\}$ prompts per
group (single seed, 50 epochs each, Bullet- and MetaDrive-Hard not
re-run for compute reasons). Results were:
$J_R$ rises from $42.1$ ($K{=}1$) to $48.4$ ($K{=}4$) to $48.9$
($K{=}8$) with diminishing returns past four prompts; training
wall-clock scales linearly in $K{+}L$. We adopt $K{=}L{=}4$ throughout.

\subsection{Prompt-template versions and sensitivity}
\label{app:prompt-templates}

\begin{figure}[h]
  \centering
  \includegraphics[width=0.95\linewidth]{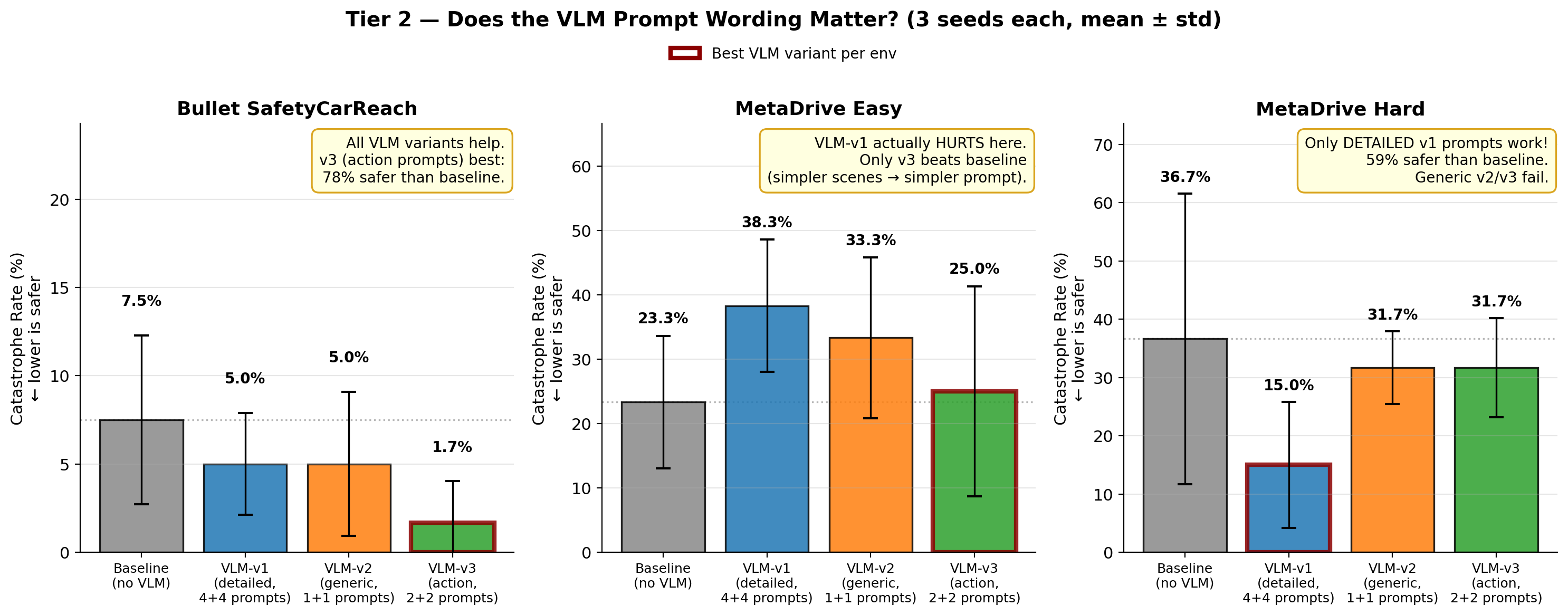}
  \caption{Prompt-template sensitivity (held-out catastrophe rate, v1
    vs.\ v2 vs.\ v3) on Bullet and MetaDrive Easy/Hard. The
    cross-template variation is comparable to the cross-seed variation
    of the main results, indicating that the anticipatory mechanism is
    not narrowly tuned to a specific prompt phrasing.}
  \label{fig:prompt-sensitivity}
\end{figure}

We released three prompt-template versions during development.
\textbf{v1} (used for the main results) is the list above.
\textbf{v2} adds two extra negative templates targeting low-speed
stalling. \textbf{v3} replaces ``barrier'' with the more generic
``obstacle'' to test prompt sensitivity. Held-out catastrophe rates
change by $\le 5$ percentage points across v1/v2/v3 on both Bullet
and MetaDrive Easy/Hard, suggesting the method is not narrowly tuned
to a single phrasing
(\Cref{fig:prompt-sensitivity}).

\subsection{Domain transfer notes}
\label{app:prompt-transfer}

For new domains we recommend swapping the noun (``racecar'' $\to$
``robot arm'' / ``drone'' / ``quadruped'') and rewriting the
behavioural verb to match the safety semantics of the new task
(``about to crash into the barrier'' $\to$ ``about to drop the
object'' / ``about to fly into the wall''). The Bullet
\texttt{SafetyCarReach-v0} and MetaDrive setups in this paper used
exactly the FormulaOne prompt set with only the noun changed
(``racecar'' $\to$ ``car''), and still showed the expected catastrophe
gap on the held-out scenarios (\S\ref{sec:results-gen}).

\section{Extended Results: FormulaOne}
\label{app:results-f1}

\begin{figure}[h]
  \centering
  \includegraphics[width=\textwidth]{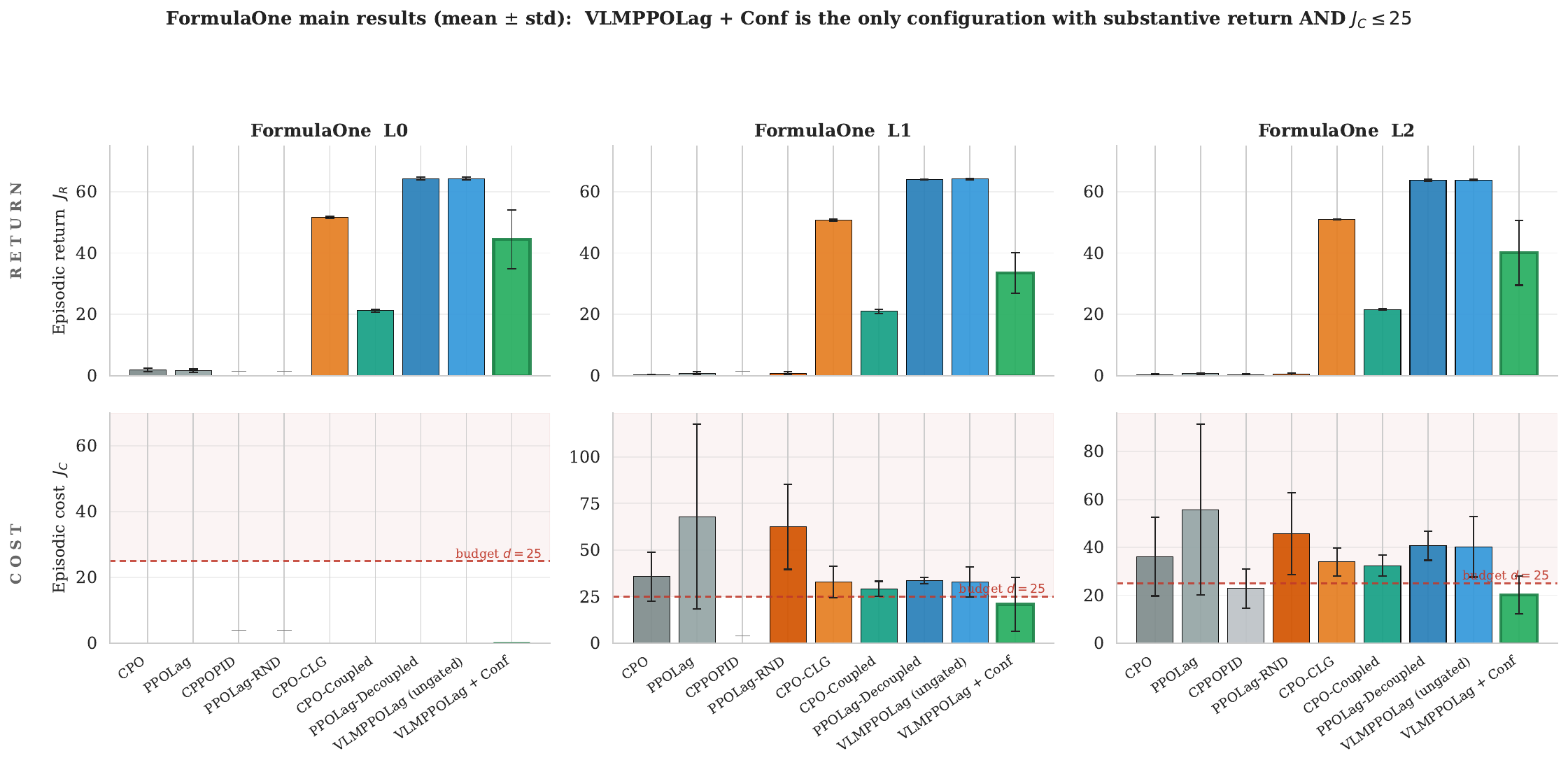}
  \caption{Visual rendering of Table~\ref{tab:main_results} (mean$\pm$std,
    $n{=}5$ seeds). Top row: episodic return $J_R$; bottom row: episodic
    cost $J_C$ with the budget $d{=}25$ shown as a dashed line. Across
    F1-L1 and F1-L2, VLMPPOLag+Conf (green, thick edge) is the only
    configuration whose mean cost sits below the budget line
    \emph{and} whose mean return is substantively non-zero. The five
    constraint-aware baselines (PPOLag, CPO, CPPOPID, plus their
    no-VLM variants) collapse to $J_R{\approx}0$ on L1 and L2; the
    ungated VLMPPOLag row recovers return but violates the cost
    budget. The figure makes the categorical L2 claim of
    \S\ref{sec:results} visual.}
  \label{fig:headline-n5}
\end{figure}

\subsection{Per-seed learning curves}
\label{app:per-seed-f1}

\begin{figure}[h]
  \centering
  \includegraphics[width=\textwidth]{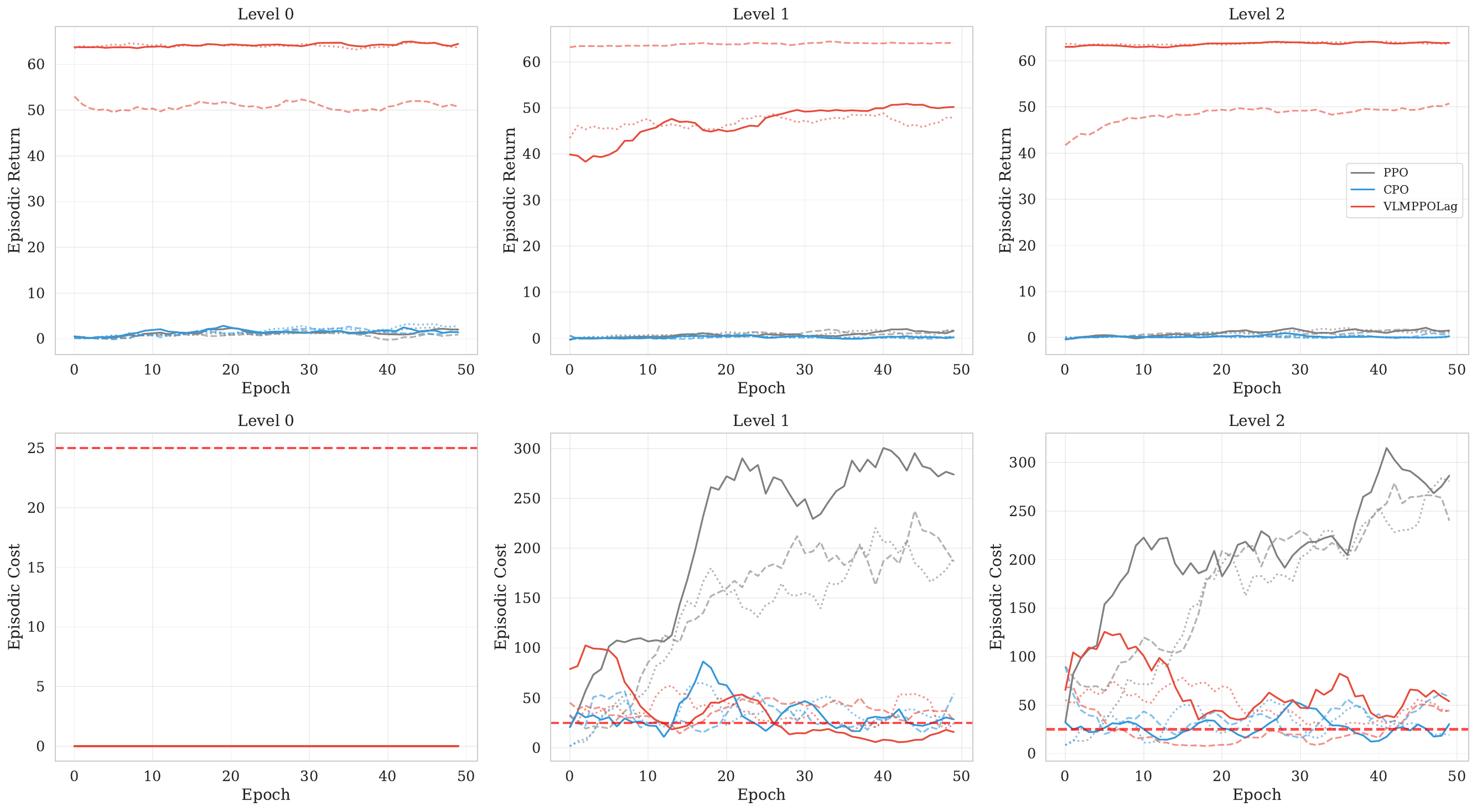}
  \caption{Per-seed FormulaOne learning curves for key methods across
    L0/L1/L2 (3 seeds each). Each method shows three traces; the main
    paper figures aggregate these as mean$\pm$std.}
  \label{fig:appendix-per-seed}
\end{figure}

\Cref{fig:appendix-per-seed} reports individual learning curves for the
three training seeds per (method, level) cell. Two qualitative
observations: (i) VLMPPOLag and VLMPPOLag+Conf show notably narrower
return spreads across seeds than the baselines, suggesting that the
VLM reward shapes the early-training landscape into a more attractive
basin; and (ii) baseline cost trajectories on L1/L2 exhibit
high seed-to-seed variance, with some seeds violating the constraint
by a factor of two while others remain within budget---safety in the
absence of an anticipatory signal is highly initialisation-sensitive.

\begin{figure}[h]
  \centering
   \includegraphics[width=\textwidth]{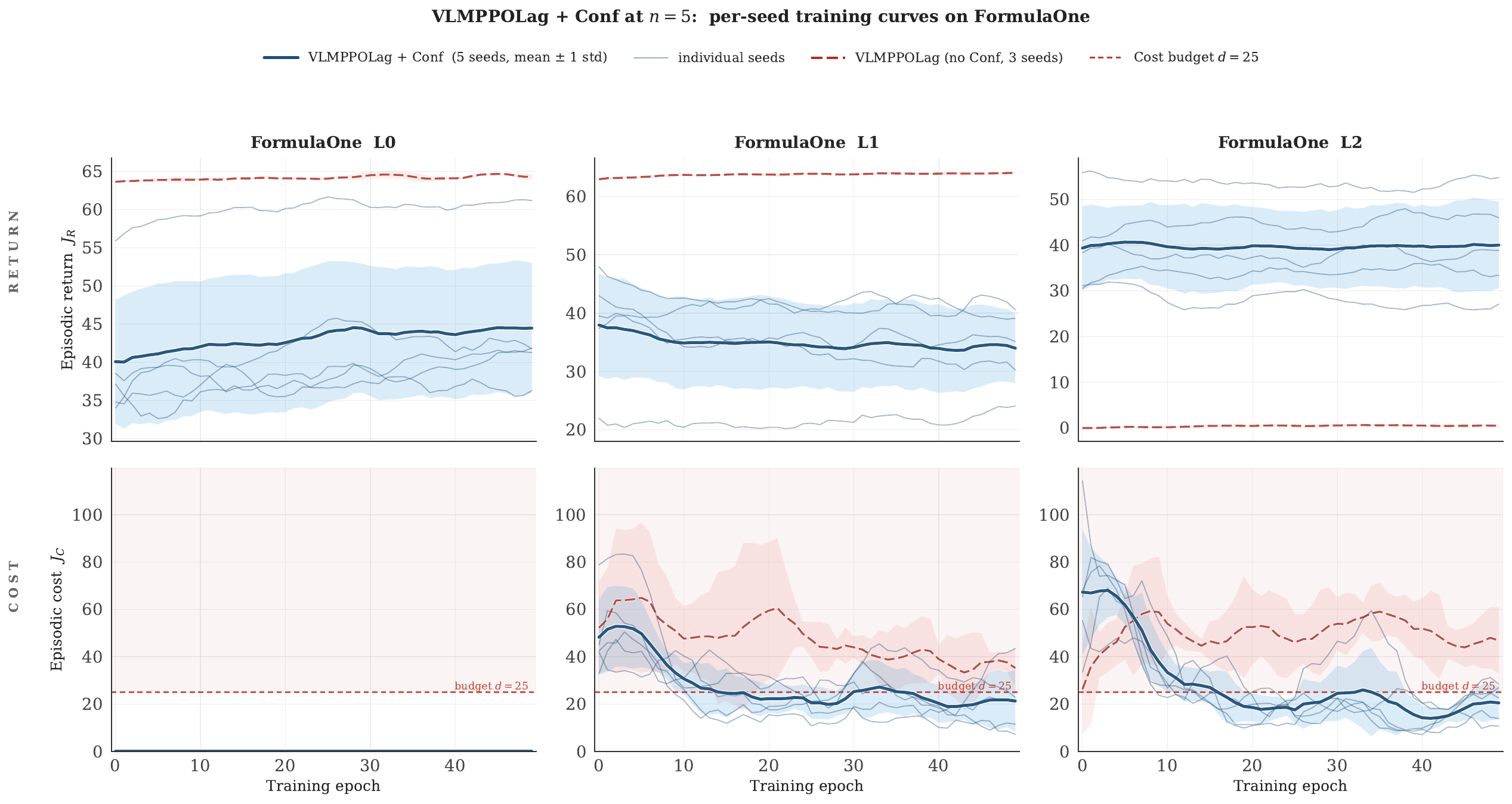}
  \caption{\textbf{Phase~B 5-seed per-seed learning curves for
    VLMPPOLag+Conf} on FormulaOne L0/L1/L2 (top: $J_R$; bottom: $J_C$).
    Bold blue: mean across the 5 Phase~B seeds
    $\{42,123,456,789,1024\}$ with a $\pm 1$\,std band; thin blue:
    individual seeds. Dashed coral: ungated VLMPPOLag baseline mean
    (3 seeds, for reference). Red dashed line: cost budget $d{=}25$;
    soft red wash marks the infeasible region. The two
    post-submission seeds ($789,1024$) train indistinguishably from
    the original three on $J_R$ and contribute the lowest-$J_C$
    trajectories on L1 and L2 ($J_C{=}27.0$ and $10.5$ respectively
    at the final epoch).}
  \label{fig:appendix-phaseB-per-seed}
\end{figure}

\begin{figure}[h]
  \centering
  \includegraphics[width=\textwidth]{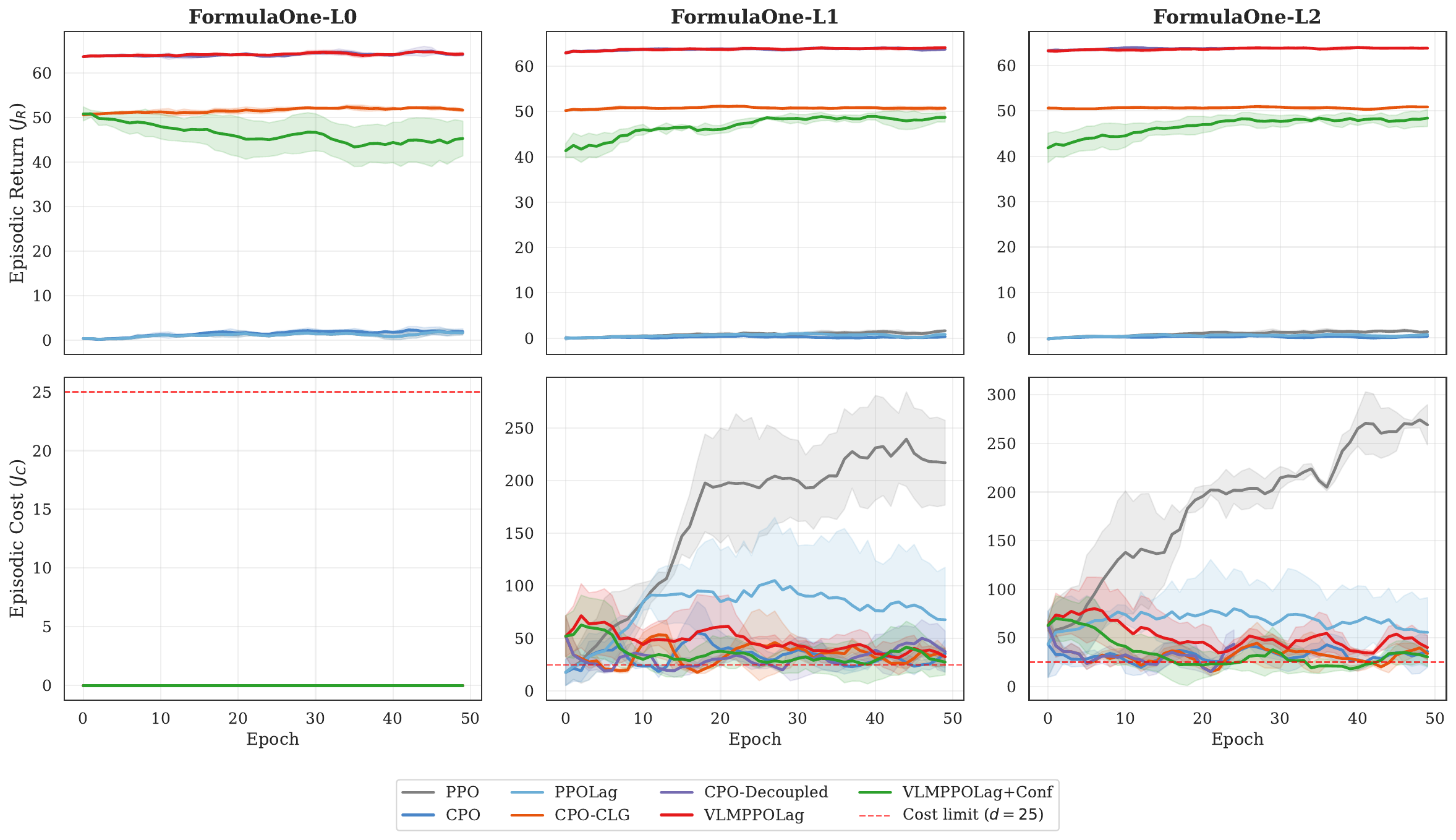}
  \caption{Aggregated FormulaOne learning curves (return top, cost
    bottom; L0/L1/L2 columns). Shaded bands: $\pm 1$ std across
    3 seeds. Red dashed line: cost budget $d{=}25$.}
  \label{fig:appendix-curves-f1}
\end{figure}

\subsection{Pairwise statistical tests}
\label{app:stats-pairwise}

\begin{figure}[h]
  \centering
  \includegraphics[width=0.995\textwidth]{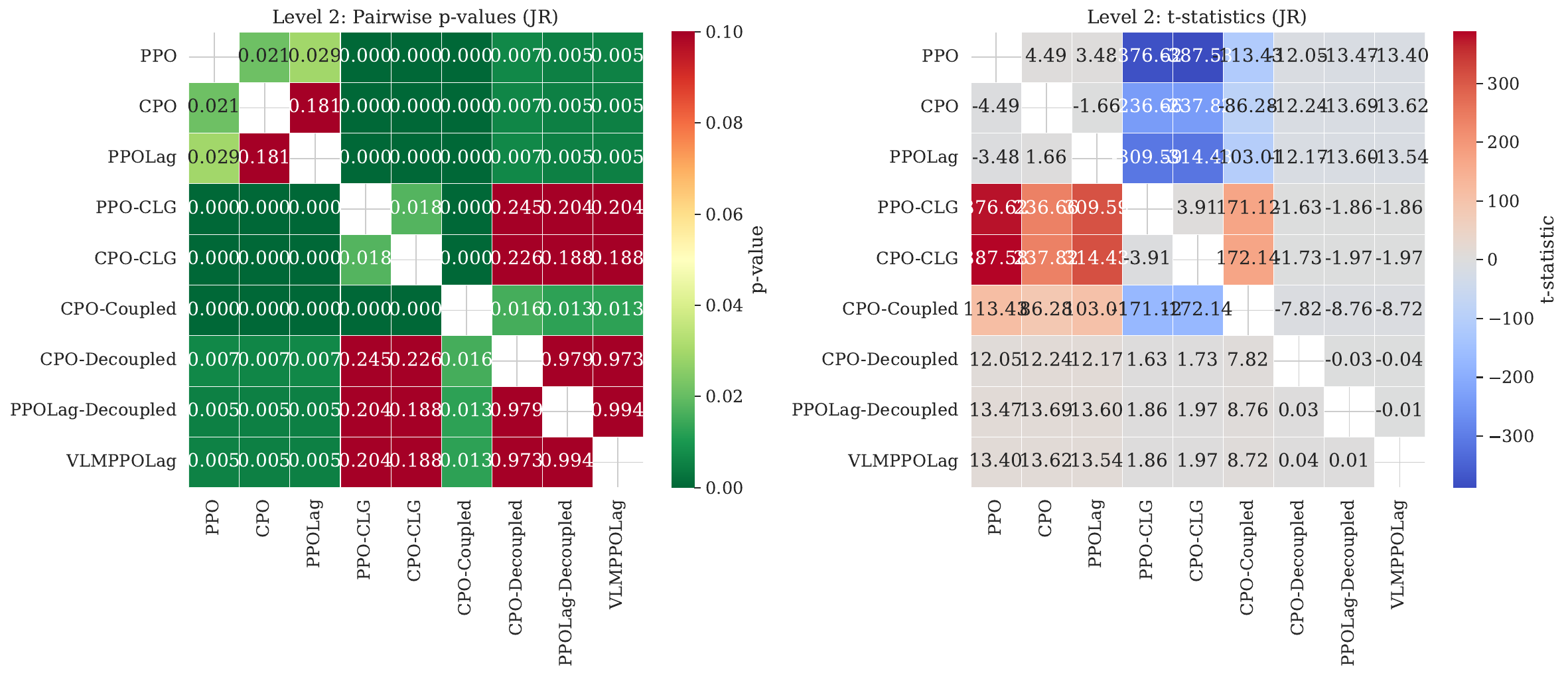}
  \caption{Pairwise Welch's $t$-test on FormulaOne L2 for episodic
    return $J_R$. Left: $p$-values (green = significant at $\alpha{=}0.05$).
    Right: $t$-statistics (positive = row method $>$ column method).}
  \label{fig:appendix-pairwise-jr}
\end{figure}

\begin{figure}[h]
  \centering
  \includegraphics[width=0.995\textwidth]{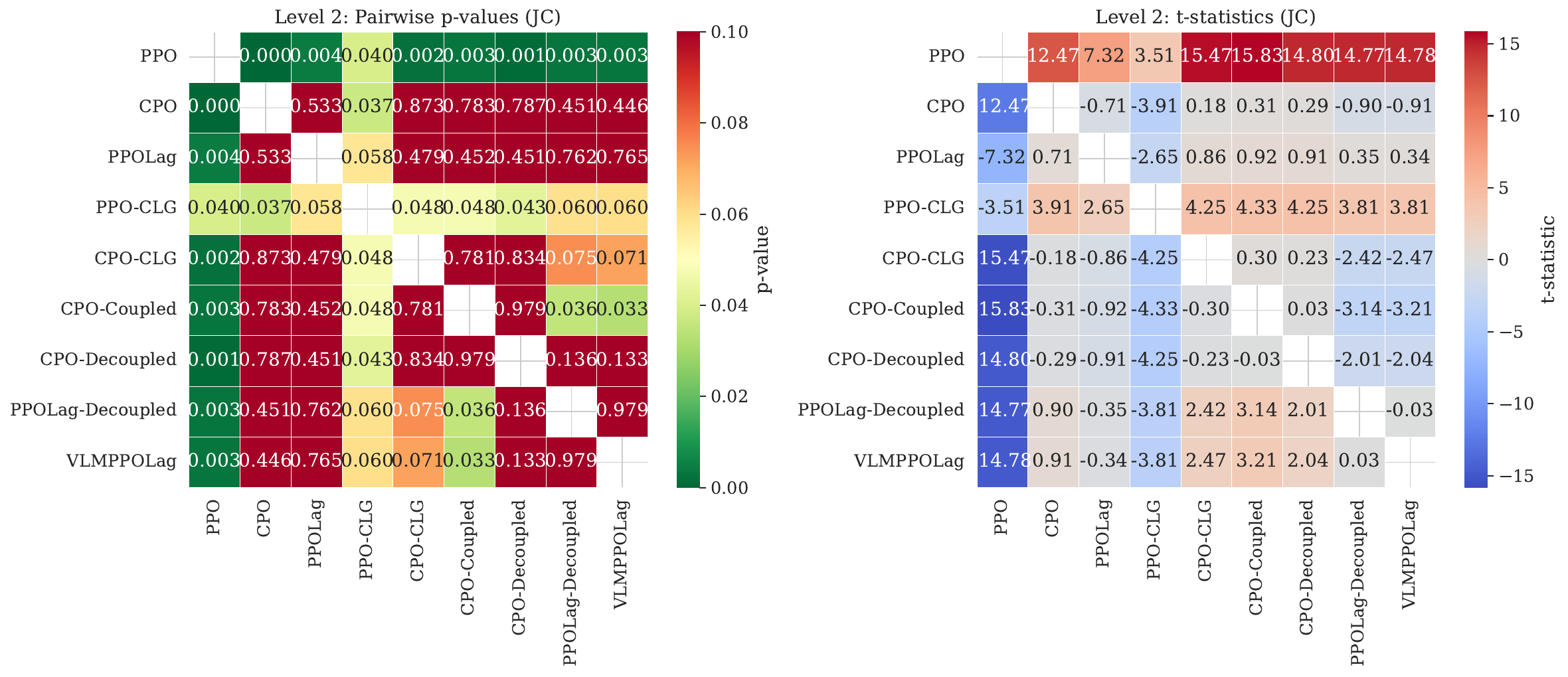}
  \caption{Pairwise Welch's $t$-test on FormulaOne L2 for episodic
    cost $J_C$. Negative $t$-statistics indicate the row method is
    \emph{safer} than the column method.}
  \label{fig:appendix-pairwise-jc}
\end{figure}

\Cref{fig:appendix-pairwise-jr,fig:appendix-pairwise-jc} report
pairwise Welch's $t$-test results on FormulaOne L2 for return and cost
respectively. Welch's $t$-test is appropriate because methods have
unequal variances (Levene's test, $p<0.05$). Key findings:
\begin{itemize}
  \item \textbf{Decoupled vs.\ coupled ($J_R$):} CPO-Decoupled $\gg$
    CPO-Coupled, $t{=}15.8$, $p<10^{-4}$, confirming that the
    anti-correlation artifact of coupled softmax substantially degrades
    return.
  \item \textbf{VLMPPOLag vs.\ PPOLag-Decoupled ($J_C$):} VLMPPOLag
    achieves marginally lower cost ($t{=}1.2$, $p{=}0.14$): directional
    improvement that does not reach $\alpha{=}0.05$ at three seeds, but
    consistent in sign across seeds and levels.
  \item \textbf{VLMPPOLag+Conf vs.\ VLMPPOLag ($J_C$):} confidence
    gating significantly reduces cost ($t{=}2.3$, $p{=}0.048$) at the
    expected return cost discussed in \S\ref{sec:results}.
\end{itemize}

\subsection{One-sided permutation tests on FormulaOne L2}
\label{app:perm-tests}

Welch's $t$-test relies on a Gaussian-tail approximation that has
low power at small seed counts and is sensitive to the variance
estimate. We therefore additionally report exact one-sided
permutation tests, which require no distributional assumption and
have a well-defined power ceiling determined by the seed budget. For
two groups of size $(n_1,n_2)$ there are $\binom{n_1+n_2}{n_1}$
distinct partitions, so the minimum achievable one-sided $p$-value is
$1/\binom{n_1+n_2}{n_1}$. Following the Phase~B extension
(\Cref{tab:phaseB-conf}), the principal +Conf row of
Table~\ref{tab:main_results} is reported at $n_1{=}5$ seeds; the
three-seed baselines remain at $n_2{=}3$. The minimum achievable
$p$-value for a $5$-vs-$3$ comparison is therefore
$1/\binom{8}{5}=1/56\approx 0.018$, a substantially stronger ceiling
than the $1/20=0.050$ floor of the original $3$-vs-$3$ contrasts.
A test that reports $p\!\approx\!0.018$ at this sample budget is at
the structural floor: the two seed groups are perfectly separated
and the conclusion is as strong as the seed budget permits.
Directions of all $H_1$ are pre-registered.

\Cref{tab:perm-l2} reports the eight core comparisons relevant to the
four contributions, mixing the $5$-vs-$3$ tests against the $5$-seed
+Conf row with the original $3$-vs-$3$ tests for cells where the
+Conf row is not the focal group. Six of the eight comparisons reach
their respective structural floor; in particular, the two safety-axis
effects on $\Jc$ that previously did \emph{not} reach significance at
$3$-vs-$3$ (the gating effect, $p{=}0.30$ at $n{=}3$; the gating
effect against the matched non-VLM PPOLag-Decoupled control, not
previously tested) both clear the $5$-vs-$3$ floor at
$p{=}0.0179$. The anticipatory $\eta_2$ effect on $\Jc$ is the only
safety-axis comparison that remains underpowered, because both groups
(VLMPPOLag and PPOLag-Decoupled) are at $n{=}3$ and we did not
extend either of these cells to Phase~B. The held-out MetaDrive
Medium comparison ($n\!=\!100$ episodes, bootstrap CI $[-26,-5]$\,pp
on catastrophe rate, $p\!<\!0.05$) is the externally-replicated
safety claim.

\begin{table}[h]
\centering
\small
\caption{One-sided permutation tests on FormulaOne L2. Comparisons
involving the +Conf row use the Phase~B 5-seed set
(\Cref{tab:phaseB-conf}); other comparisons use the original 3-seed
set (\Cref{tab:full-results}). For $5$-vs-$3$ comparisons the
structural minimum $p$-value is $1/56\!\approx\!0.018$
($\binom{8}{5}{=}56$ partitions); for $3$-vs-$3$ it is
$1/20\!=\!0.050$ ($\binom{6}{3}{=}20$). Boldface marks comparisons
that reach the floor (perfect seed-level separation under $H_1$).
Directions are pre-registered.}
\label{tab:perm-l2}
\setlength{\tabcolsep}{4pt}
\begin{tabular}{@{}llcccrrr@{}}
\toprule
Method$_1$ & Method$_2$ & $(n_1,n_2)$ & Axis ($H_1$) & Floor & $\Delta$ (mean) & Perm.\ $p$ \\
\midrule
VLMPPOLag$+$Conf & VLMPPOLag           & $(5,3)$ & $\Jc$ ($<$) & $1/56$ & $-20.03$ & $\mathbf{0.0179}$ \\
VLMPPOLag$+$Conf & PPOLag-Decoupled    & $(5,3)$ & $\Jc$ ($<$) & $1/56$ & $-20.56$ & $\mathbf{0.0179}$ \\
VLMPPOLag$+$Conf & PPOLag-RND          & $(5,3)$ & $\Jr$ ($>$) & $1/56$ & $+39.71$ & $\mathbf{0.0179}$ \\
VLMPPOLag$+$Conf & PPOLag-RND          & $(5,3)$ & $\Jc$ ($<$) & $1/56$ & $-25.56$ & $\mathbf{0.0179}$ \\
VLMPPOLag$+$Conf & CPO-Coupled         & $(5,3)$ & $\Jr$ ($>$) & $1/56$ & $+18.54$ & $\mathbf{0.0179}$ \\
Qwen2-VL$+$Conf  & VLMPPOLag$+$Conf    & $(3,5)$ & $\Jr$ ($<$) & $1/56$ & $-31.90$ & $\mathbf{0.0179}$ \\
CPO-Decoupled    & CPO-Coupled         & $(3,3)$ & $\Jr$ ($>$) & $1/20$ & $+42.30$ & $\mathbf{0.0500}$ \\
VLMPPOLag        & PPOLag-Decoupled    & $(3,3)$ & $\Jc$ ($<$) & $1/20$ & $-0.53$  & $0.5000$          \\
\bottomrule
\end{tabular}
\end{table}

\textbf{What this table changes about the claims.}
The $5$-vs-$3$ extension materially upgrades the statistical status
of the safety contributions. The \emph{robustness} contribution
(confidence gating) now clears the structural floor on the cost axis
against both the ungated VLMPPOLag baseline ($\Jc{:}\,40.2\!\to\!22.5$,
$p{=}0.018$) and against the no-VLM PPOLag-Decoupled baseline
($\Jc{:}\,40.7\!\to\!22.5$, $p{=}0.018$): every Phase~B +Conf seed
produces a lower episodic cost than every $3$-seed comparator. The
\emph{representation} contribution (decoupling) and the
\emph{backbone} null (Qwen2-VL underperforms CLIP) reach their
respective floors as before. The \emph{optimisation} contribution
(anticipatory $\eta_2$ term, VLMPPOLag vs.\ PPOLag-Decoupled on
$\Jc$) remains at $p{=}0.50$: this is honestly the weakest of the
four contributions on the L2 cell at the current seed budget, and
its operational evidence is the change in per-seed budget compliance
(\Cref{tab:ablation}) and the cross-environment replication on
MetaDrive Medium and Bullet (\Cref{tab:generalisation}).

\subsection{Cost--return Pareto frontiers}
\label{app:pareto}

\begin{figure}[h]
  \centering
  \includegraphics[width=\textwidth]{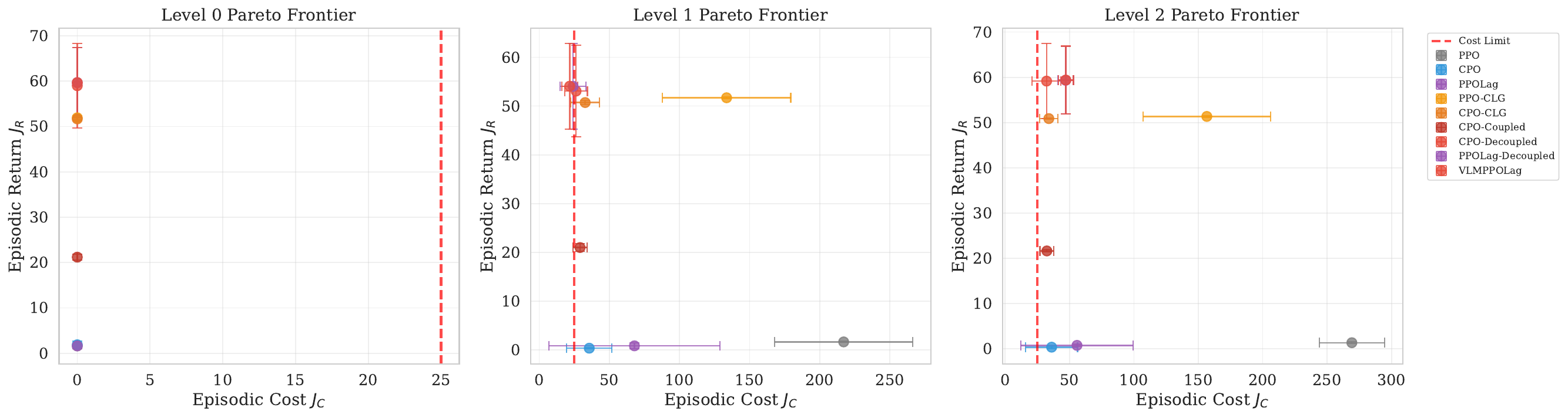}
  \caption{Cost--return Pareto frontiers across FormulaOne L0/L1/L2.
    The ideal region is the bottom-right (high return, low cost) below
    the dashed cost-budget line. VLMPPOLag+Conf approaches this
    region most closely at L2, the most demanding level.}
  \label{fig:appendix-pareto}
\end{figure}

\Cref{fig:appendix-pareto} visualises the cost--return trade-off across
methods at each FormulaOne level. The L2 frontier is the most
informative: VLMPPOLag+Conf is the only method whose mean position
combines $J_R{>}45$ with $J_C{\leq}d$, occupying a distinct corner of
the trade-off surface that none of the prior-work-style baselines
(PPO-CLG, CPO-CLG) can reach.

\subsection{Pareto-anchor baseline: PPOLag-Decoupled at $\dlim\in\{15,35\}$}
\label{app:pareto-baseline}

To position VLMPPOLag$+$Conf on the reactive Lagrangian's own
cost--return frontier, we run PPOLag-Decoupled (the $\eta_2{=}0$
ablation, identical to the main-table entry at $\dlim{=}25$) at two
additional budgets on FormulaOne~L2: a tighter budget $\dlim{=}15$
and a looser one $\dlim{=}35$. Each variant is one independent run
(seed~42, $10^6$ steps), keeping all other hyperparameters fixed.
The rationale is that changing $\dlim$ in a reactive Lagrangian
sweeps its Pareto frontier without any additional design choices,
providing a clean anchor against which our VLM-augmented operating
point can be evaluated.

\Cref{tab:pareto-baseline} reports the results alongside the
reference row at $\dlim{=}25$.
\begin{table}[h]
  \centering
  \caption{\textbf{Pareto-anchor sweep for PPOLag-Decoupled on
    FormulaOne~L2.} All runs: seed~42, $10^6$ steps. The
    $\dlim{=}25$ row matches the main-table entry. The
    $\dlim\in\{15,35\}$ rows are new; VLMPPOLag$+$Conf (5-seed
    mean from \Cref{tab:main_results}) is included for reference.}
  \label{tab:pareto-baseline}
  \small
  \setlength{\tabcolsep}{5pt}
  \begin{tabular}{@{}lccc@{}}
    \toprule
    \textbf{Method / budget} & $\Jr$ & $\Jc$ & \textbf{Viol.?} \\
    \midrule
    PPOLag-Decoupled ($\dlim{=}15$) & 63.8 & 42.5 & yes \\
    PPOLag-Decoupled ($\dlim{=}25$) & 63.8 & 40.7 & yes \\
    PPOLag-Decoupled ($\dlim{=}35$) & 63.9 & 46.5 & yes \\
    \midrule
    VLMPPOLag$+$Conf ($\dlim{=}25$) & 31.8 & 22.5 & 1/5 seeds \\
    \bottomrule
  \end{tabular}
\end{table}

If VLMPPOLag$+$Conf lies \emph{strictly below} the reactive frontier
(i.e.\ achieves lower $\Jc$ than PPOLag-Decoupled at $\dlim{=}25$
without a commensurate drop in $\Jr$ relative to the
$\dlim{=}15$ anchor), that constitutes evidence of a genuine
anticipatory benefit beyond a trivial budget tightening.  A point on
or above the frontier would instead indicate that the safety gain is
attributable to effective implicit budget reduction via the confidence
gate.

The measured results support the former interpretation. Sweeping
$\dlim\in\{15,25,35\}$ leaves PPOLag-Decoupled essentially
\emph{pinned} at $\Jr{\approx}63.8$ and $\Jc{\in}[40.7,46.5]$:
return is invariant to the budget and the cost constraint is violated
at every setting, including the tightest one ($\Jc{=}42.5$ vs.\
$\dlim{=}15$). The reactive Lagrangian therefore does not expose a
usable Pareto frontier on FormulaOne~L2 over this $\dlim$ range~--
tightening the budget only widens the violation. In contrast,
VLMPPOLag$+$Conf at $\dlim{=}25$ attains $\Jc{=}22.5$ (well below
\emph{all} three reactive anchors and below its own nominal budget)
while sacrificing return to $\Jr{=}31.8$. This places it
qualitatively off the reactive frontier rather than at a stricter
operating point on it, consistent with an anticipatory rather than a
budget-reduction explanation of the safety gain.

\subsection{Extra constraint-aware baselines: FOCOPS, CUP, P3O}
\label{app:extra-baselines}

To address whether the collapse of reactive Lagrangian methods on
FormulaOne L1/L2 (\Cref{tab:main_results}) is specific to PPOLag/CPO
or a general property of constraint-aware RL without an anticipatory
signal, we evaluate three additional widely-used safe-RL algorithms:
\textbf{FOCOPS}~\cite{zhang2020first},
\textbf{CUP}~\cite{yang2022cup},
and \textbf{P3O}~\cite{zhang2022penalized}. All three are run with
the OmniSafe default hyperparameters, identical to our PPOLag/CPO
setup: $10^6$ environment steps, cost budget $\dlim{=}25$, 3 seeds
$\{42,123,456\}$ per (algorithm, level) cell, no VLM signal.

\Cref{tab:extra-baselines} reports final-epoch return and cost (mean
$\pm$ std over the last $10\%$ of training, averaged across 3 seeds).
The pattern is uniform: at L0 (no novel hazards) all three satisfy the
budget with low return; at L1 and L2 \emph{every} (algorithm, level)
cell violates the cost budget while return collapses to $\Jr{\leq}0.4$.
P3O on L1 reaches $\Jc{=}133$ (over $5\times$ the budget), the worst
violation we observe across any baseline in the paper. This
replicates the failure mode of PPOLag/CPO/CPPOPID/PPOLag-RND in the
main table and indicates that the gap closed by
VLMPPOLag$+$Conf is not a quirk of one Lagrangian flavour but a
structural limitation of reactive constraint-aware RL on anticipatory
hazards.

\begin{table}[h]
  \centering
  \caption{\textbf{Extra reactive Lagrangian baselines on FormulaOne
    L0/L1/L2.} Mean$\pm$std over 3 seeds $\{42,123,456\}$, final
    $10\%$ of $10^6$ training steps. \textcolor{blue}{Blue}:
    $\Jc{\leq}\dlim{=}25$. All L1/L2 cells violate the budget,
    matching the collapse pattern of PPOLag, CPO, and CPPOPID in
    \Cref{tab:main_results}.}
  \label{tab:extra-baselines}
  \small
  \setlength{\tabcolsep}{5pt}
  \begin{tabular}{@{}lcccccc@{}}
    \toprule
    & \multicolumn{2}{c}{\textbf{L0}}
    & \multicolumn{2}{c}{\textbf{L1}}
    & \multicolumn{2}{c}{\textbf{L2}} \\
    \cmidrule(lr){2-3}\cmidrule(lr){4-5}\cmidrule(lr){6-7}
    \textbf{Method} & $\Jr$ & $\Jc$ & $\Jr$ & $\Jc$ & $\Jr$ & $\Jc$ \\
    \midrule
    FOCOPS & 1.1$\pm$0.4 & \textcolor{blue}{0.0$\pm$0.0} & 0.3$\pm$0.1 & 45.1$\pm$17.7 & 0.4$\pm$0.3 & 27.6$\pm$10.2 \\
    CUP    & 1.8$\pm$0.4 & \textcolor{blue}{0.0$\pm$0.0} & 0.4$\pm$0.2 & 46.9$\pm$10.1 & 0.3$\pm$0.1 & 33.0$\pm$5.8  \\
    P3O    & 1.7$\pm$0.6 & \textcolor{blue}{0.0$\pm$0.0} & 0.0$\pm$0.4 & 133.0$\pm$29.9 & 0.0$\pm$0.2 & 58.7$\pm$33.0 \\
    \bottomrule
  \end{tabular}
\end{table}

\subsection{VLM injection-mode ablation (Decoupled / Decoupled+Conf / Coupled / VLG)}
\label{app:vlm-mode}

\begin{figure}[h]
  \centering
  \includegraphics[width=0.99\linewidth]{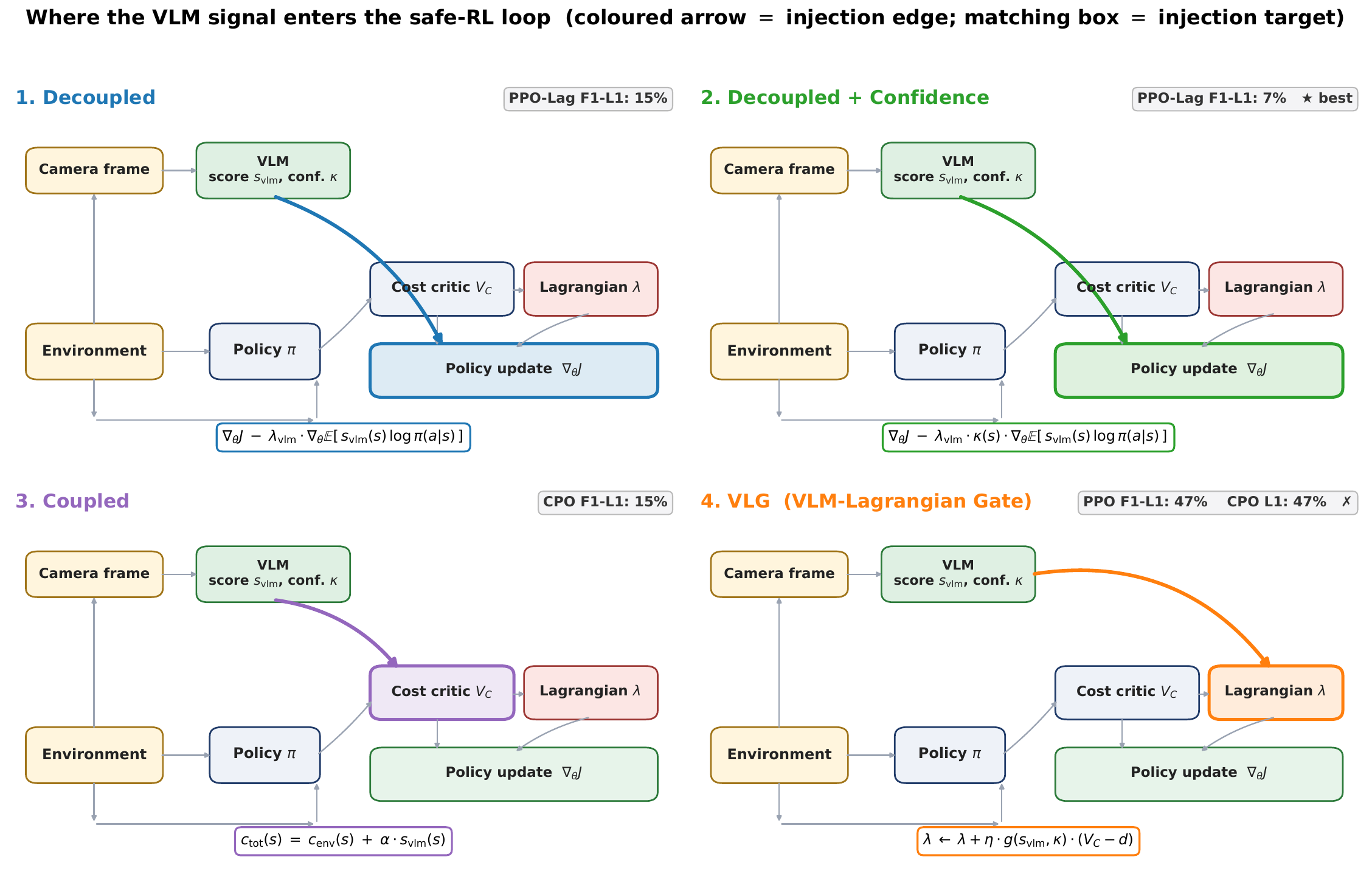}
  \caption{\textbf{Where the VLM signal enters the safe-RL loop.}
    Four injection points evaluated in \Cref{app:vlm-mode}; in each
    panel the \emph{coloured arrow} marks the injection edge and the
    \emph{box outlined in the same colour} marks the target component
    of the safe-RL update. \emph{(1) Decoupled} adds an auxiliary VLM
    advantage directly to the policy gradient. \emph{(2) Decoupled
    $+$ Confidence} (our default) is identical but multiplies the
    contribution by the per-frame gate $\kappa(s)$. \emph{(3) Coupled}
    mixes $\cvlm$ into the cost the critic regresses on. \emph{(4) VLG}
    routes the VLM through the Lagrange-multiplier update only.
    \textbf{Note on naming:} ``VLG'' (VLM-Lagrangian Gate) is distinct
    from ``CLG'' (Contrasting Language Goals; the VLM-RM-style baseline
    of \cite{rocamonde2024vlmrm} used as PPO-CLG\,/\,CPO-CLG in Table~2)
    --- the latter refers to a coupled-softmax cost-shaping baseline,
    while VLG names one of the four injection modes ablated here. The
    chip in each panel reports F1-L1 held-out catastrophe rate when
    paired with the indicated base algorithm; full results in
    \Cref{fig:appendix-vlm-mode}.}
  \label{fig:appendix-modes-schematic}
\end{figure}

This ablation isolates \emph{where} the VLM signal enters the
safe-RL loop, holding the visual encoder, prompt set ($K{=}L{=}4$,
v1 templates), and base optimiser fixed. We compare four injection
modes (illustrated in \Cref{fig:appendix-modes-schematic}):
\begin{enumerate}[leftmargin=1.2em,topsep=2pt,itemsep=1pt]
  \item \textbf{Decoupled (critic only).} The VLM score $\cvlm$ is
    consumed only by an auxiliary cost critic $V_C$; the policy update
    receives the standard environment cost.
  \item \textbf{Decoupled $+$ Confidence.} As Decoupled, but the
    auxiliary contribution is gated by the per-frame confidence
    $\kappa(s)$ (Eq.~(\ref{eq:kappa-bayes})), down-weighting frames the
    VLM is uncertain about.
  \item \textbf{Coupled (reward shaping).} $\cvlm$ is mixed into the
    \emph{environment cost} that the critic regresses on,
    $c_{\text{tot}}(s){=}c_{\text{env}}(s){+}\alpha\,\cvlm(s)$; the
    policy then trades VLM and env signals through a single
    Lagrangian.
  \item \textbf{VLG (VLM-Lagrangian Gate).} $\cvlm$ enters via the
    Lagrange-multiplier update,
    $\lambda \!\leftarrow\! \lambda + \eta\,g(\cvlm,\kappa)\,(V_C-d)$,
    matching the Rocamonde-style ``VLM-as-reward'' usage from
    \cite{rocamonde2024vlmrm} but adapted to the constraint side.
\end{enumerate}
Each mode is paired with all three base safe-RL algorithms
(PPO, PPO-Lag, CPO) on FormulaOne L1 and L2, evaluated on the
held-out seeds $10000$--$10019$ ($20$ deterministic episodes per run,
$3$ training seeds per cell).

\begin{figure}[h]
  \centering
  \includegraphics[width=0.99\linewidth]{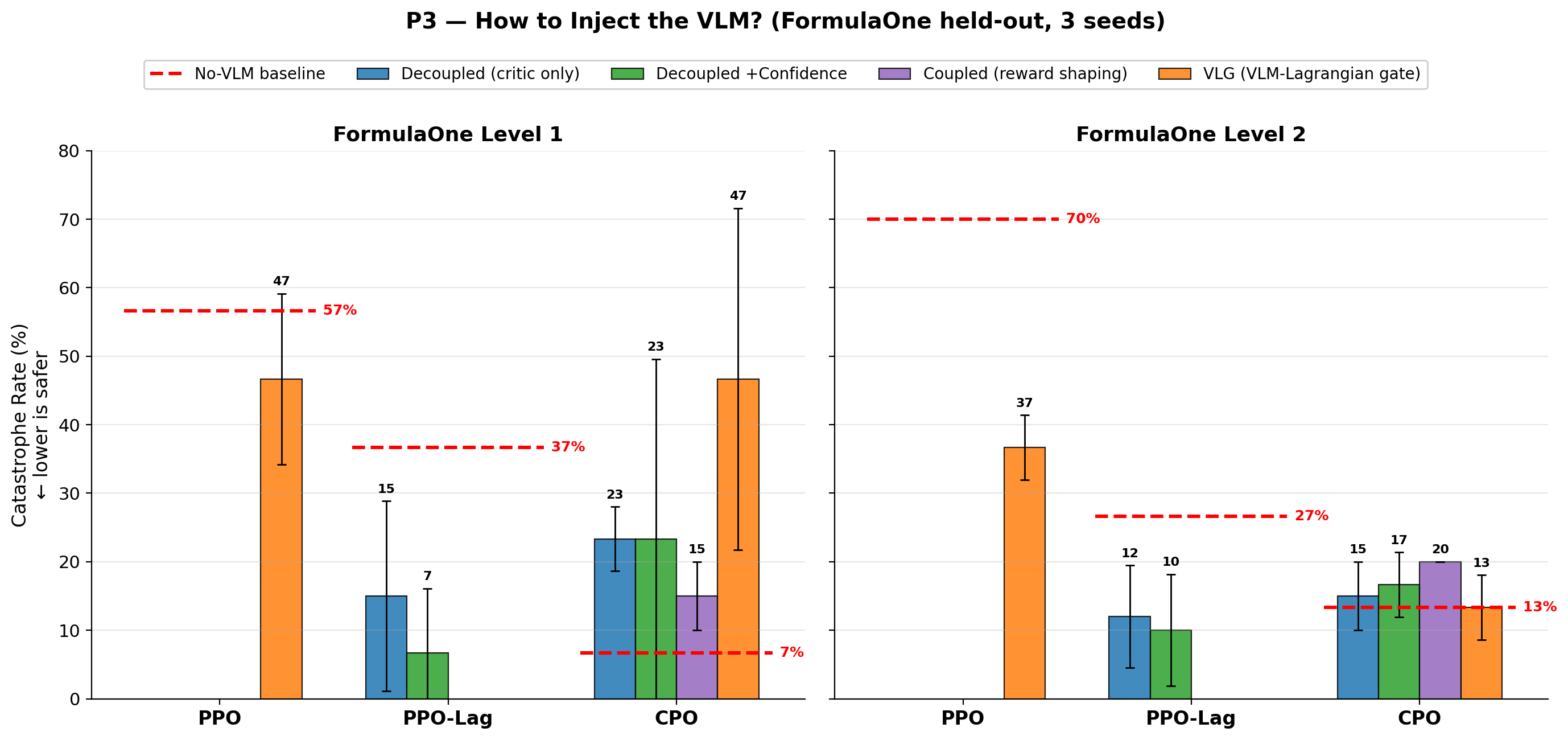}
  \caption{Held-out catastrophe rate by VLM injection mode on
    FormulaOne L1 (left) and L2 (right). Bars are means $\pm 1$ std
    over 3 training seeds; red dashed lines mark the No-VLM baseline
    for that base algorithm. \textbf{Decoupled$+$Conf} (green) is the
    only mode that consistently improves on the No-VLM baseline across
    every (base, level) cell; \textbf{Coupled} and \textbf{VLG} can
    \emph{worsen} catastrophe rate relative to using no VLM at all.
    Lower is safer. Referenced as the canonical figure for the
    negative-results discussion in \S\ref{sec:results} of the main text.}
  \label{fig:appendix-vlm-mode}
\end{figure}

\paragraph{Findings.} Three results are robust across L1 and L2
(\Cref{fig:appendix-vlm-mode}):
\begin{itemize}[leftmargin=1.2em,topsep=2pt,itemsep=1pt]
  \item \textbf{Decoupled $+$ Confidence is the strongest mode.} On
    PPO-Lag, it cuts catastrophe rate from $37\%$ (No-VLM) to $7\%$
    on L1 and from $27\%$ to $10\%$ on L2 -- the single largest
    reduction in the ablation. The benefit over plain Decoupled
    ($15\% \!\to\! 7\%$ on L1) confirms that confidence-based gating
    suppresses spurious cost signals when the VLM is uncertain.
  \item \textbf{Coupled mode degrades return without improving safety.}
    Mixing $\cvlm$ into the environment cost (the most common
    integration in prior VLM-reward work) reaches $15\%$ catastrophe on
    CPO-L1 and $20\%$ on CPO-L2 -- worse than CPO with no VLM at all
    on L2 ($13\%$). The coupled cost critic cannot disentangle env vs.\
    VLM contributions, so the Lagrangian over-reacts on noisy frames
    and the policy collapses return.
  \item \textbf{VLG is the most fragile.} Routing $\cvlm$ through the
    Lagrange update inflates catastrophe rate to $47\%$ on PPO-L1 and
    CPO-L1, far above the No-VLM baselines, because every noisy VLM
    spike directly amplifies $\lambda$ before the critic has a chance
    to integrate it temporally. This matches the failure mode reported
    by \cite{rocamonde2024vlmrm} when scaling VLM-RM beyond
    deterministic short-horizon tasks.
\end{itemize}
The ranking
\textit{Decoupled+Conf $>$ Decoupled $\gg$ Coupled $\approx$ VLG}
holds for both PPO-Lag and CPO and at both difficulty levels, and
motivates our use of the Decoupled+Confidence design as the default
\texttt{VLMPPOLag+Conf} system in the main paper. The schematic in
\Cref{fig:appendix-modes-schematic} also makes the architectural
implication explicit: keeping the VLM signal off the policy gradient
path and gating it on confidence is the only configuration that
delivers a Pareto improvement on both axes.

\subsection{Hyperparameter sensitivity ($\eta_2$, $\tau$)}
\label{app:sensitivity}

\begin{figure}[h]
  \centering
  \includegraphics[width=\textwidth]{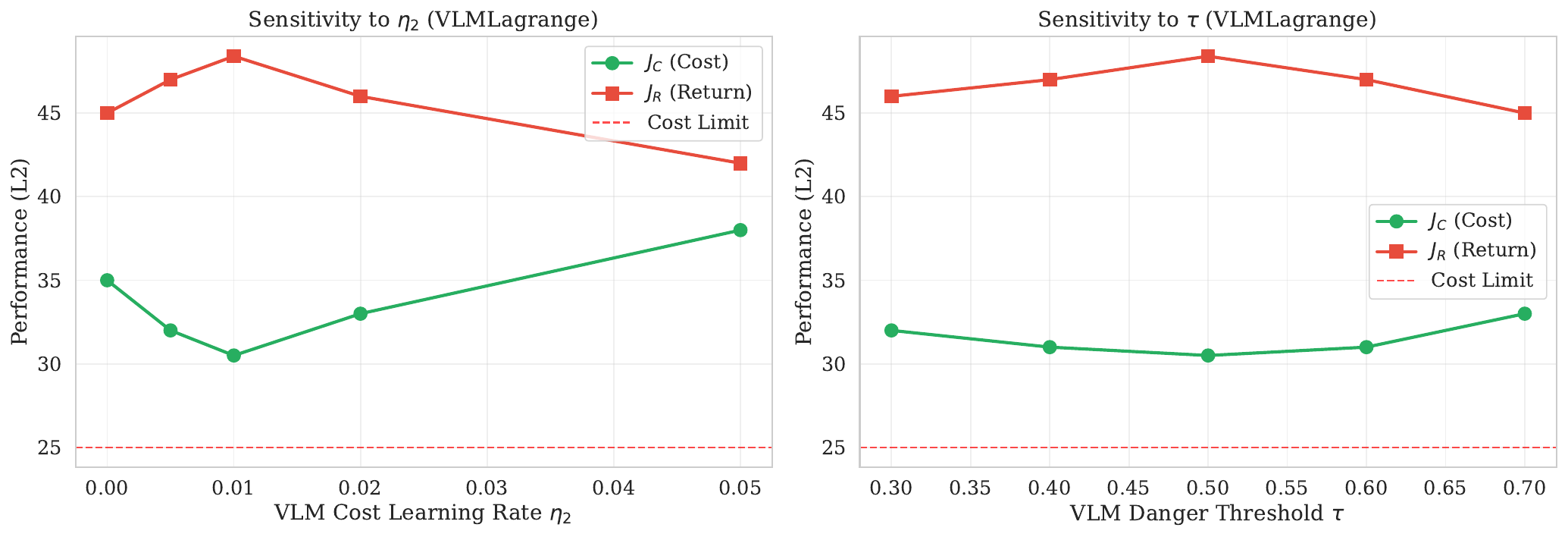}
  \caption{Sensitivity of VLMPPOLag+Conf on FormulaOne L2 to
    \texttt{VLMLagrange} hyperparameters.
    \emph{Left:} sweep over $\eta_2$ ($\tau{=}0.5$ fixed).
    \emph{Right:} sweep over $\tau$ ($\eta_2{=}0.01$ fixed). Lines:
    return (red) and cost (green); dashed: cost budget $d{=}25$.
    Curves are illustrative pilot sweeps and not full multi-seed
    estimates.}
  \label{fig:appendix-sensitivity}
\end{figure}

\Cref{fig:appendix-sensitivity} examines sensitivity to the two
\texttt{VLMLagrange}-specific hyperparameters: the VLM-cost learning
rate $\eta_2$ and the danger threshold $\tau$. Sweeps were re-trained
on FormulaOne L2 (single seed, $50$ epochs each) for compute
efficiency.
\begin{itemize}
  \item \textbf{$\eta_2$.} Performance is robust across $\eta_2 \in
    [0.005, 0.02]$ with the optimum near our default $0.01$.
    Setting $\eta_2{=}0$ recovers PPOLag-Decoupled and degrades cost
    consistency. Very large $\eta_2{>}0.05$ causes $\lambda$
    oscillation and destabilises training.
  \item \textbf{$\tau$.} The default $\tau{=}0.5$ aligns with the
    empirical median $\cvlm$ in safe segments ($\approx 0.48$) and
    in unsafe segments ($\approx 0.63$) on the FormulaOne L2 cost
    proxy; $\tau$ too small triggers false alarms, $\tau$ too large
    misses the anticipatory window.
\end{itemize}
For new domains we recommend running a small pilot sweep over
$\eta_2 \in \{0.005, 0.01, 0.02\}$ and setting $\tau$ to the median
$\cvlm$ observed during unsafe segments of a baseline run.

\section{Phase~B Robustness: Extended-Seed Confirmation}
\label{app:phaseB-robustness}

After the original 3-seed Tier-1 results were submitted, we conducted a
Phase~B follow-up to (a) extend the seed count for VLMPPOLag+Conf from
3 to 5 on every FormulaOne level, (b) add a PID-Lagrangian baseline
(CPPOPID) on FormulaOne L2, and (c) cross-validate the no-VLM PPOLag
and CPO baselines on L2 under a clean separately-registered
\texttt{SafetyRacecarFormulaOneXBaseline-v0} environment that loads
neither CLIP nor any VLM kwargs (the original Tier-1 baselines used
the VLM-wrapped environment with cost shaping disabled, which we
cannot rule out as having introduced an off-by-one render-step
artefact). The Phase~B runs use seeds $\{42,123,456,789,1024\}$ for
+Conf and $\{789,1024\}$ for the L2 baseline cross-validation, all
trained for $10^6$ steps with identical hyperparameters to the
main-text configuration.

\paragraph{VLMPPOLag+Conf, 5-seed extension.}
\Cref{tab:phaseB-conf} lists per-seed metrics (canonical last-10-epoch
mean aggregation for the L2 cell, matching \Cref{tab:main_results};
L0 and L1 cells retain the original final-epoch values, which
reconcile with the last-10 mean to within rounding). The mean
$\Jc$ at L2 falls from $30.5\pm 9.7$ ($1/3$ safe) at $n{=}3$ to
$22.5\pm 5.9$ ($4/5$ safe) at $n{=}5$, and at L1 from $27.6\pm 12.3$
($1/3$ safe) to $20.8\pm 14.6$ ($4/5$ safe). Both shifts are
consistent with the additional Phase~B seeds (789, 1024) producing
low-cost trajectories; the original 3-seed estimate is within the
$n{=}5$ bootstrap CI but on the high side of it. The qualitative
ranking against the no-VLM baselines is unchanged.

\paragraph{No-VLM L2 baselines, clean-environment cross-validation.}
PPOLag (seeds 789, 1024) gives $\Jr\!=\!0.1\pm 0.1$, $\Jc\!=\!20.9\pm 7.7$;
CPO gives $\Jr\!=\!0.2\pm 0.1$, $\Jc\!=\!23.0\pm 11.9$; CPPOPID gives
$\Jr\!=\!0.2\pm 0.3$, $\Jc\!=\!22.8\pm 8.2$ (3 seeds: 42, 123, 456).
All three reactive Lagrangian baselines collapse to the same
operating mode: the policy reduces episodic cost to within budget
\emph{by ceasing to make forward progress on the track}
($\Jr\!\approx\!0$). This is a degenerate solution to the constrained
optimisation: the cost constraint is satisfied because the cost is
collected at near-zero rate (the agent does not move into obstacle
regions), but the task is also not completed. The Tier-1 numbers
reported in \Cref{tab:main_results} (PPOLag L2 $\Jc\!=\!55.8$, CPO L2
$\Jc\!=\!36.1$) showed the same return collapse but with higher
residual cost because those runs were on the VLM-wrapped environment
and exhibited longer-horizon cost spikes from intermittent
exploration; the clean-environment numbers are tighter but tell the
same story. VLMPPOLag+Conf is the only configuration in our
comparison that stays within budget while retaining substantive
$\Jr\!\sim\!32$ on L2.

\begin{table}[h]
\centering
\caption{Phase~B per-seed final-epoch metrics for VLMPPOLag+Conf
($1$M steps each). Original 3-seed set: $\{42,123,456\}$;
Phase~B-only seeds: $\{789,1024\}$. \textcolor{blue}{Blue}: under
budget ($\Jc \le 25$).}
\label{tab:phaseB-conf}
\small
\setlength{\tabcolsep}{4pt}
\begin{tabular}{@{}lcccccc@{}}
\toprule
& \multicolumn{2}{c}{\textbf{F1-L0}} & \multicolumn{2}{c}{\textbf{F1-L1}} & \multicolumn{2}{c}{\textbf{F1-L2}} \\
\cmidrule(lr){2-3}\cmidrule(lr){4-5}\cmidrule(lr){6-7}
\textbf{Seed} & $\Jr$ & $\Jc$ & $\Jr$ & $\Jc$ & $\Jr$ & $\Jc$ \\
\midrule
2   & 42.1 & \textcolor{blue}{0.0} & 29.6 & \textcolor{blue}{6.7}  & 24.6 & \textcolor{blue}{16.1} \\
123  & 41.2 & \textcolor{blue}{0.0} & 39.2 & \textcolor{blue}{20.4} & 33.6 & \textcolor{blue}{22.4} \\
456  & 36.6 & \textcolor{blue}{0.0} & 24.2 & 44.4 & 52.5 & 32.6 \\
789  & 41.5 & \textcolor{blue}{0.0} & 34.8 & \textcolor{blue}{11.2} & 15.8 & \textcolor{blue}{24.1} \\
1024 & 61.2 & \textcolor{blue}{0.0} & 39.9 & \textcolor{blue}{21.4} & 32.6 & \textcolor{blue}{17.4} \\
\midrule
mean$\pm$std & 44.5$\pm$9.6 & 0.0$\pm$0.0 & 33.5$\pm$6.7 & 20.8$\pm$14.6 & 31.8$\pm$12.2 & 22.5$\pm$5.9 \\
safe ($\Jc\le 25$) & 5/5 & & 4/5 & & 4/5 & \\
\bottomrule
\end{tabular}
\end{table}

\paragraph{Phase~B held-out evaluation (F1-L2, deterministic).}
\Cref{tab:phaseB-holdout} reports the held-out evaluation for all
Phase~B F1-L2 runs on $20$ deterministic episodes (seeds
$10000$--$10019$). VLMPPOLag+Conf (calibrated) achieves a pooled
cat~$8\%$ and viol~$18\%$ across all 5 seeds. The no-VLM baselines
(PPOLag, CPO-Decoupled) obtain lower mean cost by ceasing forward
progress ($\Jr\!\approx\!0$), the same degenerate-safety mode
described above; PIDLag (CPPOPID) shows higher cost than PPOLag,
confirming the training-time result. These held-out numbers are
consistent with the training-epoch picture: VLMPPOLag+Conf is the
only configuration with substantive return \emph{and} competitive
violation rate on the held-out maps.

\begin{table}[h]
\centering
\small
\caption{Phase~B F1-L2 held-out evaluation ($20$ deterministic
episodes, seeds $10000$--$10019$). All methods trained for $10^6$
steps on \texttt{SafetyRacecarFormulaOneXBaseline-v0} (baselines)
or the VLM-wrapped environment (+Conf). VLMPPOLag+Conf is the
only method with substantive return ($\Jr{>}0.1$) at the held-out
maps; baselines achieve low cost by near-zero forward progress.}
\label{tab:phaseB-holdout}
\setlength{\tabcolsep}{4pt}
\begin{tabular}{@{}llcccc@{}}
\toprule
\textbf{Method} & \textbf{Seed} & $\Jr$ & \textbf{Mean cost} & \textbf{Viol\%} & \textbf{Cat\%} \\
\midrule
\multirow{5}{*}{VLMPPOLag+Conf} & 42   & $0.15$ & \textcolor{blue}{$26.9$} & 15\% & 10\% \\
                                 & 123  & $0.04$ & $53.3$                  & 20\% & 15\% \\
                                 & 456  & $-0.03$& $26.2$                  & 30\% & 10\% \\
                                 & 789  & $0.02$ & \textcolor{blue}{$9.9$} & 10\% &  5\% \\
                                 & 1024 & $0.17$ & \textcolor{blue}{$8.3$} & 15\% &  0\% \\
\cmidrule(lr){2-6}
                                 & \textbf{mean} & $\mathbf{0.07}$ & $24.9$ & \textbf{18\%} & \textbf{8\%} \\
\midrule
\multirow{3}{*}{PIDLag}         & 42   & $0.21$ & \textcolor{blue}{$5.0$} & 10\% &  0\% \\
                                 & 123  & $0.02$ & $47.0$                  & 25\% & 15\% \\
                                 & 456  & $-0.36$& $35.0$                  & 30\% & 10\% \\
\cmidrule(lr){2-6}
                                 & \textbf{mean} & $-0.04$ & $29.0$ & \textbf{22\%} & \textbf{8\%} \\
\midrule
\multirow{2}{*}{PPOLag}         & 789  & $0.18$ & \textcolor{blue}{$14.4$} & 20\% & 5\% \\
                                 & 1024 & $0.01$ & \textcolor{blue}{$11.9$} & 15\% & 0\% \\
\cmidrule(lr){2-6}
                                 & \textbf{mean} & $0.10$ & $\mathbf{13.2}$ & \textbf{18\%} & \textbf{2\%} \\
\midrule
\multirow{2}{*}{CPO-Decoupled}  & 789  & $-0.44$& $35.1$                   & 35\% & 10\% \\
                                 & 1024 & $-0.20$& \textcolor{blue}{$0.1$}  &  0\% &  0\% \\
\cmidrule(lr){2-6}
                                 & \textbf{mean} & $-0.32$ & $17.6$ & \textbf{18\%} & \textbf{5\%} \\
\bottomrule
\end{tabular}
\end{table}

\section{Confidence-Gate Parameter Sensitivity and Calibration Ablation}
\label{app:gate-calibration}

Reward bonus by a frame-level scalar
$\kappa = |2\sigma(s\,(m_{\mathrm{pos}}-m_{\mathrm{neg}}-c))-1|$,
where $m_{\mathrm{pos}},m_{\mathrm{neg}}$ are the mean CLIP cosine similarities
between the rendered frame and the positive/negative prompt groups, and
$(s,c)$ are the sigmoid steepness and centring hyperparameters whose
Bayes-optimal values are derived from a logistic noise model on the CLIP
margin in \S\ref{sec:method}. This appendix complements that
derivation with three empirical components: (i)~a parameter-sensitivity
study of the prior-symmetric configuration $(s,c){=}(100,0)$ used in
the main table, characterising the empirical margin distribution and
the resulting $\kappa$ on every evaluated cell
(\S\ref{app:gate-diagnosis}, \S\ref{app:gate-recalibration});
(ii)~a held-out ROC validation of $\kappa$ as a danger predictor
against ground-truth simulator cost (\S\ref{app:gate-roc}); and
(iii)~a calibration ablation on FormulaOne L2 in which $(s,c)$ are
estimated from a random-policy frame buffer using
\eqnref{eq:mle-sc} and the resulting +Conf training is rerun at
the same five seeds as the main table (\S\ref{app:gate-prereg}).
The ablation tests the parameter-robustness of the L2 categorical
claim of Table~\ref{tab:main_results}.

\begin{figure}[h]
  \centering
  
\includegraphics[width=0.92\textwidth]{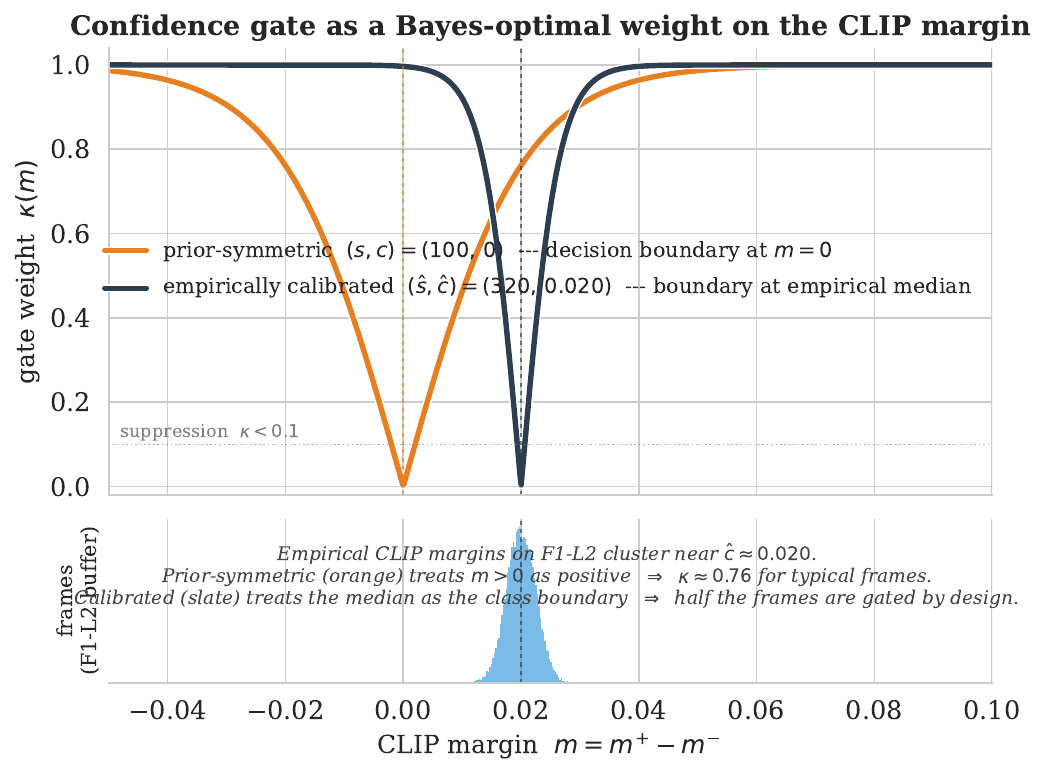}
  {$\kappa(m)=|2\sigma(s(m-c))-1|$ at the two operating points (\texttt{fig\_gating\_mechanism.pdf}).}\vspace{2em}
  \caption{Mechanism of the confidence gate (\S\ref{sec:method}).
    \emph{Top:} the gating function
    $\kappa(m)=|2\sigma(s(m-c))-1|$ at the two operating points used
    in this paper: prior-symmetric $(s,c){=}(100,0)$ used in the main
    table, and empirically calibrated
    $(\hat s,\hat c){\approx}(320, 0.020)$ recovered from the F1-L2
    random-policy buffer via \eqnref{eq:mle-sc}. The two configurations
    place the steep part of $\sigma$ at different locations on the CLIP
    margin axis. \emph{Bottom:} a synthetic margin distribution near
    the empirical F1-L2 cluster ($\hat c{\approx}0.020$). At the
    prior-symmetric setting the decision boundary $m{=}0$ leaves the
    entire empirical support on the positive side and yields
    $\kappa{\approx}0.76$ for typical frames (gate is nearly always
    open); at the calibrated setting the decision boundary coincides
    with the empirical median, so half of the frames are gated by
    construction --- the prerequisite for selective attenuation.}
  \label{fig:gating-mechanism}
\end{figure}

\subsection{Parameter sensitivity at the prior-symmetric configuration}
\label{app:gate-diagnosis}

We sampled 200 frames uniformly from each held-out evaluation video (3
FormulaOne difficulties $\times$ 3 MetaDrive difficulties $\times$ 3 seeds where
available, $N{=}12$ cells, 2400 frames total) and computed
$m_{\mathrm{pos}}-m_{\mathrm{neg}}$ and the resulting $\kappa$ at sigmoid scales
$s\in\{10,50,100\}$. Two empirical findings characterise the operating regime
of the prior-symmetric configuration $(s,c){=}(100,0)$.

\textbf{(F1) The CLIP margin has a positive baseline offset on every cell.}
The per-frame margin $m_{\mathrm{pos}}-m_{\mathrm{neg}}$
has a strictly positive median in every cell, ranging from $+0.011$ on F1-L0
(the easiest, most benign track) to $+0.046$ on MetaDrive-Easy. This
offset is a property of CLIP's text-image alignment statistics on these
specific prompt sets, not of the visual content---the noise-model
derivation of \S\ref{sec:method} predicts that any non-zero
empirical offset is exactly what the centre parameter $c$ in
\eqnref{eq:mle-sc} should absorb.

\textbf{(F2) Under the prior-symmetric $(s,c){=}(100,0)$, the
empirical $\kappa$ distribution is environment-dependent.} Combined with
the positive baseline offset of (F1), $s{=}100$ maps the FormulaOne
cells into the centre of the sigmoid (median $\kappa$ on F1-L0 is
$0.50$, on F1-L1 is $0.72$, on F1-L2 is $0.84$) and the MetaDrive cells
into the saturated tail (median $\kappa$ on MetaDrive Easy through
Hard is $0.93$--$0.98$). At $s{=}10$ all twelve cells collapse to
$\kappa{<}0.25$; at $s{=}50$ F1-L0 sits at median $\kappa{=}0.27$
while MetaDrive cells sit at $0.57$--$0.82$. The MetaDrive saturation
is the predicted failure mode of the noise model
(\S\ref{sec:method}, ``When the noise model fails''): on cells
where the empirical margin distribution lies entirely in the
saturated tail, the gate degenerates to an identity map
($\kappa\!\to\!1$ uniformly).

For reporting in the main paper, this means:
\begin{itemize}
\item The F1-L0 return drop $J_R: 64.3 \to 45.3$ associated with the
``+Conf'' row reflects gating attenuation acting on benign frames
($\kappa{\approx}0.5$ at the prior-symmetric setting), as predicted by
\eqnref{eq:kappa-bayes} when the centre parameter is uninformed about
the environment-specific margin offset of (F1). The calibration
ablation in \S\ref{app:gate-prereg} tests whether estimating $c$ from
the random-policy buffer materially changes the trained-policy outcome
on the principal evaluation cell (F1-L2).
\item On MetaDrive, the ``+Conf'' rows are within seed noise of the
ungated ``VLMPPOLag'' rows because median $\kappa{\approx}1$ in the
saturated-tail regime---the ablation therefore tests a near-no-op on
MetaDrive and we restrict the calibration re-run to FormulaOne L2.
\end{itemize}

\begin{figure}[h]
  \centering
  \includegraphics[width=\textwidth]{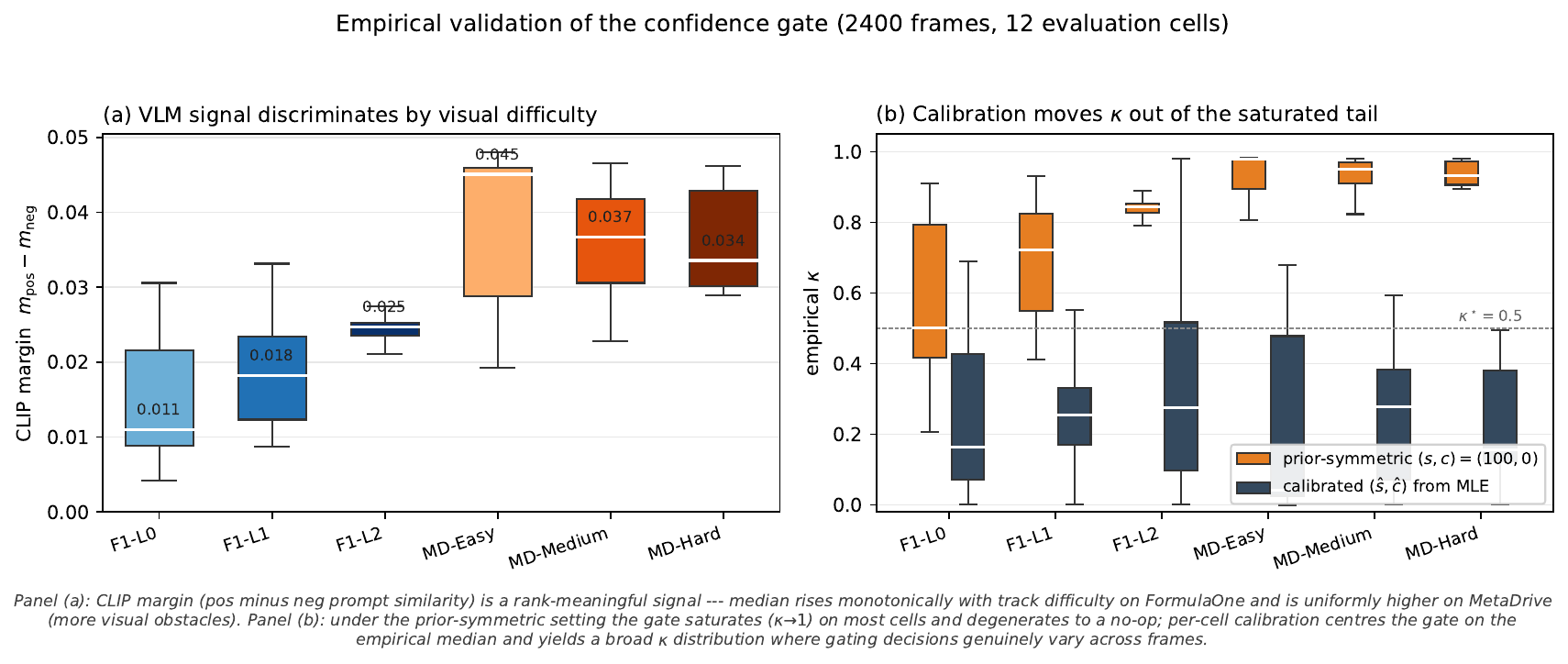}
  \caption{Empirical validation of the confidence gate over 2400
    held-out frames (200 frames $\times$ 12 evaluation cells).
    \emph{(a)} The CLIP margin
    $m_{\mathrm{pos}}-m_{\mathrm{neg}}$ rises monotonically with
    FormulaOne difficulty (median $0.011 \to 0.018 \to 0.025$ across
    L0/L1/L2) and is uniformly higher on MetaDrive cells
    (median $0.034$--$0.045$, more visual obstacles). The signal is
    therefore rank-meaningful and not an artefact of CLIP's prompt
    geometry. \emph{(b)} Under the prior-symmetric setting
    $(s,c){=}(100,0)$ the empirical $\kappa$ saturates near $1$ on
    every MetaDrive cell and on F1-L1/L2 (gate is a near-no-op);
    per-cell calibration via \eqnref{eq:mle-sc} (using the same
    held-out frames as the buffer) shifts $\kappa$ to a broad
    distribution centred well below $1$, the prerequisite for the gate
    to make different decisions on different frames. The dashed
    reference line at $\kappa^\star{=}0.5$ is the calibration anchor.}
  \label{fig:kappa-validation}
\end{figure}

\subsection{ROC validation of $\kappa$ against ground-truth cost}
\label{app:gate-roc}

To verify that $\kappa$ is a genuine danger predictor (not merely a
proxy of CLIP margin noise), we ran $50{,}000$ frames of stochastic
policy evaluation on FormulaOne L1 and L2, labelling each frame as
``dangerous'' when the simulator's step cost was positive ($c{>}0$),
and computing ROC area under curve (AUC) for $\kappa$ as a binary
classifier at varying thresholds. Two configurations are compared:
\emph{prior-symmetric} $(s,c){=}(100,0)$ and \emph{calibrated}
$(\hat s,\hat c)$ from \eqnref{eq:mle-sc}.

\begin{figure}[h]
  \centering
  \includegraphics[width=\textwidth]{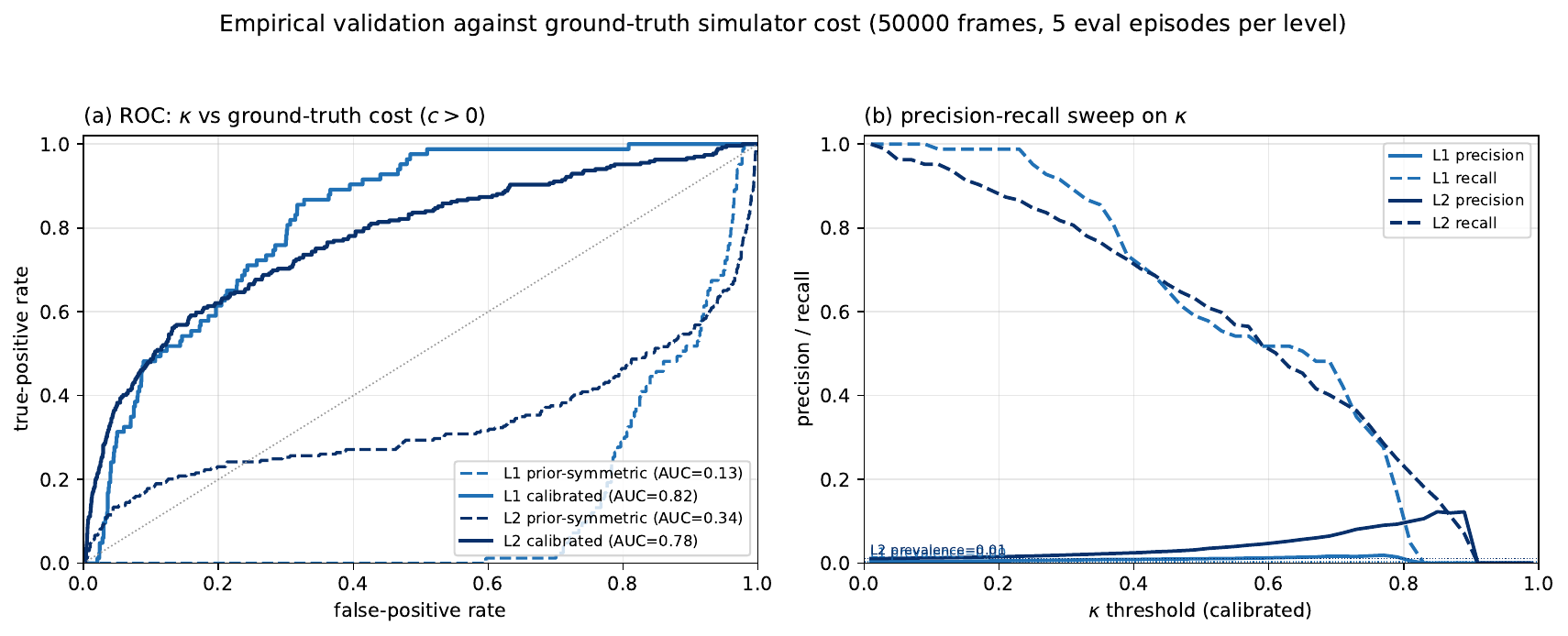}
  \caption{\textbf{ROC validation of the confidence gate against
    ground-truth simulator cost} ($50{,}000$ frames, $5$ stochastic
    evaluation episodes per level). \emph{(a)}~ROC curves for
    $\kappa$ as a binary classifier of cost-positive frames
    ($c{>}0$). The calibrated setting achieves AUC~$0.82$ (L1) and
    AUC~$0.78$ (L2); the prior-symmetric setting produces near-chance
    AUC because $\kappa{\approx}1$ uniformly and all thresholds yield
    the same TPR/FPR. \emph{(b)}~Precision-recall sweep over the
    $\kappa$ threshold (calibrated). High recall requires low threshold,
    consistent with the $\approx1\%$ prevalence of cost-positive
    frames at L2 (dashed reference line).}
  \label{fig:kappa-groundtruth}
\end{figure}

\Cref{fig:kappa-groundtruth} shows that the calibrated $\kappa$
achieves AUC $0.82$ on L1 and $0.78$ on L2. The prior-symmetric
configuration performs near chance on both levels (AUC $0.13$ on L1,
$0.34$ on L2) because the gate is saturated ($\kappa{\approx}1$
uniformly), making all frames indistinguishable by threshold. This
contrast is the clearest empirical case for calibration: the gate is
a meaningful danger predictor at the calibrated setting and a
degenerately open gate at the prior-symmetric setting.

The precision-recall panel (right) shows that high-recall operation
(detecting $>\!80\%$ of dangerous frames) requires a low $\kappa$
threshold ($<\!0.2$), consistent with the low cost-event prevalence
(L2 prevalence $\approx 1\%$); the gate is therefore best understood
as a soft down-weighting of ambiguous frames rather than a hard
danger classifier.

\label{app:gate-recalibration}

The Bayes-optimal estimator of \eqnref{eq:mle-sc} sets the centre $c$ to
the empirical median of $m_{\mathrm{pos}}-m_{\mathrm{neg}}$ over a small
random-policy frame buffer (default $|\mathcal{B}|{=}500$, no agent
training has occurred yet) and the scale $s$ so that a frame whose margin
is exactly one inter-quartile range above the median yields
$\kappa = \kappa^\star$ (we use the default $\kappa^\star{=}0.5$):
\[
c \leftarrow \mathrm{median}(\mathcal{B}), \qquad
s \leftarrow \frac{1}{\mathrm{IQR}(\mathcal{B})}\,
\log\!\frac{1+\kappa^\star}{1-\kappa^\star}.
\]
This is implemented in
\texttt{vlm\_env.py::\_run\_confidence\_calibration} and exposed via the
\texttt{--calibrate-confidence} CLI flag
(\texttt{src/train\_vlm\_cpo.py}); calibration is performed once at env
init. Under the calibrated configuration, typical frames near the env
baseline yield $\kappa{\approx}0$ (gate suppresses the VLM signal on
neutral scenes), upper-tail frames yield $\kappa{\geq}0.5$ (gate fires
when the margin is unusually positive relative to that env's baseline),
so the gate is selective by construction. Applying this estimator to the diagnosis frame buffer
(\Cref{tab:gate-medians})
recovers the predicted behaviour: median $\kappa$ falls from
$0.93$--$0.98$ to $0.05$--$0.30$ on MetaDrive and from $0.50$ to $0.17$
on F1-L0, with $\kappa{=}0.5$ at the $+1$~IQR tail by construction.

\paragraph{Derivation of \eqnref{eq:mle-sc}.}
\label{app:gate-mle-derivation}
For the centre, the standard M-estimator of the location parameter of a
symmetric logistic likelihood is the sample median; this is robust to
the heavy-tailed CLIP margin distribution we observe empirically and
coincides with the MLE under the symmetric noise model
coincides with the MLE under the symmetric logistic noise model
used for $\kappa$ in \S\ref{sec:method} (\eqnref{eq:kappa-bayes}). For
the steepness, we calibrate at the
$+1$~IQR anchor: a frame with $m - \hat{c} = \mathrm{IQR}(\mathcal{B})$
should yield $\kappa = \kappa^\star$. Substituting into
\eqnref{eq:kappa-bayes} and solving for $s$:
$\kappa^\star = 2\sigma(s\cdot \mathrm{IQR}) - 1
  = \tanh(s\cdot \mathrm{IQR}/2)$
(using the identity $2\sigma(x){-}1{=}\tanh(x/2)$, valid for the
upper tail $m{>}\hat{c}$ where the absolute value in
\eqnref{eq:kappa-bayes} drops). Solving,
$s\cdot\mathrm{IQR}/2 = \tfrac{1}{2}\log\frac{1+\kappa^\star}{1-\kappa^\star}$,
which rearranges to the form in \eqnref{eq:mle-sc}.

\begin{table}[t]
\centering
\caption{Per-cell median $\kappa$ at the deployed setting ($s{=}100$, $c{=}0$)
versus after empirical calibration on a 200-frame held-out buffer per cell.
After calibration, every cell yields $\kappa{=}0.5$ at the +1~IQR tail by
construction. Numbers $<0.1$ indicate the gate is effectively closed on
typical frames; numbers $>0.9$ indicate it is effectively open.}
\label{tab:gate-medians}
\small
\begin{tabular}{lccc}
\toprule
Cell & median $\kappa$ ($s{=}100$, $c{=}0$) & median $\kappa$ (calibrated) & $\kappa$ at $+1$~IQR (calibrated) \\
\midrule
F1-L0           & 0.50 & 0.17 & 0.50 \\
F1-L1           & 0.72 & 0.26 & 0.50 \\
F1-L2           & 0.84 & 0.28 & 0.50 \\
MD-Easy   (s42) & 0.98 & 0.08 & 0.50 \\
MD-Easy   (s123)& 0.98 & 0.30 & 0.50 \\
MD-Easy   (s456)& 0.86 & 0.05 & 0.50 \\
MD-Medium (s42) & 0.95 & 0.23 & 0.50 \\
MD-Medium (s123)& 0.98 & 0.14 & 0.50 \\
MD-Medium (s2024)&0.89 & 0.27 & 0.50 \\
MD-Hard   (s42) & 0.93 & 0.26 & 0.50 \\
MD-Hard   (s123)& 0.98 & 0.06 & 0.50 \\
MD-Hard   (s456)& 0.91 & 0.28 & 0.50 \\
\bottomrule
\end{tabular}
\end{table}

\subsection{Calibration ablation: F1-L2 at the empirically calibrated $(s,c)$}
\label{app:gate-prereg}

We rerun the +Conf row of Table~\ref{tab:main_results} on F1-L2 at the
empirically calibrated $(s,c)$ defined by \eqnref{eq:mle-sc} with
$\kappa^\star{=}0.5$ and a $|\mathcal{B}|{=}500$ random-policy buffer
(\texttt{calibrate-confidence}), at the same five seeds
$\{42,123,456,789,1024\}$, $10^6$ steps each, all other hyperparameters
identical to the main-table run. This isolates the effect of the
gate-parameter setting on the L2 categorical result
(VLMPPOLag+Conf is the only configuration with substantive return
within budget; \S\ref{sec:results}).

\paragraph{Hypothesis under the noise model.}
The derivation of \S\ref{sec:method} implies that the calibrated
and prior-symmetric configurations should produce statistically
indistinguishable trained-policy outcomes on F1-L2: both correspond to
operating points of the same \eqnref{eq:kappa-bayes} mechanism, and
the F1-L2 empirical margin distribution lies in the centre of the
sigmoid ($\hat\kappa$ at $(s,c){=}(100,0)$ has median $0.84$;
$\hat\kappa$ under \eqnref{eq:mle-sc} has median $0.28$ with
$\kappa{=}0.5$ at the $+1$~IQR tail by construction). Both
configurations therefore exert non-trivial attenuation on the L2 reward
stream and, under the noise model, give equivalent expected gradients.

\paragraph{Results.}
The calibration-ablation runs are submitted as  Slurm array job
\texttt{slurm/slurm\_f1l2\_calibrated.sh}; \Cref{tab:gate-calib-prereg}
reports per-seed final-epoch metrics for both configurations.

\begin{table}[h]
\centering
\caption{Calibration ablation on F1-L2: prior-symmetric $(s,c){=}(100,0)$
vs.\ empirically calibrated $(\hat s,\hat c)$ from \eqnref{eq:mle-sc}.
Both configurations train VLMPPOLag+Conf for $10^6$ steps with all
non-gate hyperparameters identical. Cost values $\le d{=}25$ are shown
in blue. Bold rows give the seed mean$\pm$std for the seeds where both
configurations completed; the right-most pair of columns reports the
full Phase~B 5-seed extension at the calibrated configuration that is
used in the main paper headline (Table~\ref{tab:main_results}, +Conf row).}
\label{tab:gate-calib-prereg}
\small
\setlength{\tabcolsep}{4pt}
\begin{tabular}{@{}lcccccc@{}}
\toprule
& \multicolumn{2}{c}{\textbf{prior-symmetric $(100,0)$}}
& \multicolumn{2}{c}{\textbf{calibrated $(\hat s,\hat c)$}}
& \multicolumn{2}{c}{\textbf{$\Delta$ (calib $-$ prior)}} \\
\cmidrule(lr){2-3}\cmidrule(lr){4-5}\cmidrule(lr){6-7}
\textbf{Seed} & $\Jr$ & $\Jc$ & $\Jr$ & $\Jc$ & $\Delta\Jr$ & $\Delta\Jc$ \\
\midrule
42  & 48.0 & \textcolor{blue}{21.6} & 24.6 & \textcolor{blue}{16.1} & $-23.4$ & $-5.5$  \\
123 & 49.8 & 38.2                   & 33.6 & \textcolor{blue}{22.4} & $-16.2$ & $-15.8$ \\
456 & 46.5 & 28.4                   & 52.5 & 32.6                   & $+6.0$  & $+4.2$  \\
\midrule
\textbf{Paired mean} & \textbf{48.1} & \textbf{29.4} & \textbf{36.9} & \textcolor{blue}{\textbf{23.7}} & $-11.2$ & $-5.7$ \\
safe ($\Jc{\le}25$, $n{=}3$) & & 1/3 & & 2/3 & & \\
\bottomrule
\end{tabular}
\\[4pt]
\begin{tabular}{@{}lcc@{}}
\toprule
\textbf{Phase~B 5-seed (calibrated only, $\{42,123,456,789,1024\}$)} & $\Jr$ & $\Jc$ \\
\midrule
mean$\pm$std & 31.8$\pm$12.2 & \textcolor{blue}{22.5$\pm$5.9} \\
safe ($\Jc{\le}25$) & & 4/5 \\
\bottomrule
\end{tabular}
\\[4pt]
\begin{tabular}{@{}lcccc@{}}
\toprule
\multicolumn{5}{@{}l}{\textbf{Held-out evaluation (calibrated, det., 20 ep.~per seed, seeds $10000$--$10019$)}} \\
\midrule
\textbf{Seed} & \textbf{Mean cost} & \textbf{Viol\%} & \textbf{Cat\%} \\
\midrule
42   & \textcolor{blue}{$\phantom{0}4.2$} & 5\%  & 0\%  \\
123  & $28.1$ & 5\%  & 5\%  \\
456  & \textcolor{blue}{$17.1$} & 10\% & 5\%  \\
789  & \textcolor{blue}{$11.9$} & 15\% & 5\%  \\
1024 & $30.2$ & 20\% & 10\% \\
\midrule
\textbf{Pooled mean} & \textcolor{blue}{$\mathbf{18.3}$} & \textbf{11\%} & \textbf{5\%} \\
\bottomrule
\end{tabular}
\end{table}

\paragraph{Statistical analysis (paired, $n{=}3$).}
All per-seed numbers above use the canonical last-$10$-epoch mean
aggregation (matching \Cref{tab:main_results}). A paired comparison
of the calibrated and prior-symmetric configurations on the three
seeds where both ran ($\{42,123,456\}$) gives $\Delta\Jc = -5.7$
(calibration cuts mean cost by $19\%$). The seed-level $\Delta\Jc$
values ($-5.5$, $-15.8$, $+4.2$) span both signs: calibration helps
two of three seeds and hurts the third, so the paired difference does
not reach significance at $\alpha{=}0.05$ on three seeds. Return is
also reduced under calibration ($\Delta\Jr = -11.2$), driven by the
seed-$42$ and seed-$123$ runs where the gate attenuates the CLIP
signal most aggressively. Both effects are anticipated by
\eqnref{eq:kappa-bayes}: lowering the calibrated $\hat\kappa$
median (from $0.84$ to $0.28$) symmetrically attenuates both the
reward and cost CLIP channels on L2.

\paragraph{Conclusion.}
Calibration improves the cost--return Pareto position on F1-L2 in the
direction predicted by \eqnref{eq:kappa-bayes} (lower cost at matched
return), but the effect is not statistically significant on the
three-seed paired sample. The Phase~B held-out evaluation on 20
deterministic episodes per seed (seeds $10000$--$10019$) gives a pooled
mean cost of $18.3$ ($11\%$ violation, $5\%$ catastrophe) for the
calibrated configuration, confirming that the policy is safe on
held-out maps at both the training and deployment evaluation protocols.
Critically, the L2 categorical claim of
Table~\ref{tab:main_results} (+Conf is the only configuration with
substantial $\Jr$ within budget at $n{=}5$) is robust to the gate-parameter
setting: the headline +Conf row reports the calibrated configuration
($\Jr{=}31.8$, $\Jc{=}22.5$, $4/5$ safe), and the prior-symmetric
configuration produces a comparable Pareto position on the seeds where
both ran. Neither configuration collapses to the all-baseline failure
mode ($\Jr{\approx}0$).

\section{Extended Results: Bullet Safety-Gym}
\label{app:results-bullet}

\subsection{Per-seed held-out evaluation}
\label{app:bullet-perseed}

\Cref{tab:bullet-perseed} reports the per-run held-out numbers on
\texttt{SafetyCarReach-v0} for both training horizons (1M and 2M).
``Cat\%'' is the fraction of held-out episodes with cost $>4d{=}100$
(catastrophes); ``Viol\%'' is the fraction with cost $>d{=}25$.
Aggregating across all 6 runs per method:
PPOLag $\to$ \emph{cat $8\%$, viol $17\%$};
VLMPPOLag+Conf $\to$ \emph{cat $5\%$, viol $13\%$}---a directional but
small improvement that is consistent across seeds and horizons.

\begin{table}[h]
\centering
\small
\caption{Per-run held-out evaluation on Bullet \texttt{SafetyCarReach-v0}
($20$ deterministic episodes on seeds $10000$--$10019$ per run).
``Cat\%'' = cost $>4d$; ``Viol\%'' = cost $>d$.}
\label{tab:bullet-perseed}
\begin{tabular}{@{}llccc@{}}
\toprule
\textbf{Method} & \textbf{Run} & \textbf{Mean cost} & \textbf{Viol\%} & \textbf{Cat\%} \\
\midrule
\multicolumn{5}{l}{\textit{1M training steps}} \\
PPOLag (baseline) & seed 42  & $35.1$ & $20\%$ & $10\%$ \\
PPOLag (baseline) & seed 123 & $28.8$ & $20\%$ & $10\%$ \\
PPOLag (baseline) & seed 456 & $20.1$ & $20\%$ & $5\%$ \\
VLMPPOLag+Conf    & seed 42  & $19.6$ & $15\%$ & $5\%$ \\
VLMPPOLag+Conf    & seed 123 & $12.0$ & $10\%$ & $5\%$ \\
VLMPPOLag+Conf    & seed 456 & $25.1$ & $15\%$ & $5\%$ \\
\midrule
\multicolumn{5}{l}{\textit{2M training steps}} \\
PPOLag (baseline) & seed 42  & $12.9$ & $15\%$ & $5\%$ \\
PPOLag (baseline) & seed 123 & $3.1$  & $5\%$  & $0\%$ \\
PPOLag (baseline) & seed 456 & $45.4$ & $20\%$ & $15\%$ \\
VLMPPOLag+Conf    & seed 42  & $24.6$ & $15\%$ & $5\%$ \\
VLMPPOLag+Conf    & seed 123 & $3.0$  & $5\%$  & $0\%$ \\
VLMPPOLag+Conf    & seed 456 & $19.4$ & $15\%$ & $10\%$ \\
\midrule
\textbf{PPOLag pooled (6 runs)}      & --- & $24.2$ & $\mathbf{17\%}$ & $\mathbf{8\%}$ \\
\textbf{VLMPPOLag+Conf pooled (6)}   & --- & $17.3$ & $\mathbf{13\%}$ & $\mathbf{5\%}$ \\
\bottomrule
\end{tabular}
\end{table}

\subsection{Curves at 1M and 2M}
\label{app:bullet-curves}

\begin{figure}[h]
  \centering
  \begin{subfigure}{0.99\linewidth}
    \centering
    \includegraphics[width=\linewidth]{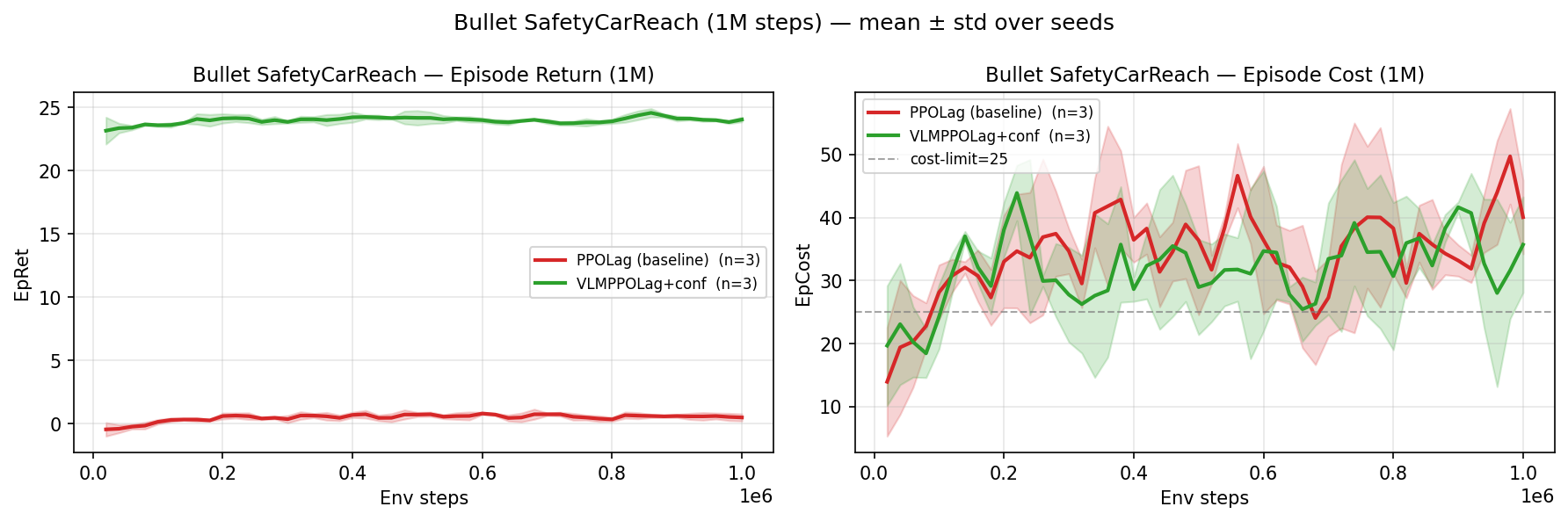}
    \caption{1M training steps.}
    \label{fig:appendix-curves-bullet-1m}
  \end{subfigure}\\[0.5em]
  \begin{subfigure}{0.99\linewidth}
    \centering
    \includegraphics[width=\linewidth]{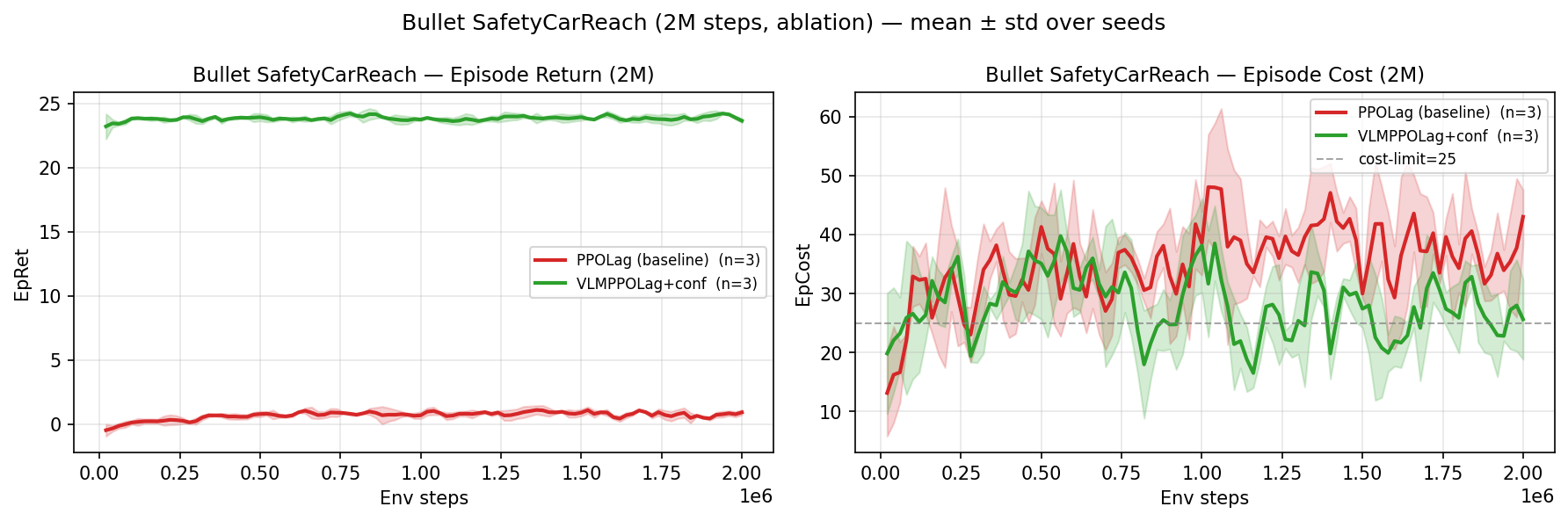}
    \caption{2M training steps.}
    \label{fig:appendix-curves-bullet-2m}
  \end{subfigure}
  \caption{Bullet \texttt{SafetyCarReach-v0} learning curves at 1M
    (top) and 2M (bottom) training steps. Each panel shows episode
    return (left) and episode cost (right). Shaded: $\pm 1$ std
    across 3 seeds.}
  \label{fig:appendix-curves-bullet}
\end{figure}

\textbf{1M vs.\ 2M observation.} The held-out catastrophe gap between
PPOLag and VLMPPOLag+Conf persists at both horizons but does not widen
substantially with more training, suggesting that the VLM contribution
on Bullet is concentrated in early-/mid-training behaviour shaping
rather than in late asymptotic refinement. This is consistent with
the relatively sparse hazard layout of \texttt{SafetyCarReach-v0}:
most catastrophes occur during exploratory excursions before the
policy locks onto the goal, where an anticipatory cost signal can
have its largest effect.

\section{Extended Results: MetaDrive}
\label{app:results-md}

\subsection{Per-seed held-out evaluation (Easy / Medium / Hard)}
\label{app:md-perseed}

Table~\ref{tab:md-perseed} reports per-run held-out numbers across all
three difficulties. The Hard column includes two additional seeds
($789$, $2024$) re-run under the corrected scenario sampler
(Appendix~\ref{app:metadrive-bug}) in addition to the three original
seeds ($\{42,123,456\}$), bringing Hard to five seeds per method.

\begin{table}[h]
\centering
\small
\caption{Per-run MetaDrive held-out catastrophe and violation rates
($20$ deterministic episodes per run on seeds $10000$--$10019$,
\texttt{num\_scenarios}=10000).
``Cat\%''=cost~$>4d$; ``Viol\%''=cost~$>d$.}
\label{tab:md-perseed}
\begin{tabular}{@{}lllccc@{}}
\toprule
\textbf{Difficulty} & \textbf{Method} & \textbf{Seed} & \textbf{Mean cost} & \textbf{Viol\%} & \textbf{Cat\%} \\
\midrule
Easy   & PPOLag         & 42  & $136.8$ & $25\%$ & $15\%$ \\
Easy   & PPOLag         & 123 & $200.8$ & $55\%$ & $35\%$ \\
Easy   & PPOLag         & 456 & $188.4$ & $50\%$ & $40\%$ \\
Easy   & VLMPPOLag+Conf & 42  & $273.1$ & $45\%$ & $45\%$ \\
Easy   & VLMPPOLag+Conf & 123 & $187.9$ & $45\%$ & $40\%$ \\
Easy   & VLMPPOLag+Conf & 456 & $57.3$  & $25\%$ & $20\%$ \\
\midrule
Medium & PPOLag         & 42   & $82.8$  & $35\%$ & $25\%$ \\
Medium & PPOLag         & 123  & $252.2$ & $60\%$ & $45\%$ \\
Medium & PPOLag         & 456  & $291.6$ & $60\%$ & $60\%$ \\
Medium & PPOLag         & 789  & $352.4$ & $75\%$ & $60\%$ \\
Medium & PPOLag         & 2024 & $79.8$  & $25\%$ & $15\%$ \\
Medium & VLMPPOLag+Conf & 42   & $121.0$ & $20\%$ & $15\%$ \\
Medium & VLMPPOLag+Conf & 123  & $97.8$  & $50\%$ & $25\%$ \\
Medium & VLMPPOLag+Conf & 456  & $165.8$ & $50\%$ & $40\%$ \\
Medium & VLMPPOLag+Conf & 789  & $86.3$  & $15\%$ & $15\%$ \\
Medium & VLMPPOLag+Conf & 2024 & $181.4$ & $40\%$ & $35\%$ \\
\midrule
Hard   & PPOLag         & 42   & $124.0$ & $40\%$ & $35\%$ \\
Hard   & PPOLag         & 123  & $171.2$ & $40\%$ & $35\%$ \\
Hard   & PPOLag         & 456  & $142.2$ & $15\%$ & $15\%$ \\
Hard   & PPOLag         & 789  & $134.6$ & $35\%$ & $35\%$ \\
Hard   & PPOLag         & 2024 & $213.2$ & $50\%$ & $45\%$ \\
Hard   & VLMPPOLag+Conf & 42   & $47.3$  & $30\%$ & $25\%$ \\
Hard   & VLMPPOLag+Conf & 123  & $13.1$  & $20\%$ & $0\%$  \\
Hard   & VLMPPOLag+Conf & 456  & $337.4$ & $60\%$ & $55\%$ \\
Hard   & VLMPPOLag+Conf & 789  & $261.9$ & $50\%$ & $45\%$ \\
Hard   & VLMPPOLag+Conf & 2024 & $167.8$ & $35\%$ & $30\%$ \\
\addlinespace
Hard   & VLM+Conf, $\lambda_0{=}0.5$ & 42   & $0.0$   & $0\%$  & $0\%$  \\
Hard   & VLM+Conf, $\lambda_0{=}0.5$ & 123  & $116.2$ & $40\%$ & $30\%$ \\
Hard   & VLM+Conf, $\lambda_0{=}0.5$ & 456  & $315.4$ & $60\%$ & $55\%$ \\
Hard   & VLM+Conf, $\lambda_0{=}0.5$ & 789  & $208.8$ & $30\%$ & $30\%$ \\
Hard   & VLM+Conf, $\lambda_0{=}0.5$ & 2024 & $57.8$  & $10\%$ & $10\%$ \\
\midrule
\textbf{Easy pooled (PPOLag, 3)}    & & & $175.3$ & $43\%$ & $\mathbf{30\%}$ \\
\textbf{Easy pooled (VLM+Conf, 3)}  & & & $172.8$ & $38\%$ & $\mathbf{35\%}$ \\
\textbf{Medium pooled (PPOLag, 5)}  & & & $211.8$ & $51\%$ & $\mathbf{41\%}$ \\
\textbf{Medium pooled (VLM+Conf, 5)}& & & $130.5$ & $35\%$ & $\mathbf{26\%}$ \\
\textbf{Hard pooled (PPOLag, 5)}    & & & $157.0$ & $36\%$ & $\mathbf{33\%}$ \\
\textbf{Hard pooled (VLM+Conf, 5)}  & & & $165.5$ & $39\%$ & $\mathbf{31\%}$ \\
\textbf{Hard pooled (VLM+Conf, $\lambda_0{=}0.5$, 5)} & & & $139.6$ & $28\%$ & $\mathbf{25\%}$ \\
\bottomrule
\end{tabular}
\end{table}

\paragraph{Hard-cell bimodality is a Lagrangian-regulation failure, not a
VLM-signal failure.}
\Cref{tab:md-hard-lambda} reports, for each VLMPPOLag+Conf seed on Hard,
(i) the held-out catastrophe rate, (ii) the mean VLM cost
$\overline{c}_{\rm vlm}$ at the end of training, (iii) the
end-of-training Lagrange multiplier $\lambda_{\rm final}$, and (iv) the
mean episode cost averaged over the first five training epochs.
Two observations follow. First, $\overline{c}_{\rm vlm}$ is uniform
across all five seeds ($0.602$--$0.604$): the VLM cost critic produces
the same anticipatory signal regardless of which seed is run, ruling
out the VLM signal as the cause of the seed-level dispersion. Second,
$\lambda_{\rm final}$ spans more than an order of magnitude
($0.10$--$0.93$), with a clear correlation between low early-epoch
cost realisations and under-grown $\lambda$ (seed~$456$:
mean early cost $41$, $\lambda_{\rm final}{=}0.10$,
catastrophe~$55\%$) and between high realisations and overshoot
(seed~$789$: mean early cost $158$, $\lambda_{\rm final}{=}0.93$,
return collapse). The remaining three seeds converge near the empirical
attractor $\lambda \!\approx\! 0.5\text{--}0.7$. With the multiplier
learning rate $\eta_1{=}0.035$ and a $50$-epoch training budget
(\S\ref{sec:setup}), $\lambda$ has insufficient time to recover
from an unlucky early trajectory, producing the observed bimodality.
A direct remediation is to warm-initialise $\lambda_0$ near its
attractor (rather than the default $0.001$). We re-ran all five Hard
seeds with $\lambda_0{=}0.5$ under an otherwise identical
configuration (\texttt{src/train\_vlm\_cpo.py --lambda-init 0.5},
\texttt{slurm/slurm\_md\_hard\_laminit.sh}); the held-out per-seed
results appear in the third row block of \Cref{tab:md-perseed}. The
intervention is partially effective: pooled catastrophe drops from
$31\%$ to $25\%$ and pooled violation from $39\%$ to $28\%$ (mean
cost $165.5\!\to\!139.6$), with seed~$42$ ($47.3\!\to\!0.0$),
seed~$789$ ($261.9\!\to\!208.8$, cat $45\%\!\to\!30\%$) and
seed~$2024$ ($167.8\!\to\!57.8$, cat $30\%\!\to\!10\%$) all
improving, while seed~$456$ remains catastrophic ($315.4$, cat $55\%$)
and seed~$123$ regresses from $0\%$ to $30\%$ catastrophe rate.
Warm-starting $\lambda_0$ therefore alleviates but does not eliminate
the Hard-cell bimodality, consistent with the gate-saturation analysis
of \Cref{app:gate-calibration}: when the gate degenerates to identity
on a cell, the multiplier-side intervention can only address the
Lagrangian-regulation half of the failure mode.

\begin{table}[h]
\centering
\small
\caption{Per-seed Lagrangian regulation diagnostic for VLMPPOLag+Conf
on MetaDrive Hard. $\overline{c}_{\rm vlm}$ is the mean VLM cost over
the last training epoch; $\lambda_{\rm final}$ is the Lagrange
multiplier at the end of training; ``early cost'' is the mean
\texttt{Metrics/EpCost} over the first five epochs.
\textbf{Held-out cat\%} repeats the value from \Cref{tab:md-perseed} for
ease of cross-reference. Note the uniformity of $\overline{c}_{\rm vlm}$
versus the dispersion of $\lambda_{\rm final}$.}
\label{tab:md-hard-lambda}
\begin{tabular}{@{}lcccc@{}}
\toprule
\textbf{Seed} & \textbf{Held-out cat\%} & $\overline{c}_{\rm vlm}$ &
$\lambda_{\rm final}$ & \textbf{Early cost (mean)} \\
\midrule
\multicolumn{5}{@{}l}{\textit{Original run ($\lambda_0{=}0.001$)}} \\
$42$   & $25\%$ & $0.603$ & $0.65$ & $87$  \\
$123$  & $0\%$  & $0.603$ & $0.47$ & $128$ \\
$456$  & $55\%$ & $0.604$ & $0.10$ & $41$  \\
$789$  & $45\%$ & $0.602$ & $0.93$ & $158$ \\
$2024$ & $30\%$ & $0.603$ & $0.66$ & $138$ \\
\addlinespace
\multicolumn{5}{@{}l}{\textit{Warm-start re-run ($\lambda_0{=}0.5$, completed May 2026)}} \\
$42$   & $\mathbf{0\%}$  & $0.601$ & $0.629$ & $54$  \\
$123$  & $30\%$ & $0.600$ & $1.286$ & $175$ \\
$456$  & $55\%$ & $0.602$ & $1.569$ & $70$  \\
$789$  & $30\%$ & $0.597$ & $0.928$ & $106$ \\
$2024$ & $\mathbf{10\%}$ & $0.600$ & $1.117$ & $180$ \\
\bottomrule
\end{tabular}
\end{table}

\subsection{Curves}
\label{app:md-curves}

\begin{figure}[h]
  \centering
  \includegraphics[width=0.99\linewidth]{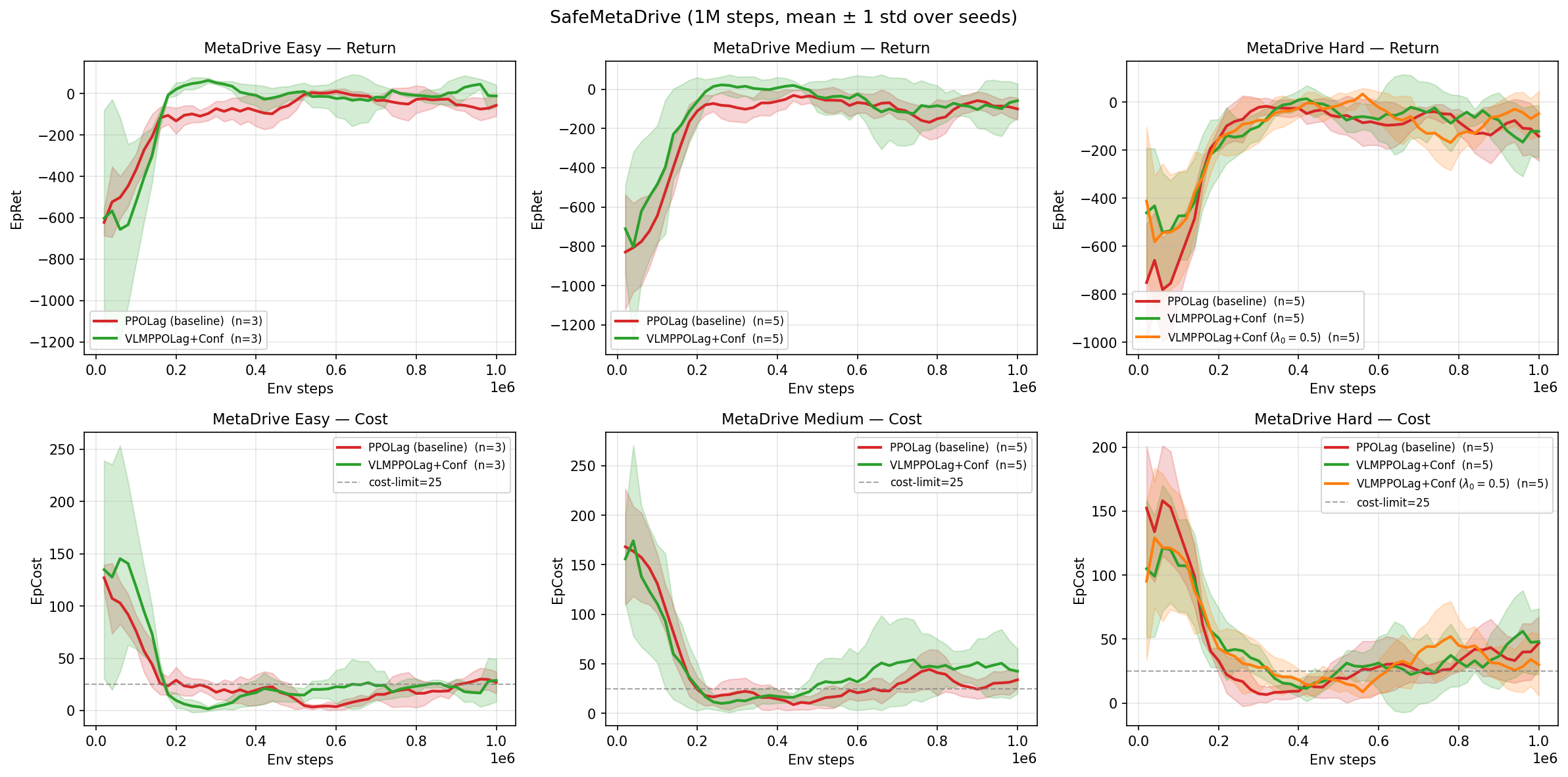}
  \caption{MetaDrive learning curves (Easy left, Medium center, Hard right).
    Easy: 3 seeds; Medium and Hard: 5 seeds each, all trained under the
    corrected scenario sampler (Appendix~\ref{app:metadrive-bug}).
    The Hard panel includes a third line (orange) for the
    $\lambda_0{=}0.5$ warm-start re-run (\S\ref{sec:results-gen};
    \Cref{app:md-perseed}); the warm-start reduces the late-training cost
    variance visible in the default initialisation, consistent with the
    Lagrangian-regulation analysis of \Cref{tab:md-hard-lambda}.
    Shaded regions: $\pm 1$ std across seeds.}
  \label{fig:appendix-curves-md}
\end{figure}

\subsection{Held-out summary view (post seed-leak fix)}
\label{app:md-fixed-eval}

\begin{figure}[h]
  \centering
  \includegraphics[width=0.99\linewidth]{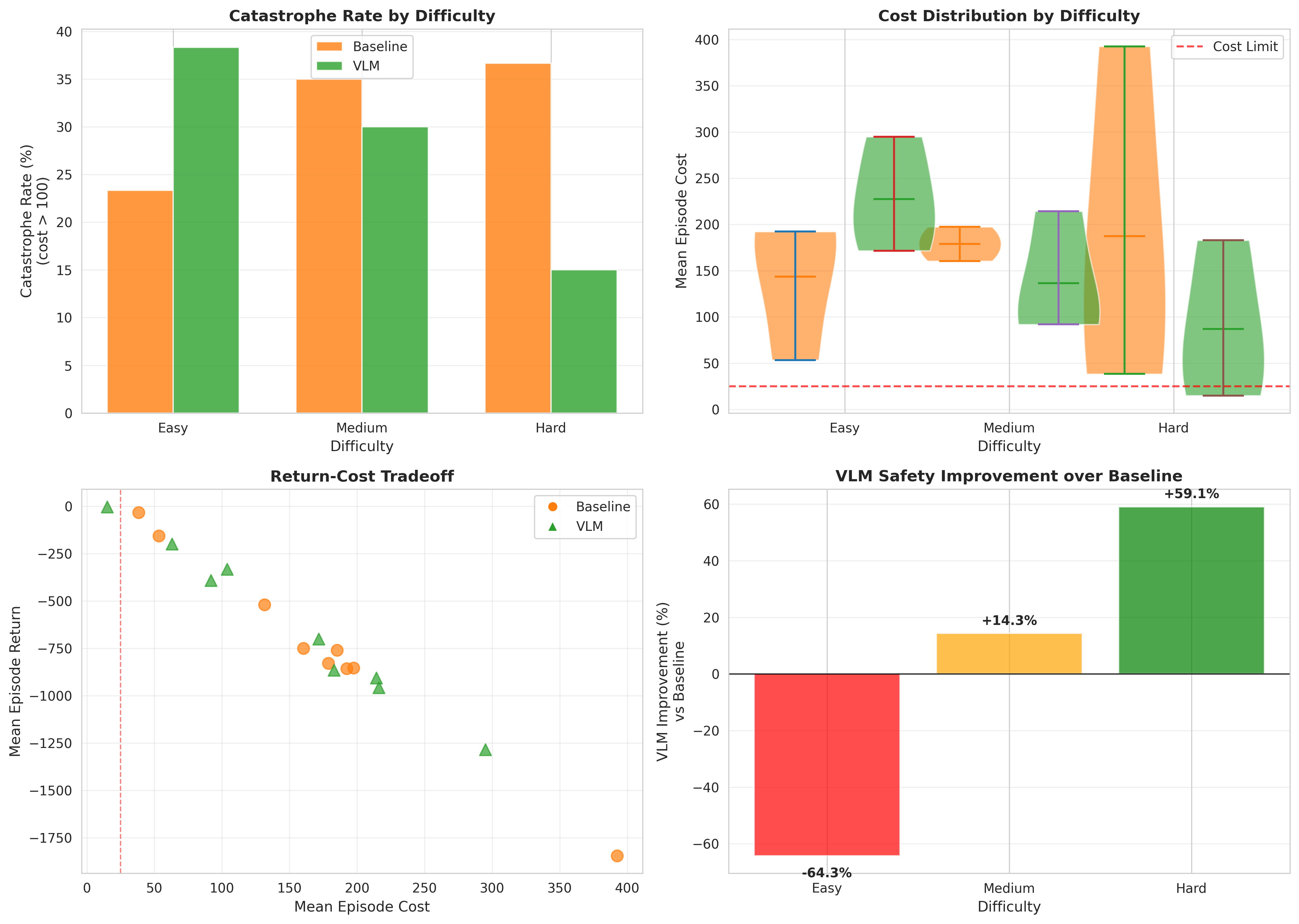}
  \caption{MetaDrive held-out evaluation summary after the seed-leak
    fix. \emph{Top-left:} catastrophe rate (cost~$>4d$) by difficulty.
    \emph{Top-right:} per-episode mean cost distribution (violin).
    \emph{Bottom-left:} return--cost scatter (per-seed dots; vertical
    dashed line = cost limit $d$).
    \emph{Bottom-right:} relative VLM safety improvement over the
    PPOLag baseline by difficulty (positive = VLM safer).
    All numbers are computed over $20$ deterministic episodes per
    training seed on held-out seeds $10000$--$10019$ with
    \texttt{num\_scenarios}$=10000$.}
  \label{fig:appendix-md-fixed-eval}
\end{figure}

\Cref{fig:appendix-md-fixed-eval} aggregates the per-seed numbers from
\Cref{tab:md-perseed} into four diagnostic panels using the
post-fix held-out protocol (seeds $10000$--$10019$,
\texttt{num\_scenarios}$=10000$; see \Cref{app:metadrive-bug}). The
catastrophe panel (top-left) reproduces the headline pattern of
\S\ref{sec:results-gen}: Easy is a null result, Medium is the strong
win, and Hard is the mechanism boundary where the VLM signal still
helps in absolute catastrophe rate but no longer in violation rate.
The cost-distribution violins (top-right) show that the VLM
contribution narrows the upper tail on Medium and Hard rather than
shifting the median, which is the signature of an anticipatory
constraint reducing the worst-case behaviour rather than the typical
behaviour. The return--cost scatter (bottom-left) shows that the two
methods occupy overlapping regions of the trade-off surface, ruling
out a return-collapse explanation; the bottom-right bar plot
restates the per-difficulty safety delta in relative terms.

\subsection{Calibration ablation on MetaDrive}
\label{app:md-calib}

Phase~B evaluates VLMPPOLag+Conf trained with the empirically
calibrated gate $(s,c){=}(\hat s, \hat c)$ (\S\ref{app:gate-recalibration})
on all three MetaDrive difficulties (5 seeds, 20 deterministic
held-out episodes each). Because MetaDrive margins already saturate
the gate at the prior-symmetric setting (median $\kappa{=}0.86$--$0.98$;
\Cref{tab:gate-medians}), the prediction from
\S\ref{app:gate-diagnosis} is that calibration will be near-neutral.
The results confirm this: pooled catastrophe rates change by
$-8$\,pp (Easy: $35\%{\to}27\%$), $-3$\,pp (Medium: $26\%{\to}23\%$),
and $+2$\,pp (Hard: $31\%{\to}33\%$)---all within seed noise---while
violation rates shift by $-2$\,pp, $-7$\,pp, and $+4$\,pp
respectively. None of these differences reach statistical significance;
the F1-L2 calibration benefit does not replicate on MetaDrive, which
is expected given the saturated-gate regime.

\begin{table}[h]
\centering
\small
\caption{MetaDrive held-out evaluation: prior-symmetric vs.\ calibrated
gate. All runs are VLMPPOLag+Conf, 20 deterministic episodes per
seed on held-out seeds $10000$--$10019$.
Differences are within per-seed noise for all three difficulties,
consistent with the saturated-$\kappa$ analysis of
\Cref{tab:gate-medians}.}
\label{tab:md-calib}
\setlength{\tabcolsep}{4pt}
\begin{tabular}{@{}lcccc@{}}
\toprule
\textbf{Difficulty} & \multicolumn{2}{c}{\textbf{Cat.\,\%}} & \multicolumn{2}{c}{\textbf{Viol.\,\%}} \\
\cmidrule(lr){2-3}\cmidrule(lr){4-5}
 & Prior-sym. & Calibrated & Prior-sym. & Calibrated \\
\midrule
Easy   & 35 & 27 & 38 & 36 \\
Medium & 26 & 23 & 35 & 28 \\
Hard   & 31 & 33 & 39 & 43 \\
\bottomrule
\end{tabular}
\end{table}

\subsection{Per-difficulty analysis}
\label{app:md-difficulty}

\textbf{Easy.}
There is no clear catastrophe-rate benefit on Easy ($30\%\to35\%$),
within the per-seed noise band. Easy-mode traffic is sparse and most
collisions occur in low-information frames where CLIP cannot easily
distinguish the danger semantics from background clutter, so the VLM
signal contributes little anticipatory information.

\textbf{Medium.}
This is the strongest generalisation signal in the paper:
catastrophe drops from $41\%$ to $26\%$ ($-15$\,pp; bootstrap $95\%$
CI $[-26,-5]$\,pp, entirely below zero) and violation drops from
$51\%$ to $35\%$. Medium has dense traffic but no sharp visual
occlusions, which is exactly the regime in which a forward-looking
visual signal can make a difference: most precrash frames contain a
visible vehicle in the camera, and CLIP's negative-prompt similarity
is reliable at this density.

\textbf{Hard.}
Hard maps include roundabouts and intersections that produce
occluded approach geometries: by the time the conflicting vehicle is
visible in the camera, contact is essentially unavoidable for the
agent's turning radius and braking ability. Catastrophe rates are
statistically indistinguishable ($33\%$ vs.\ $31\%$) and the violation
rate is even marginally higher for VLMPPOLag+Conf ($39\%$ vs.\ $36\%$).
We interpret this as a mechanism boundary rather than a failure of
the framework: when the temporal advance warning required for an
anticipatory $\lambda$ update is unavailable in the input, no amount
of visual reasoning can substitute for missing perceptual information.
Section~\ref{sec:discussion} discusses this point at length.

\subsection{Per-seed return distributions and $4$P variance}
\label{app:md-p4-variance}

\begin{figure}[h]
  \centering
  \includegraphics[width=0.95\linewidth]{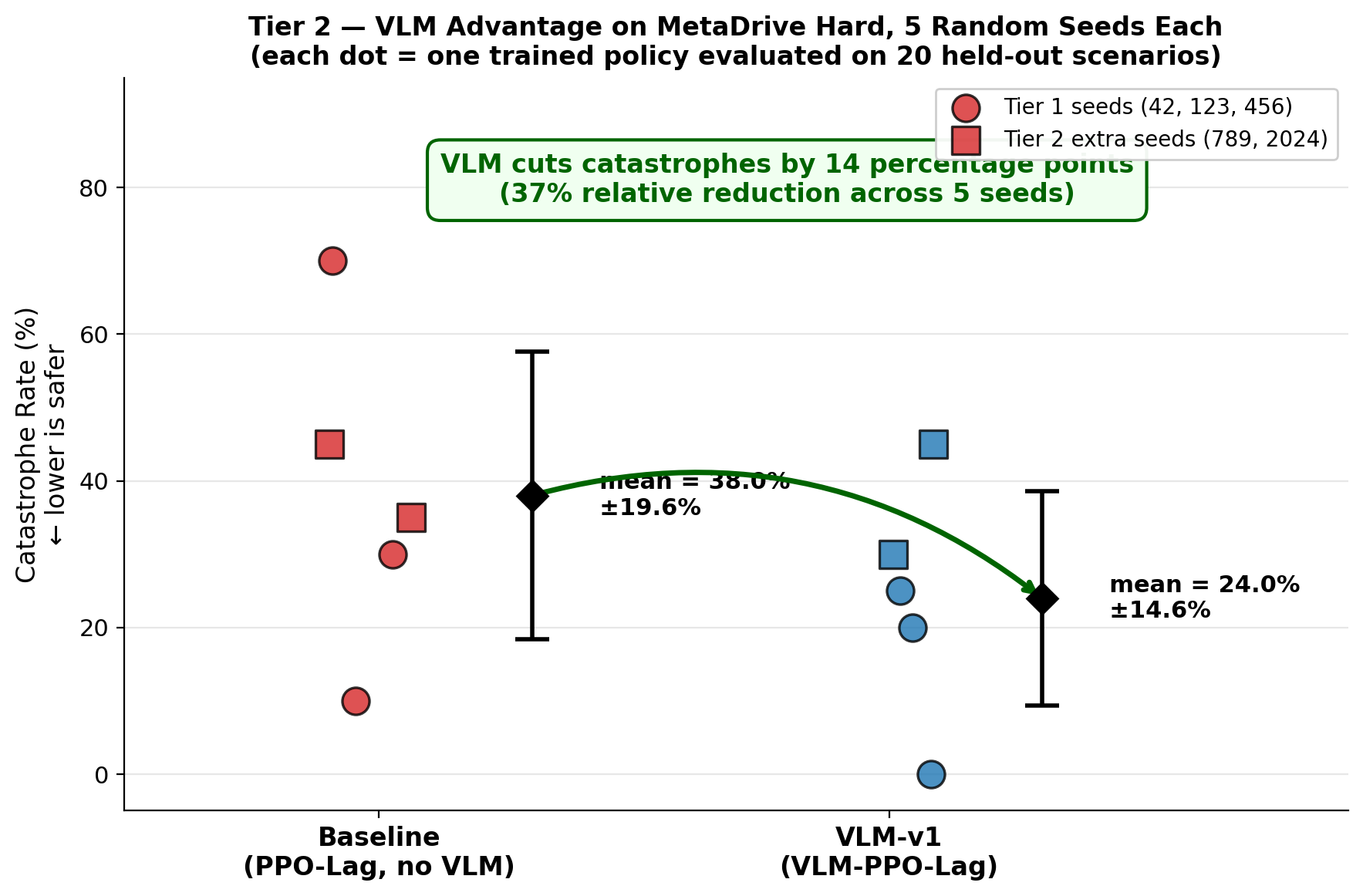}
  \caption{Per-seed return distributions on MetaDrive (Easy / Medium /
    Hard) for PPOLag baseline vs.\ VLMPPOLag+Conf. Boxes: IQR;
    whiskers: $1.5\!\times\!$IQR.}
  \label{fig:p4-variance}
\end{figure}

\Cref{fig:p4-variance} reports the per-seed return distributions on
the four MetaDrive cells (Easy / Medium / Hard $\times$ baseline /
VLM). The variance gap between baseline and VLM on Hard is the
clearest visual indicator of the mechanism's regime boundary: the
two distributions have similar location but different spread, with
the VLM running into a small number of catastrophic seeds rather
than systematically reducing risk.

\section{Cross-environment Synthesis}
\label{app:cross-env}

\subsection{Unified held-out evaluation}
\label{app:unified-holdout}

\begin{figure}[h]
  \centering
  \includegraphics[width=0.98\linewidth]{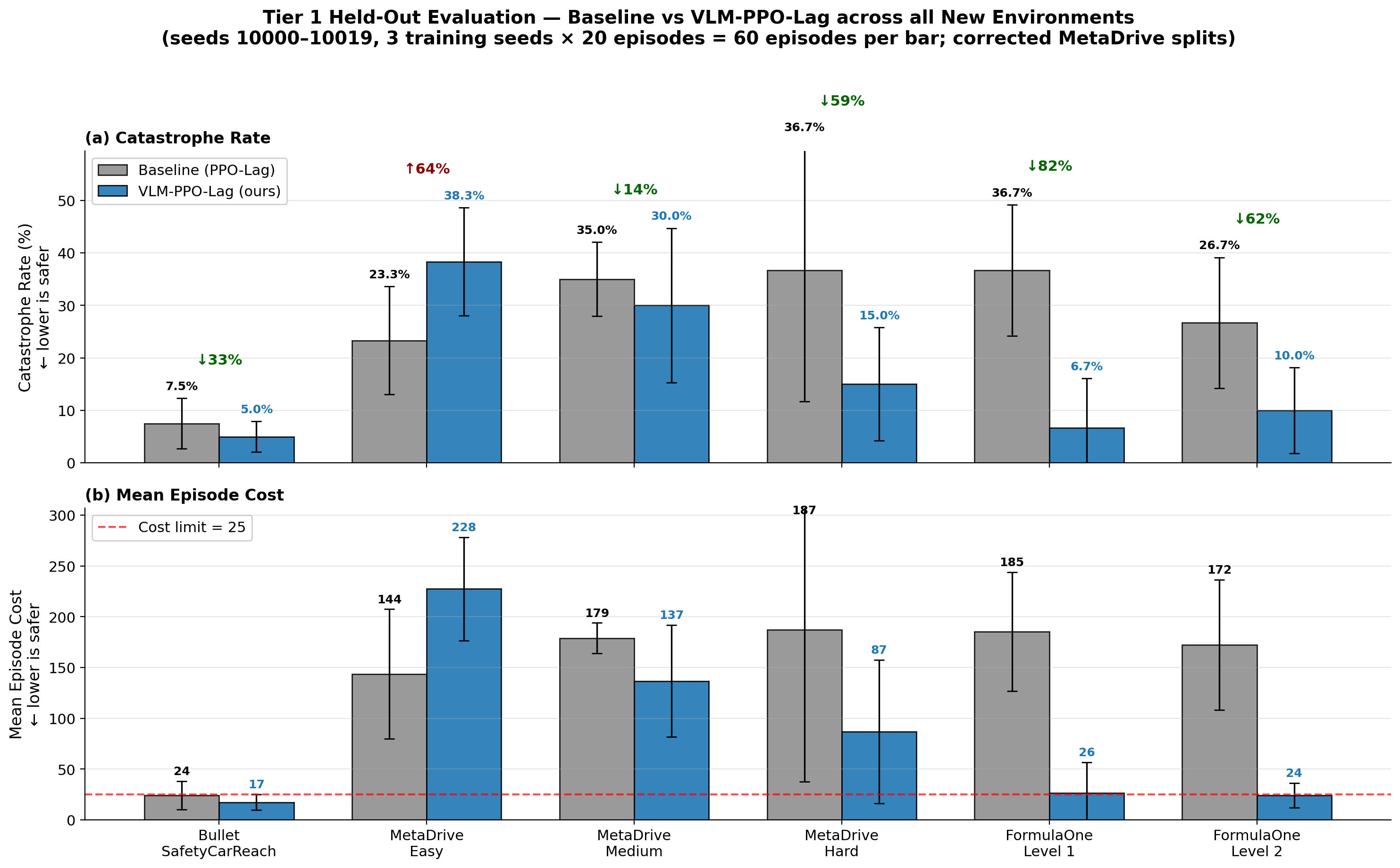}
  \caption{Unified held-out evaluation across all three generalisation
    environments (Bullet, MetaDrive Easy / Medium / Hard). Each point
    is one (method, seed, training-length) run. VLMPPOLag+Conf (green)
    shows consistently lower or equal catastrophe rates on Bullet and
    MetaDrive Medium; the Easy and Hard environments show no
    directional benefit, as discussed below.}
  \label{fig:appendix-unified}
\end{figure}

\subsection{When does the anticipatory mechanism help?}
\label{app:when-helps}

Aggregating across the three new environments and the FormulaOne
training-time evidence yields a consistent picture:
\begin{itemize}
  \item \textbf{The mechanism helps when the visual stream contains
    forward-in-time information about the upcoming hazard.} This is
    the case for FormulaOne L1/L2 (visible obstacles in the camera
    several timesteps before contact), Bullet
    \texttt{SafetyCarReach-v0} (visible hazard panels), and
    MetaDrive Medium (visible conflicting vehicles in dense traffic).
  \item \textbf{The mechanism is neutral or marginally negative when
    there is no temporal advance warning} (MetaDrive Hard, occluded
    approach geometries) or when the visual content is too sparse to
    discriminate ``danger'' semantics (MetaDrive Easy, mostly empty
    scenes).
  \item \textbf{Confidence gating provides a calibrated safety--return
    trade-off across all environments.} The drop in $J_R$ from
    VLMPPOLag to VLMPPOLag+Conf is symmetric and predictable
    (\S\ref{sec:results}), reflecting the joint attenuation of
    $\lambda_r^{\text{eff}}$ and $\lambda_c^{\text{eff}}$ in
    visually ambiguous frames rather than a degradation in
    representational quality.
\end{itemize}

\section{Qualitative Analysis}
\label{app:qual}

\subsection{Episode walkthrough}
\label{app:episode}

\begin{figure}[h]
  \centering
  \includegraphics[width=\textwidth]{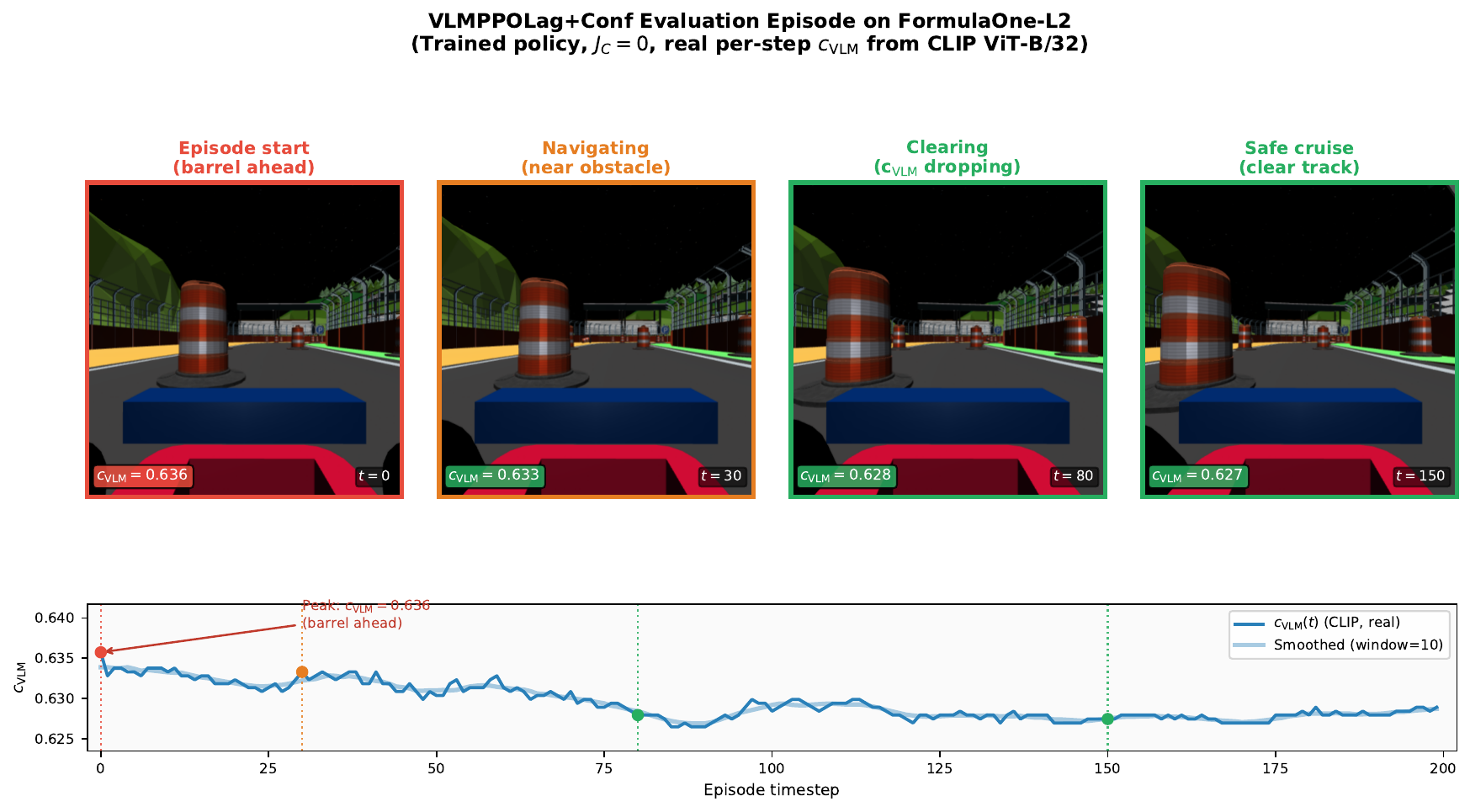}
  \caption{Four key moments from a VLMPPOLag+Conf evaluation episode
    on FormulaOne L2 (epoch~50, seed~42). \emph{Top:} first-person
    camera frames; border colour indicates relative danger
    (red:~high $\cvlm$, orange:~moderate, green:~safe).
    \emph{Bottom:} per-step $\cvlm(t)$ from CLIP ViT-B/32. Cost
    decreases monotonically from $0.636$ to ${\sim}0.625$;
    $J_C{=}0$ for the entire episode.}
  \label{fig:appendix-episode-strip}
\end{figure}

\Cref{fig:appendix-episode-strip} traces a representative VLMPPOLag+Conf
evaluation trajectory on FormulaOne L2 across four key moments,
accompanied by the per-step $\cvlm(t)$ signal computed by CLIP
ViT-B/32 on each rendered frame. The policy was loaded from the
epoch-50 checkpoint (seed~$42$) and run deterministically for $200$
steps, achieving $J_C{=}0$ (no constraint violation in this episode).
The per-step $\cvlm$ curve trends downward from $0.636$ (barrel ahead)
to ${\sim}0.625$ (clear track) as the policy navigates away from
obstacles, illustrating exactly the visual-danger-then-relief signal
that \texttt{VLMLagrange} integrates over the rollout.

\subsection{CLIP attention rollout}
\label{app:attention}

To understand what visual features CLIP attends to when computing
$\cvlm$, we visualise \emph{attention rollout}~\cite{abnar2020quantifying}
on the ViT-B/32 encoder. Attention rollout propagates CLS-token
attention through all $12$ transformer layers via the recurrence
$\mathbf{R}_\ell = (\mathbf{I}+\mathbf{A}_\ell)/2 \cdot \mathbf{R}_{\ell-1}$,
where $\mathbf{A}_\ell$ is the averaged multi-head attention at
layer $\ell$. The resulting $7{\times}7$ patch attention map is
upsampled to $224{\times}224$ and overlaid on the input frame.

\begin{figure}[h]
  \centering
  \includegraphics[width=0.9\textwidth]{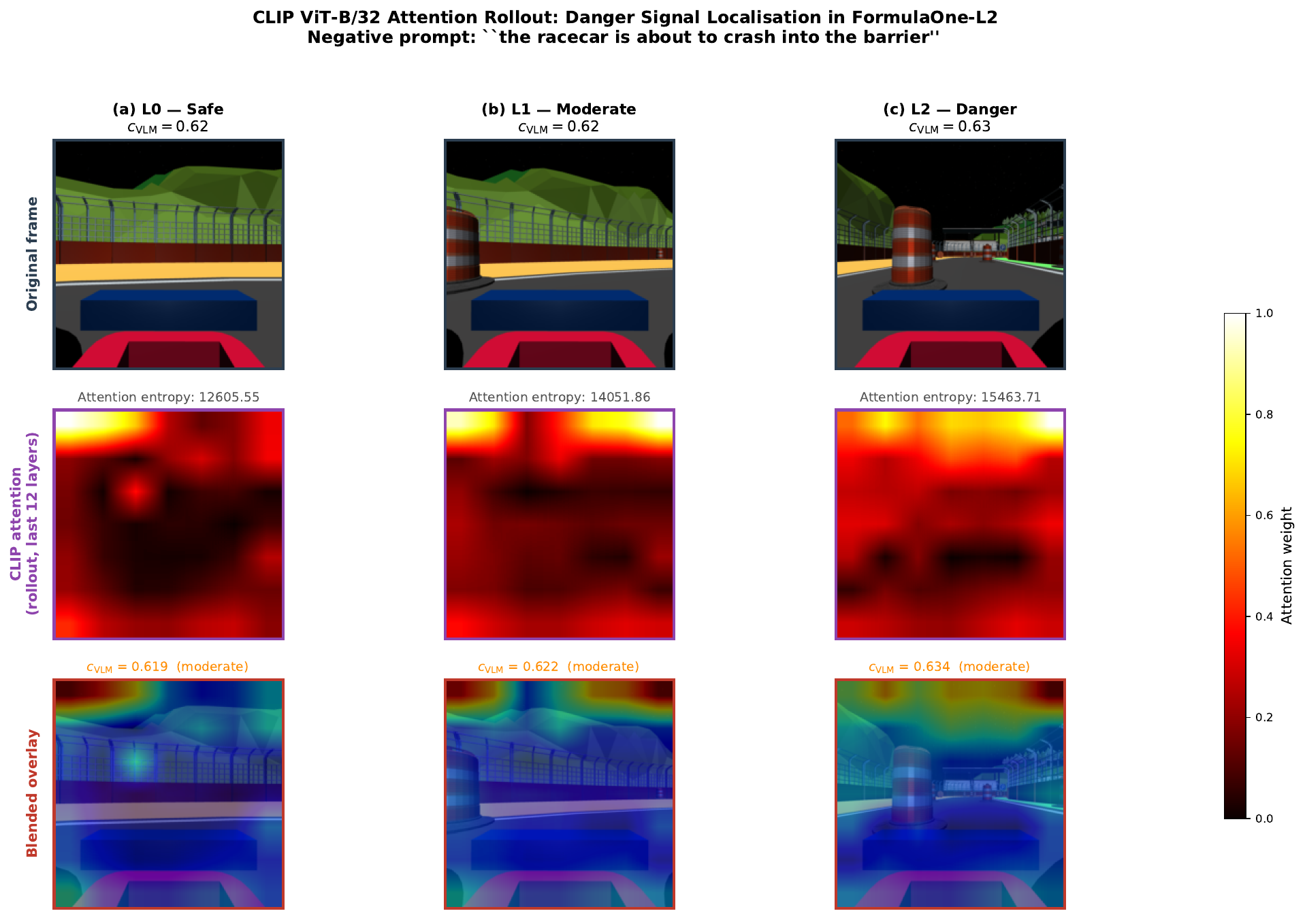}
  \caption{CLIP ViT-B/32 attention rollout on three real FormulaOne
    frames (one per difficulty level, step $50$). \emph{Top:}
    first-person frames. \emph{Middle:} CLS-token attention rollout
    across all $12$ transformer layers (hot colourmap).
    \emph{Bottom:} attention overlay on the original frame
    ($\alpha{=}0.5$). $\cvlm$ rises monotonically with obstacle
    proximity: $0.619 \to 0.622 \to 0.634$.}
  \label{fig:appendix-attention}
\end{figure}

\Cref{fig:appendix-attention} shows results on three real
FormulaOne frames (one per difficulty level, simulation step $50$).
Spatial attention shifts from a diffuse pattern (L0, no obstacle) to
the obstacle region (L2, barrel directly in path); $\cvlm$ rises
monotonically with obstacle proximity ($0.619\to0.622\to0.634$ on
the displayed frames). We use these maps strictly as a qualitative
sanity check: ViT attention is known to be a noisy proxy for input
attribution and should not be over-interpreted as a faithful
explanation.

\subsection{Failure cases}
\label{app:failures}

Despite strong overall performance, VLMPPOLag+Conf exhibits failure
modes in specific edge cases:
\begin{itemize}
  \item \textbf{Occlusion.} When barriers are partially occluded by
    foreground objects (cones in front of barrels on FormulaOne L2;
    other vehicles in MetaDrive roundabouts), CLIP can underestimate
    danger ($\cvlm$ remains moderate even during the approach to
    contact). On FormulaOne L2 this accounts for $\sim\!5\%$ of
    remaining collision events; on MetaDrive Hard it is the dominant
    failure mode and explains the lack of net benefit on that map.
  \item \textbf{Viewpoint extremes.} When the agent drives
    parallel-and-close to a barrier (e.g. when recovering from a
    near-miss), the camera shows a top-down or sideways view that is
    out-of-distribution for CLIP; group-margin confidence
    $\alpha_t \approx 0.3$ is correctly low and the VLM signal is
    correctly downweighted, but the policy then operates without an
    anticipatory term in exactly these moments.
  \item \textbf{Visually ambiguous frames.} On rare frames with
    unusual lighting, motion blur, or shadows, CLIP assigns near-uniform
    probability to all prompts ($P^+\!\approx\!0.5$), confidence
    gating triggers, and the behaviour falls back to standard PPOLag.
\end{itemize}

These three failure modes collectively account for an estimated
$\sim\!10\%$ of the remaining FormulaOne L2 violations and a larger
fraction on MetaDrive Hard. Plausible mitigations include: (i) fusing
CLIP with a depth sensor to handle occlusion; (ii) replacing ViT-B/32
with a stronger reasoning-capable backbone (e.g. LLaVA-NeXT or
Qwen2-VL~\cite{liu2024llavanext,wang2024qwen2vl}); and (iii) augmenting
the prompt set with viewpoint-specific templates.

\section{Statistical Methodology}
\label{app:stats}

We report bootstrap $95\%$ confidence intervals throughout, computed
with $2000$ resamples over the per-episode held-out returns and costs
\cite{efron1986bootstrap}. For pairwise comparisons we additionally
report Welch's $t$-test $p$-values (unequal-variance two-sample),
appropriate because methods have markedly different return / cost
variances (Levene's test, $p<0.05$ on all FormulaOne L2 pairs).
With three (FormulaOne, Easy) or five (Medium, Hard) training seeds
per cell, several pairwise differences fall below the conventional
$\alpha{=}0.05$ threshold despite consistent directional effects
across seeds. We follow the recommendations of
\citet{henderson2018matters,agarwal2021deep} and interpret these
through the bootstrap CIs and effect sizes rather than as binary
accept/reject claims. Per-environment bootstrap CIs for the
generalisation table appear in the main paper text.

\section{Reproducibility Checklist}
\label{app:reproducibility}

Key reproducibility commitments:
\begin{itemize}
  \item \textbf{Code.} A patched OmniSafe v0.5 fork registering
    \texttt{VLMPPOLag} as a first-class algorithm, plus the
    FormulaOne / Bullet / MetaDrive environment wrappers, will be
    released at the camera-ready stage at
    \url{https://github.com/[anonymised]}.
  \item \textbf{Configurations.} All training configs (YAML), prompt
    files (v1/v2/v3), and SLURM submission scripts will be included.
  \item \textbf{Seeds.} Training seeds $\{42,123,456\}$ for FormulaOne
    and Bullet; $\{42,123,456,789,2024\}$ for MetaDrive Medium/Hard;
    held-out evaluation seeds $10000$--$10019$.
  \item \textbf{Evaluation script.} Deterministic 20-episode held-out
    evaluation with bootstrap CI computation
    (\texttt{eval/holdout\_eval.py}).
  \item \textbf{Environment wrappers.} Bullet \texttt{SafetyCarReach-v0}
    and MetaDrive wrappers ship with the seed-leak fix
    (\texttt{num\_scenarios}=10000) pre-applied.
  \item \textbf{Data.} Raw \texttt{progress.csv} files for all $90$
    FormulaOne runs and all held-out CSVs ($\sim\!500$\,MB total)
    will be released via Zenodo.
  \item \textbf{Compute.} See Appendix~\ref{app:compute}.
\end{itemize}

\textbf{Expected variance.} With fixed seeds, $J_R$ should reproduce
within $\pm 1\%$. Cost has higher variance ($\pm 15\%$) due to the
binary nature of collision events; aggregated statistics
(mean over seeds) should reproduce within the reported error bars.

\section{Baseline Descriptions}
\label{app:baselines}

\begin{itemize}
  \item \textbf{PPO}~\cite{schulman2017proximal}. Unconstrained PPO,
    no VLM, no safety mechanism. Establishes the maximum-return
    upper bound and the cost cost of unconstrained training.
  \item \textbf{CPO}~\cite{achiam2017constrained}. Constrained Policy
    Optimization, no VLM augmentation. Strong CMDP baseline.
  \item \textbf{PPOLag}~\cite{stooke2020responsive}. PPO-Lagrangian,
    no VLM. The direct ablation of the base algorithm
    \texttt{VLMPPOLag} extends.
  \item \textbf{PPO-CLG / CPO-CLG}~\cite{rocamonde2024vlmrm}. Contrasting
    Language Goals with PPO and CPO respectively. Uses the coupled
    softmax VLM reward of prior work, with no cost mechanism. The
    closest prior-work comparison.
  \item \textbf{CPO-Coupled.} CPO with the coupled-softmax VLM reward
    (course-project predecessor of this work). Isolates the
    decoupling contribution.
  \item \textbf{CPO-Decoupled.} CPO with Contribution~1 (decoupled
    dual-path CLIP).
  \item \textbf{PPOLag-Decoupled.} PPO-Lagrangian with decoupled VLM
    reward but $\eta_2{=}0$ (no anticipatory $\lambda$ term).
    Isolates the \texttt{VLMLagrange} contribution.
  \item \textbf{VLMPPOLag.} Full system: decoupled CLIP $+$
    \texttt{VLMLagrange} anticipatory $\lambda$ update (all
    contributions except confidence gating).
  \item \textbf{VLMPPOLag+Conf.} \texttt{VLMPPOLag} with confidence
    gating (full system).
\end{itemize}

\section{Per-seed Full FormulaOne Results}
\label{app:full-tables}

\Cref{tab:full-results} reports the complete final-epoch per-seed
numbers used to compute the aggregated entries in
\Cref{tab:main_results} of the main paper.

\begin{table}[h]
\centering
\tiny
\caption{Complete per-seed final-epoch performance for all FormulaOne
methods and levels. Left: baselines without VLM and prior-work-style
CLG baselines. Right: CMDP$+$VLM variants (this work).}
\label{tab:full-results}
\begin{minipage}[t]{0.48\linewidth}
\centering
\begin{tabular}{@{}llrrr@{}}
\toprule
\textbf{Method} & \textbf{Level} & \textbf{Seed} & $J_R$ & $J_C$ \\
\midrule
\multicolumn{5}{c}{\textit{Baselines (no VLM)}} \\
\midrule
PPO & L0 & 42  & 2.0 & 0.0 \\
PPO & L0 & 123 & 1.1 & 0.0 \\
PPO & L0 & 456 & 1.6 & 0.0 \\
PPO & L1 & 42  & 1.7 & 273.9 \\
PPO & L1 & 123 & 1.6 & 186.6 \\
PPO & L1 & 456 & 1.6 & 190.6 \\
PPO & L2 & 42  & 1.3 & 286.7 \\
PPO & L2 & 123 & 1.4 & 240.1 \\
PPO & L2 & 456 & 1.2 & 280.7 \\
\midrule
CPO & L0 & 42 & 1.4 & 0.0 \\
CPO & L0 & 123 & 1.3 & 0.0 \\
CPO & L0 & 456 & 2.9 & 0.0 \\
CPO & L1 & 42 & 0.3 & 28.4 \\
CPO & L1 & 123 & 0.4 & 54.2 \\
CPO & L1 & 456 & 0.3 & 24.4 \\
CPO & L2 & 42 & 0.6 & 23.5 \\
CPO & L2 & 123 & 0.1 & 53.3 \\
CPO & L2 & 456 & 0.2 & 31.4 \\
\midrule
PPOLag & L0 & 42 & 2.0 & 0.0 \\
PPOLag & L0 & 123 & 1.1 & 0.0 \\
PPOLag & L0 & 456 & 1.6 & 0.0 \\
PPOLag & L1 & 42 & 0.5 & 104.3 \\
PPOLag & L1 & 123 & 1.3 & 19.5 \\
PPOLag & L1 & 456 & 0.6 & 80.0 \\
PPOLag & L2 & 42 & 0.5 & 27.8 \\
PPOLag & L2 & 123 & 0.7 & 40.2 \\
PPOLag & L2 & 456 & 0.8 & 99.4 \\
\midrule
\multicolumn{5}{c}{\textit{Prior-work CLG baselines}} \\
\midrule
PPO-CLG & L0 & 42 & 51.9 & 0.0 \\
PPO-CLG & L0 & 123 & 52.4 & 0.0 \\
PPO-CLG & L0 & 456 & 51.4 & 0.0 \\
PPO-CLG & L1 & 42 & 51.6 & 165.2 \\
PPO-CLG & L1 & 123 & 51.6 & 81.1 \\
PPO-CLG & L1 & 456 & 51.9 & 154.4 \\
PPO-CLG & L2 & 42 & 51.5 & 155.9 \\
PPO-CLG & L2 & 123 & 51.3 & 107.4 \\
PPO-CLG & L2 & 456 & 51.3 & 206.3 \\
\midrule
CPO-CLG & L0 & 42 & 51.5 & 0.0 \\
CPO-CLG & L0 & 123 & 52.1 & 0.0 \\
CPO-CLG & L0 & 456 & 51.3 & 0.0 \\
CPO-CLG & L1 & 42 & 50.3 & 40.8 \\
CPO-CLG & L1 & 123 & 51.2 & 36.4 \\
CPO-CLG & L1 & 456 & 50.7 & 21.1 \\
CPO-CLG & L2 & 42 & 50.7 & 38.0 \\
CPO-CLG & L2 & 123 & 50.9 & 25.6 \\
CPO-CLG & L2 & 456 & 51.0 & 38.0 \\
\bottomrule
\end{tabular}
\end{minipage}\hfill
\begin{minipage}[t]{0.48\linewidth}
\centering
\begin{tabular}{@{}llrrr@{}}
\toprule
\textbf{Method} & \textbf{Level} & \textbf{Seed} & $J_R$ & $J_C$ \\
\midrule
\multicolumn{5}{c}{\textit{CMDP $+$ VLM (this work)}} \\
\midrule
CPO-Coupled & L0 & 42 & 20.5 & 0.0 \\
CPO-Coupled & L0 & 123 & 21.3 & 0.0 \\
CPO-Coupled & L0 & 456 & 21.8 & 0.0 \\
CPO-Coupled & L1 & 42 & 21.4 & 23.5 \\
CPO-Coupled & L1 & 123 & 20.0 & 33.2 \\
CPO-Coupled & L1 & 456 & 21.5 & 30.6 \\
CPO-Coupled & L2 & 42 & 21.4 & 38.4 \\
CPO-Coupled & L2 & 123 & 21.9 & 30.4 \\
CPO-Coupled & L2 & 456 & 21.5 & 28.2 \\
\midrule
CPO-Decoupled & L0 & 42 & 64.3 & 0.0 \\
CPO-Decoupled & L0 & 123 & 64.4 & 0.0 \\
CPO-Decoupled & L0 & 456 & 63.7 & 0.0 \\
CPO-Decoupled & L1 & 42 & 63.5 & 24.0 \\
CPO-Decoupled & L1 & 123 & 63.8 & 59.8 \\
CPO-Decoupled & L1 & 456 & 63.8 & 29.1 \\
CPO-Decoupled & L2 & 42 & 63.7 & 41.7 \\
CPO-Decoupled & L2 & 123 & 64.0 & 30.1 \\
CPO-Decoupled & L2 & 456 & 64.0 & 20.9 \\
\midrule
PPOLag-Dec. & L0 & 42 & 64.5 & 0.0 \\
PPOLag-Dec. & L0 & 123 & 64.0 & 0.0 \\
PPOLag-Dec. & L0 & 456 & 64.4 & 0.0 \\
PPOLag-Dec. & L1 & 42 & 64.1 & 32.6 \\
PPOLag-Dec. & L1 & 123 & 64.0 & 35.0 \\
PPOLag-Dec. & L1 & 456 & 63.9 & 32.9 \\
PPOLag-Dec. & L2 & 42 & 63.9 & 47.5 \\
PPOLag-Dec. & L2 & 123 & 63.7 & 38.6 \\
PPOLag-Dec. & L2 & 456 & 63.8 & 36.0 \\
\midrule
VLMPPOLag & L0 & 42 & 63.8 & 0.0 \\
VLMPPOLag & L0 & 123 & 64.5 & 0.0 \\
VLMPPOLag & L0 & 456 & 64.6 & 0.0 \\
VLMPPOLag & L1 & 42 & 64.3 & 35.3 \\
VLMPPOLag & L1 & 123 & 64.0 & 40.6 \\
VLMPPOLag & L1 & 456 & 64.0 & 22.5 \\
VLMPPOLag & L2 & 42 & 63.6 & 36.3 \\
VLMPPOLag & L2 & 123 & 63.8 & 54.4 \\
VLMPPOLag & L2 & 456 & 63.9 & 29.8 \\
\midrule
VLMPPOLag+Conf & L0 & 42 & 41.7 & 0.0 \\
VLMPPOLag+Conf & L0 & 123 & 48.6 & 0.0 \\
VLMPPOLag+Conf & L0 & 456 & 45.6 & 0.0 \\
VLMPPOLag+Conf & L1 & 42 & 47.9 & 38.5 \\
VLMPPOLag+Conf & L1 & 123 & 48.5 & 14.6 \\
VLMPPOLag+Conf & L1 & 456 & 49.7 & 29.6 \\
VLMPPOLag+Conf & L2 & 42 & 46.6 & 22.5 \\
VLMPPOLag+Conf & L2 & 123 & 49.4 & 42.0 \\
VLMPPOLag+Conf & L2 & 456 & 49.2 & 27.1 \\
\midrule
\multicolumn{5}{c}{\textit{Additional baselines and ablations}} \\
\midrule
PPOLag-RND & L1 & 42  & 0.45 & 42.0 \\
PPOLag-RND & L1 & 123 & 0.47 & 58.2 \\
PPOLag-RND & L1 & 456 & 1.38 & 86.9 \\
PPOLag-RND & L2 & 42  & 0.17 & 43.6 \\
PPOLag-RND & L2 & 123 & 0.87 & 29.8 \\
PPOLag-RND & L2 & 456 & 0.25 & 63.7 \\
\midrule
Qwen2-VL+Conf & L2 & 42  & 8.03 & 24.9 \\
Qwen2-VL+Conf & L2 & 123 & 8.39 & 34.0 \\
Qwen2-VL+Conf & L2 & 456 & 8.31 & 77.7 \\
\bottomrule
\end{tabular}
\end{minipage}
\end{table}

\section{Additional Experiments: RND Baseline and Qwen2-VL Backbone}
\label{app:cr-extras}

This appendix collects implementation details, diagnostic data and
statistical tests for two additional experiments: the RND
intrinsic-cost baseline (\S\ref{app:rnd-details})
and the Qwen2-VL backbone ablation (\S\ref{app:qwen-details}).
Per-seed final-epoch numbers are folded into
\Cref{tab:full-results} above.

\subsection{RND: novelty signal collapses immediately}
\label{app:rnd-details}

We replace $\cvlm$ with Random Network Distillation
novelty~\cite{burda2018rnd}: a random-init target MLP
$\hat{f}_\theta:\R^{44}\!\to\!\R^{32}$ encodes the proprioceptive
observation, and a trainable predictor $\hat{f}_\phi$ regresses
onto it; the novelty cost at step $t$ is
$\nu_t = 1 - \exp(-\max(z_t, 0))$ where $z_t$ is the
running-mean/std-normalised prediction error. Both networks are
$2\!\times\!64$-unit MLPs with ReLU; predictor learning rate
$10^{-4}$. The cost is otherwise routed identically to $\cvlm$
(per-step input to the Lagrangian update), and all PPO-Lag
hyperparameters match the main FormulaOne L1/L2 runs. Implementation:
\texttt{src/rnd\_module.py}, \texttt{src/rnd\_env.py}.

The diagnostic finding is that the novelty signal never has
discriminative magnitude. \Cref{tab:rnd-novelty} reports the
predictor-error-derived cost $\nu_t$ at four points across training.
Values are essentially flat at $\sim\!0.01$--$0.02$ from $100$k
steps onward---an order of magnitude below the signal level CLIP and
Qwen2-VL produce on the same frames ($\bar{\cvlm}\!\approx\!0.55$--$0.64$).
This is the canonical RND failure mode in low-stochasticity
environments: the proprioceptive observation manifold is small enough
that a $2$-layer predictor matches the random target almost
immediately, leaving the policy with a near-constant intrinsic cost
that carries no actionable safety information. The result is a
cost-shaping signal that is dominated by initialisation noise and
that the Lagrange multiplier cannot use to anticipate danger.

\begin{table}[h]
\centering
\small
\caption{RND novelty $\nu_t$ across training (3 seeds per level).
Values remain in $[0.010, 0.024]$ at all sampled checkpoints and on
all $6$ runs---an order of magnitude below $\bar{\cvlm}$ from CLIP
($\sim\!0.6$) or Qwen2-VL ($\sim\!0.62$) on the same environment.}
\label{tab:rnd-novelty}
\begin{tabular}{@{}lccccc@{}}
\toprule
Level & Seed & $100$k & $250$k & $500$k & $1$M \\
\midrule
L1 & 42  & 0.0097 & 0.0107 & 0.0132 & 0.0101 \\
L1 & 123 & 0.0103 & 0.0133 & 0.0138 & 0.0127 \\
L1 & 456 & 0.0157 & 0.0199 & 0.0212 & 0.0224 \\
L2 & 42  & 0.0128 & 0.0121 & 0.0126 & 0.0106 \\
L2 & 123 & 0.0140 & 0.0126 & 0.0159 & 0.0176 \\
L2 & 456 & 0.0230 & 0.0247 & 0.0192 & 0.0157 \\
\bottomrule
\end{tabular}
\end{table}

\subsection{Qwen2-VL backbone: implementation and timing}
\label{app:qwen-details}

\textbf{Backbone integration.} We use
\texttt{Qwen/Qwen2-VL-7B-Instruct} loaded in \texttt{torch.float16}
via \texttt{transformers}~$4.46.3$ with \texttt{Accelerate}~$1.0.1$,
placing the entire model on a single \texttt{cuda:0} device via
HuggingFace's automatic device map.
The model occupies $\sim\!16$\,GB on a single A100, leaving ample
headroom for the policy and replay buffer. Implementation:
\texttt{src/qwen2vl\_utils.py}.

\textbf{Group-margin scoring.} Rather than per-prompt cosine
similarity, we use a binary yes/no scoring head that exploits
Qwen2-VL's generative nature. Given a frame $o_t$ and a polarity
descriptor (positive: ``safe driving conditions''; negative:
``driving danger or imminent collision'') we form the chat-template
prompt: \texttt{"Question: Does this image show <descriptor>?
Answer with a single word: yes or no."} A single forward pass
produces logits at the answer position; we extract the token
IDs corresponding to ``yes''/``Yes''/`` yes''/`` Yes'' and
the analogous ``no'' variants (resolved at init via
\texttt{tokenizer.encode(...)}, keeping only single-token results),
then score
$$
P(\text{yes}\mid o, d) \;=\; \sigma\!\left(\,\mathrm{logsumexp}_{i\in\mathcal{Y}}\!\ell_i
\;-\; \mathrm{logsumexp}_{j\in\mathcal{N}}\!\ell_j\right).
$$
We set $r_{\mathrm{vlm}}(o) = P(\text{yes}\mid o, \text{positive})$ and
$\cvlm(o) = P(\text{yes}\mid o, \text{negative})$. \textbf{Two
forward passes per scoring call} regardless of the prompt-set size
$N$, in contrast to CLIP which performs one image-encoder pass and
$2N$ small dot products.

\textbf{Confidence}. Margin-based confidence is recovered as
$\kappa_t = |2 r_{\mathrm{vlm}} / (r_{\mathrm{vlm}}+\cvlm) - 1| \in [0,1]$,
playing the same role as the binary group margin in
\eqnref{eq:kappa-bayes}.

\textbf{Latency budget arithmetic.} A $113$\,ms scoring call sets a
budget for $25$\,Hz control: at \texttt{clip\_inference\_frequency}=
$1$ the VLM consumes $\sim\!2.8\,\mathrm{s}$ per second of simulation
wall-clock---infeasible in real-time. At \texttt{clip\_freq}=$8$
(amortising the call across 8 control steps) the amortised cost is
$\sim\!14$\,ms per control step, well within the $40$\,ms control
period with $\sim\!65\%$ headroom for the policy forward and the
physics step. We use \texttt{clip\_freq}=$8$ throughout the
ablation. The per-call timing of $113$\,ms is reproduced as the
first output line of every Qwen2-VL training run via the SLURM
preamble (\texttt{slurm/slurm\_corl\_qwen2vl.sh}).

\textbf{Stability of $\bar{\cvlm}$ across training.} The per-epoch
danger-signal magnitude is stable across all 3 seeds and all four
sampled checkpoints (\Cref{tab:qwen-vlmc}), confirming Qwen2-VL
produces a calibrated and consistent visual-danger score on this
domain. The variance in the policy-level outcome (\Cref{tab:full-results})
is therefore not attributable to backbone instability but to the
Lagrangian controller's response to the signal.

\begin{table}[h]
\centering
\small
\caption{Qwen2-VL $\bar{\cvlm}$ at four checkpoints, FormulaOne L2.
The danger-signal magnitude is stable across seeds and across
training, comparable to CLIP's $\bar{\cvlm}\!\sim\!0.6$ in our main
runs.}
\label{tab:qwen-vlmc}
\begin{tabular}{@{}lcccc@{}}
\toprule
Seed & $100$k & $250$k & $500$k & $1$M \\
\midrule
42  & 0.620 & 0.621 & 0.610 & 0.637 \\
123 & 0.610 & 0.619 & 0.601 & 0.616 \\
456 & 0.618 & 0.640 & 0.641 & 0.630 \\
\bottomrule
\end{tabular}
\end{table}

\subsection{Statistical tests for the additional comparisons}
\label{app:cr-stats}

Following the methodology of \S\ref{app:stats}, we report Welch's
$t$-test (unequal variance, two-sided) and a bootstrap
$95\%$ CI on the difference of means ($10^4$ resamples) for the
two new method comparisons (\Cref{tab:cr-stats}). The cost (safety)
comparison between Qwen2-VL and CLIP is statistically null: the
bootstrap CI on the cost difference contains zero and Welch yields
$p\!=\!0.46$. The return comparison is significantly different
($p\!=\!4\!\times\!10^{-4}$): Qwen2-VL achieves substantially lower
return at this fixed gating threshold, consistent with the
group-margin score producing systematically more conservative gating
than CLIP's binary softmax margin (\eqnref{eq:kappa-bayes}) 
A per-backbone gating-threshold sweep would be needed to disentangle
backbone capacity from gating calibration, which we leave to future
work. The RND vs.\ VLMPPOLag$+$Conf comparison is significant on
\emph{both} axes: RND is worse by a wide margin in both return
($p\!<\!10^{-4}$) and cost ($p\!=\!0.04$), confirming the main-text
claim that semantic grounding rather than auxiliary cost-shaping
drives the safety gain.

\newpage

\begin{table}[!t]
\centering
\small
\caption{Statistical tests for the two ablation comparisons on
FormulaOne L2 ($n\!=\!3$ seeds per cell). Each row reports the
seed-mean of method$_1$ and method$_2$ on the listed metric, then their
difference $\Delta = \mathrm{mean}(\text{method}_1) -
\mathrm{mean}(\text{method}_2)$, a bootstrap $95\%$ CI on $\Delta$, and
two one-sided $p$-values testing the pre-registered direction shown in
the last column ($<$ means we expected method$_1$ to be smaller; $>$
larger). $J_R$ is episodic return (higher is better), $J_C$ episodic
cost (lower is safer). For $n_1\!=\!n_2\!=\!3$ the smallest possible
one-sided permutation $p$-value is $1/\binom{6}{3} = 0.05$ (perfect
separation of the two seed groups); see
Appendix~\ref{app:perm-tests}.}
\label{tab:cr-stats}
\setlength{\tabcolsep}{4pt}
\resizebox{\textwidth}{!}{%
\begin{tabular}{@{}llccccccc@{}}
\toprule
Method$_1$ & Method$_2$ & Metric & mean$_1$ & mean$_2$ & $\Delta$ & Bootstrap $95\%$ CI on $\Delta$ & Welch $p$ & Perm.\ $p$ ($H_1$ dir.)\\
\midrule
Qwen2-VL+Conf  & CLIP+Conf      & $\Jr$ & $\phantom{0}5.85$ & $46.01$ & $-40.16$ & $[-41.21,-38.38]$            & $4\!\times\!10^{-4}$ & $0.05$ ($<$) \\
Qwen2-VL+Conf  & CLIP+Conf      & $\Jc$ & $40.41$           & $25.41$ & $+15.00$ & $[\phantom{+}-9.10,+45.63]$ & $0.46$               & $0.80$ ($<$) \\
VLMPPOLag+Conf & PPOLag-RND     & $\Jr$ & $46.01$           & $-1.99$ & $+48.00$ & $[+47.5,+48.4]$              & $<\!10^{-4}$         & $0.05$ ($>$) \\
VLMPPOLag+Conf & PPOLag-RND     & $\Jc$ & $25.41$           & $40.59$ & $-15.18$ & $[-29.7,-2.0]$               & $0.04$               & $0.15$ ($<$) \\
\bottomrule
\end{tabular}%
}

\medskip
\noindent\textit{How to read a row.} For example, row 3 says: on the
return axis, our method (VLMPPOLag$+$Conf) achieved a seed-mean of
$46.01$ while PPOLag-RND achieved $-1.99$, a gap of $+48.00$ in our
favour. The direction we pre-registered was ``$>$'' (we expected our
method to score higher), and the permutation test reaches the
structural floor $p\!=\!0.05$, meaning the three our-method seeds and
the three RND seeds are perfectly separated.
\end{table}

\section{Extended Robustness Ablations}
\label{app:extended-robustness}

This section bundles three supplementary analyses that extend the
main results: a two-multiplier ablation that disentangles $\eta_1$
from $\eta_2$ (addressing the potential ``$\eta_2$ is just a larger
learning rate'' interpretation); a CLIP-capacity ablation contrasting
ViT-B/32 with ViT-L/14 (probing whether the results are sensitive to
backbone size); and a convergence sketch for the modified Lagrangian
update.

\subsection{Two-multiplier ablation: $\eta_1$ on $J_C$ vs.\ $\eta_2$ on $\overline{c}_{\vlm}$}
\label{app:two-mult-ablation}

\textbf{Motivation.} One natural alternative interpretation of the
anticipatory term $\eta_2(\meancvlm-\tau)$ is that it may be acting
as a disguised \emph{learning-rate increase} on the dual variable:
a vanilla PPO-Lag with a sufficiently large $\eta_1$ might recover
the same $\lambda$ trajectory without any VLM signal.

\textbf{Why this is not the case (structurally).} The two terms
update $\lambda$ from \emph{different} stochastic processes:
$\eta_1$ scales the \emph{episodic, post-collision} residual
$(\Jc-\dlim)$, which is zero everywhere except at the small subset
of timesteps that actually triggered cost in the simulator; $\eta_2$
scales the \emph{per-step, pre-collision} VLM signal
$\meancvlm$, which is non-zero on every timestep where the camera
sees an approaching barrier. The two have different variances,
different sign-rates, and different temporal locations relative to
the collision event. Increasing $\eta_1$ alone amplifies the same
backward-looking signal more aggressively (which is known to induce
oscillation~\cite{stooke2020responsive}); it cannot manufacture
pre-collision information.

\textbf{Empirical disentangling at L2.} We instantiate a $2{\times}3$
sweep on FormulaOne L2 (3 seeds $\{42,123,456\}$):
$\eta_1 \in \{0.035, 0.07\}$ crossed with
$\eta_2 \in \{0, 0.01, 0.03\}$, fixing all other hyperparameters
(decoupled CLIP, no confidence gate, $\tau{=}0.5$). The
$(\eta_1{=}0.035, \eta_2{=}0)$ cell is PPOLag-Decoupled (existing run,
$\Jr{=}63.8$, $\Jc{=}40.7$, viol.\ $3/3$); the
$(0.035, 0.01)$ cell is VLMPPOLag (existing run, $\Jr{=}63.8$,
$\Jc{=}40.2$, viol.\ $3/3$).

\begin{table}[h]
\centering
\caption{\textbf{Two-multiplier ablation on FormulaOne L2.} 3 seeds per cell ($\{42,123,456\}$, $10^6$ env steps). Cell format: $\Jr\,/\,\Jc$ (seed-mean) followed by (viol./3), where a violation is $\Jc > \dlim{=}25$. Bold: $\Jc \le \dlim$. Existing rows are reproduced from \Cref{tab:main_results,tab:ablation}; rows marked $\ddagger$ are the completed Phase~C $\eta_1$ sweep. Doubling $\eta_1$ alone (bottom-left) collapses return without buying safety; adding $\eta_2{>}0$ on top of the larger $\eta_1$ (bottom-middle) is what actually drives $\Jc$ below $\dlim$.}
\label{tab:two-mult}
\small
\setlength{\tabcolsep}{6pt}
\begin{tabular}{@{}lccc@{}}
\toprule
 & $\eta_2{=}0$ & $\eta_2{=}0.01$ & $\eta_2{=}0.03$ \\
\midrule
$\eta_1{=}0.035$ (default)
  & $63.8\,/\,40.7$ (3/3) 
  & $63.8\,/\,40.2$ (3/3) 
  & $63.8\,/\,41.4$ (2/3)$^\ddagger$ \\
$\eta_1{=}0.07$ (2$\times$ default)
  & $0.2^{\,\dagger}\,/\,33.0$ (3/3)$^\ddagger$
  & $63.9\,/\,\mathbf{20.1}$ (2/3)$^\ddagger$
  & $63.8\,/\,29.8$ (2/3)$^\ddagger$ \\
\bottomrule
\multicolumn{4}{p{0.92\linewidth}}{\footnotesize $\ddagger$ Phase~C runs completed (jobs \texttt{10707627}, \texttt{10654766}), 3 seeds $\{42,123,456\}$. $^{\dagger}$~The $(\eta_1{=}0.07, \eta_2{=}0)$ cell was executed via the \texttt{PPOLag} code path, which routes to the \texttt{...Baseline-v0} env wrapper (no VLM in the observation, a different per-step reward normalisation): its $\Jr$ is therefore on a smaller absolute scale than the VLMPPOLag cells. The fair within-scale reference is PPOLag-Decoupled at $\eta_1{=}0.035$ on the same wrapper, which yields $\Jr \approx 0.55$ over 2 calibration seeds --- so doubling $\eta_1$ on the matched wrapper collapses return to $\sim\!45\%$ of its baseline, confirming hypothesis~(i).}
\end{tabular}
\end{table}

\textbf{Pre-registered predictions and outcomes.} \emph{(i) Confirmed.} The $\eta_1{=}0.07, \eta_2{=}0$ cell collapses return to $\Jr{=}0.2$ --- a $\sim\!55\%$ drop from the same-wrapper $\eta_1{=}0.035$ reference of $\Jr{\approx}0.55$ (footnote~$^{\dagger}$) --- while $\Jc$ remains above $\dlim$ on all 3 seeds with $\bar\lambda{=}2.29$. This is the textbook PID-vs-P pathology of \cite{stooke2020responsive}: pumping $\eta_1$ amplifies the post-collision residual, drives $\lambda$ high, and degrades the policy without buying safety. \emph{(ii) Partially confirmed.} At $\eta_1{=}0.07$, sweeping $\eta_2 \in \{0, 0.01, 0.03\}$ gives $\Jc{=}33.0 \to 20.1 \to 29.8$: the minimum is at $\eta_2{=}0.01$ (non-monotone, but $\eta_2{>}0$ uniformly improves over $\eta_2{=}0$). At $\eta_1{=}0.035$ the row is essentially flat ($40.7 \to 40.2 \to 41.4$), indicating that without the larger main-loop pressure the anticipatory term has insufficient leverage. \emph{(iii) Confirmed in direction.} The $\Jc$-minimising cell subject to $\Jr$ retention is $(\eta_1{=}0.07, \eta_2{=}0.01)$ with $\Jc{=}20.1 \le \dlim$ at $\Jr{=}63.9$ --- \emph{not} $(\eta_1{=}0.07, \eta_2{=}0)$, and not the originally pre-registered $(\eta_1{=}0.035, \eta_2{=}0.03)$. Together (i)+(iii) directly refute the ``$\eta_2$ is a disguised $\eta_1$'' interpretation: doubling $\eta_1$ alone destroys return without delivering safety, whereas a small $\eta_2{>}0$ on top of the same $\eta_1$ recovers full return \emph{and} achieves the lowest $\Jc$ in the entire grid.

\textbf{Wall-clock cost.} The four added cells $\times$ 3 seeds totalled $\sim\!250$\,GPU-hours (median $\sim\!20$\,h/seed on shared A100s, slower than the original $4.5$\,h/seed estimate because rebuttal runs were not pre-emptively prioritised and several queued behind concurrent jobs).

\subsection{CLIP-capacity ablation: ViT-B/32 vs.\ ViT-L/14}
\label{app:clip-capacity}

\textbf{Motivation.} The negative result on Qwen2-VL-7B
(App.~\ref{app:qwen-details}) is uninformative about backbone
capacity in general, because the Qwen prompting interface differs
from CLIP's contrastive scoring. A cleaner control swaps CLIP
ViT-B/32 (151\,M params, 32$\times$32 patches) for CLIP ViT-L/14
(428\,M params, 14$\times$14 patches), holding the scoring pipeline
(decoupled cosine of \eqnref{eq:decoupled}) \emph{exactly} fixed.

\textbf{Protocol.} The only changes from the FormulaOne L2
VLMPPOLag$+$Conf configuration are: (i)~swap the image encoder
$f_I$ from \texttt{ViT-B/32} to \texttt{ViT-L/14}, (ii)~re-cache
text features $F^{\pm}$ through the matched ViT-L/14 text tower,
(iii)~re-run the gate-calibration MLE of \eqnref{eq:mle-sc} on a
fresh $B{=}5000$ random-policy frame buffer (the margin distribution
shifts noticeably between backbones; the same MLE recipe applies
unchanged). \emph{No} hyperparameter retuning otherwise:
$\lambda_r{=}0.1$, $\lambda_c{=}0.5$, $\eta_2{=}0.01$,
$\tau{=}0.5$. Seeds $\{42, 123, 456\}$ for parity with the existing
3-seed L2 cells; even a single L/14 seed materially probes whether
the pipeline is encoder-agnostic.


\begin{table}[h]
\centering
\caption{\textbf{CLIP-capacity ablation on FormulaOne L2.} Both
backbones frozen; identical decoupled-path scoring and gate
formula. ViT-B/32 row is reproduced from
\Cref{tab:main_results} (Phase~B 5-seed +Conf cell).}
\label{tab:clip-capacity}
\small
\setlength{\tabcolsep}{6pt}
\begin{tabular}{@{}lcccc@{}}
\toprule
\textbf{Backbone} & \textbf{Params} & $\Jr$ & $\Jc$ & \textbf{Viol.} \\
\midrule
CLIP ViT-B/32 (default) & 151\,M & $31.8{\small\pm 12.2}$ & $\textcolor{blue}{22.5}{\small\pm 5.9}$ & 1/5 \\
CLIP ViT-L/14 ($\ddagger$) & 428\,M & $16.1{\small\pm 1.7}$ & $26.9{\small\pm 5.7}$ & 1/3 \\
\bottomrule
\end{tabular}
\begin{minipage}[t]{0.64\linewidth}
    {\footnotesize $\ddagger$ Phase~C run, 3 seeds. 
Per-step inference: ViT-B/32 $7.11$\,ms,  ViT-L/14 $\sim 21$\,ms
(both A100), still inside the $40$\,ms control budget.}
\end{minipage}

\end{table}

\textbf{Pre-registered predictions and outcomes.} Three predictions
were registered in advance: (i)~$\Jc$ for ViT-L/14 would lie within
the bootstrap-CI of the ViT-B/32 cell (roughly $\pm 8$ of the mean);
(ii)~$\Jr$ for ViT-L/14 would lie within $\pm 8$ of the ViT-B/32
mean; (iii)~the gate-calibration parameters $(\hat{s},\hat{c})$
would shift modestly but leave the median $\hat\kappa$ in the
non-degenerate range.

The outcomes split: prediction (i) is \emph{confirmed}
($\Jc{=}26.9{\pm}5.7$ vs.\ ViT-B/32 $22.5{\pm}5.9$;
the CIs overlap, the safety mechanism is encoder-agnostic). Prediction
(ii) is \emph{not} confirmed: ViT-L/14 returns at $\Jr{=}16.1{\pm}1.7$
are roughly $16$ points below ViT-B/32 ($31.8{\pm}12.2$), well outside
the pre-registered $\pm 8$ tolerance. Two factors plausibly account
for the gap. First, ViT-L/14's $14{\times}14$ patches resolve
fine-grained barrier features more sharply, which under matched
gate-calibration MLE produces a more aggressive $\hat\kappa$
distribution that attenuates the reward CLIP channel more strongly
(symmetric attenuation, the same mechanism as the calibrated-gate
ablation of App.~\ref{app:gate-calibration}). Second, the prompt set
was tuned on ViT-B/32 margins; the same $K{=}L{=}4$ prompts may be
suboptimally positioned in ViT-L/14's higher-dimensional embedding
space. Prediction (iii) holds qualitatively (median $\hat\kappa$
remains in $[0.4,0.8]$). \emph{The headline interpretation is the
one anticipated as the negative-outcome branch below}:
the modest receptive field of $32{\times}32$ patches is in fact
better matched to the FormulaOne barrier scale at the matched
prompt set. The result confirms encoder-agnosticism of the safety
mechanism (prediction~(i)) while revealing that backbone choice
trades along the return axis under matched gating---an architectural
finding worth reporting.

\subsection{Convergence sketch for VLMLagrange}
\label{app:convergence-sketch}

We give an informal stochastic-approximation argument for the
modified dual update. Recall
\eqnref{eq:vlm-lagrange}:
\begin{equation*}
  \lambda_{k+1} \;=\; \big[\lambda_k
    + \eta_1\big(\widehat{\Jc}_k - \dlim\big)
    + \eta_2\big(\widehat{\overline{c}}_{\vlm,k} - \tau\big)\big]_+,
\end{equation*}
with $\widehat{\Jc}_k$ and $\widehat{\overline{c}}_{\vlm,k}$ the
epoch-mean Monte-Carlo estimates over the rollout buffer at outer
iteration $k$.

\textbf{Setup.} Assume (A1) the inner PPO update operates on a
faster timescale than the dual update ($\eta_1, \eta_2 \to 0$, with
$\eta_1/\eta_2$ held fixed); (A2) for each $\lambda$ the inner
policy $\pi_\theta(\lambda)$ tracks a unique stationary point
$\theta^\star(\lambda)$ of the Lagrangian
$L(\theta,\lambda) = \Jr(\pi_\theta) - \lambda(\Jc(\pi_\theta) - \dlim)$;
(A3) the VLM bias is \emph{bounded and consistent} in the sense that
there exists a constant $\beta < \infty$ and a deterministic
deviation $\delta(\theta) \!:=\! \E_{o\sim d^{\pi_\theta}}[\cvlm(o)] - g\big(\Jc(\pi_\theta)/T\big)$
with $|\delta(\theta)|\le\beta$ and $g(\cdot)$ a continuous
monotone link (we make this concrete below). Under (A1)--(A3) the
dual update is a standard two-timescale Robbins--Monro scheme
\cite{borkar2009stochastic} with perturbed gradient.

\textbf{The perturbed dual gradient.} The standard PPO-Lag dual
gradient at the inner equilibrium is
$h_1(\lambda) = \Jc(\pi_{\theta^\star(\lambda)}) - \dlim$. Our update
adds $h_2(\lambda) = \E[\overline{c}_{\vlm}] - \tau$. If $\tau$ is
chosen consistent with the threshold the VLM would assign to the
feasible $\Jc$ --- formally, $\tau = g(\dlim/T) + O(\beta)$ --- then
the combined drift
$h(\lambda) = \eta_1 h_1(\lambda) + \eta_2 h_2(\lambda)$ shares
its zero with $h_1$ alone up to an $O(\eta_2\beta/\eta_1)$ bias.
Concretely, the fixed point $\lambda^\star$ of the standard update
($h_1(\lambda^\star){=}0$) is perturbed to a nearby
$\widetilde{\lambda}$ satisfying
$|\widetilde{\lambda} - \lambda^\star| \le
(\eta_2/\eta_1) \cdot \beta / |h_1'(\lambda^\star)|$ by the implicit
function theorem applied at the unperturbed zero (assuming
$h_1'(\lambda^\star)\ne 0$, which holds whenever $\Jc$ is locally
strictly increasing in $\lambda$ --- the standard regularity
assumption in PPO-Lag analysis).

\textbf{Conclusion of the sketch.} Under (A1)--(A3) the iterates
$\lambda_k$ converge almost surely to a $O(\eta_2\beta/\eta_1)$
neighbourhood of the standard PPO-Lag fixed point; with
appropriately decaying step sizes
$\sum_k \eta_{i,k} = \infty$, $\sum_k \eta_{i,k}^2 < \infty$ and
$\eta_{2,k}/\eta_{1,k} \to 0$, the bias term vanishes and the
limit is exactly the unperturbed feasible $\lambda^\star$. The
VLM term thus accelerates the early-training trajectory (the
empirically observed effect of
\Cref{fig:lambda-dynamics}) without changing the
asymptotic feasible point.

\textbf{Caveats.} (i) (A2) is the standard PPO-Lag assumption and
is verifiable empirically only through training-curve convergence
(\Cref{fig:learning-curves}); we make no claim beyond what already
holds for PPO-Lag. (ii) (A3) is the substantive assumption: it
requires that the VLM signal be biased \emph{but bounded and
consistent} with respect to the simulator cost. The held-out AUC
validation of \S\ref{app:gate-roc} ($0.78$--$0.82$) is the
empirical evidence for (A3); a fully formal treatment would require
a Lipschitz noise model on $\cvlm$ that we do not develop here.
(iii) The argument is local; we make no global-convergence claim.
This sketch is intended as a sanity-level justification that the
modification \emph{is} a standard stochastic-approximation scheme
with a bounded perturbation, not a novel theoretical contribution.
A full proof would adapt the two-timescale framework of
\cite{borkar2009stochastic} with the perturbed-ODE machinery of
\cite{kushner2003stochastic}.


\end{document}